\documentclass[sigconf]{acmart}

\AtBeginDocument{%
  }

\copyrightyear{2022} 
\acmYear{2022} 
\setcopyright{rightsretained} 
\acmConference[FAccT '22]{2022 ACM Conference on Fairness, Accountability, and Transparency}{June 21--24, 2022}{Seoul, Republic of Korea}
\acmBooktitle{2022 ACM Conference on Fairness, Accountability, and Transparency (FAccT '22), June 21--24, 2022, Seoul, Republic of Korea}\acmDOI{10.1145/3531146.3533199}
\acmISBN{978-1-4503-9352-2/22/06}

\usepackage{multirow} 
\usepackage{caption}
\usepackage{subcaption} 
\usepackage{algorithm} 
\usepackage{wrapfig}
\usepackage{amsthm}
\usepackage{amsmath}
\usepackage{amsfonts}

\usepackage{tikz} 
\usetikzlibrary{bayesnet}
\usetikzlibrary{arrows}

\definecolor{ruby}{rgb}{0.88, 0.07, 0.37}
\definecolor{royalazure}{rgb}{0.0, 0.22, 0.66}

\definecolor{asparagus}{rgb}{0.01, 0.75, 0.24}
\definecolor{bittersweet}{rgb}{1.0, 0.44, 0.37}

\definecolor{asparagus}{rgb}{0.53, 0.66, 0.42}
\definecolor{chocolate}{rgb}{0.82, 0.41, 0.12}
\definecolor{cadmiumgreen}{rgb}{0.0, 0.42, 0.24}
\definecolor{ao(english)}{rgb}{0.0, 0.5, 0.0}

\newrobustcmd{\B}{\bfseries}

\newcommand{\Ecal}{ {  \mathcal{E} } }
\newcommand{\Yprox}{ {  {\tilde{Y}} } }
\newcommand{\Y}{ {  {{Y}} } }
\newcommand{\X}{ {  {\mathbf{X}} } }
\newcommand{\Z}{ {  {\mathbf{Z}} } }

\newcommand{\E}{ {  {\mathbb{E}} } }
\newcommand{\V}{ {  {\mathbf{V}} } }

\newcommand{\z}{ {  {\boldsymbol{z}} } }
\newcommand{\x}{ {  {\boldsymbol{x}} } }
\newcommand{\xf}{ {  {\boldsymbol{x}^F} } }
\newcommand{\xcf}{ {  {\boldsymbol{x}^{CF}} } }

\newcommand{\yprox}{ {  {\tilde{y}} } }

\newcommand{\y}{ {  {{y}} } }

\newcommand{\real}{ {\mathbb{R}} }
\newcommand{\dep}{ { \not\perp} }

\newcommand{\mudp}{ {\mu_{\operatorname{DPU}}} }
\newcommand{\muut}{ {\mu_{\widetilde{\operatorname{UT}}}} }
\newcommand{\tvdp}{ {\operatorname{TV}_{\operatorname{DPU}}} }
\newcommand{\tvut}{ {\operatorname{TV}_{\widetilde{\operatorname{UT}}}} }
\newcommand{\bld}[1]{\mathbf{#1}}

\newcommand{\indep}{ { {\perp} } }

\newcommand{\name}{{FairAll}} 
\newcommand{\nameone}{{VAE}}
\newcommand{\nametwo}{{FairAll (II)}}
\newcommand{\nameonetwo}{{FairAll (I+II)}}
\newcommand{\phasetwo}{{Phase~II}}
\newcommand{\phaseone}{{Phase~I}}
\newcommand{\Uprox}{{{\tilde{U}}}}
\newcommand{\uprox}{{{\tilde{u}}}}

\newcommand{\policlf}{{{\pi^{(X,S)}_\text{clf}}}}
\newcommand{\polidec}{{{\pi^{(X,S)}_\text{dec}}}}

\newcommand{\UT}{{{$\operatorname{UT}$}}}
\newcommand{\UTprox}{{{$\widetilde{\operatorname{UT}}$}}}
\newcommand{\DPU}{{$\operatorname{DPU}$}}
\newcommand{\CFU}{{$\operatorname{CFU}$}}

\newcommand{\TV}{{$\operatorname{TV}$}}

\newcommand{\niki}{{UnfairLog}}
\newcommand{\nikifair}{{FairLog}}
\newcommand{\ipsvae}{{FairLab (I+II)}}
\newcommand{\ipsvaelab}{{FairLabVAE-}}

\newcommand{\optimalunfair}{{OPT-UNFAIR}}
\newcommand{\optimalfair}{{OPT-FAIR}}

\newcommand{\COMPAS}{{COMPAS}}
\newcommand{\CREDIT}{{CREDIT}}

\newcommand{\MEPS}{{MEPS}}
\newcommand{\SCB}{{Synthetic}}

\newcommand{\HARSH}{HARSH} 
\newcommand{\LENI}{LENI}

\newcommand{\A}{{{\bld{A}}}}

\newcommand{\ALt}{{{\bld{A}^{L}_t}}}
\newcommand{\AULt}{{{\bld{A}^{UL}_t}}}

\newtheorem{lemma}{Lemma}

\newtheorem{definition}{Definition}[section]

\begin{document}

\title{Don’t Throw It Away! The Utility of Unlabeled Data in Fair Decision Making}

\author{Miriam Rateike}
\authornote{Both authors contributed equally to this research.}
\affiliation{%
\institution{MPI for Intelligent Systems}
  \city{Tübingen}
  \country{Germany}
}
\affiliation{%
  \institution{Saarland University}
  \city{Saarbrücken}
  \country{Germany}
}
\email{mrateike@tue.mpg.de}

\author{Ayan Majumdar}
\authornotemark[1]
\affiliation{%
  \institution{MPI for Software Systems}
    \city{}
  \country{}
}
\affiliation{%
  \institution{Saarland University}
    \city{Saarbrücken}
  \country{Germany}
}
\email{ayanm@mpi-sws.org}

\author{Olga Mineeva}
\affiliation{%
  \institution{ETH Zürich}
      \city{Zürich}
  \country{Switzerland}
  }
  \affiliation{%
    \institution{MPI for Intelligent Systems}
        \city{Tübingen}
  \country{Germany}
  }
  \email{omineeva@ethz.ch}

\author{Krishna P. Gummadi}
\affiliation{%
  \institution{MPI for Software Systems}
    \city{Saarbrücken}
  \country{Germany}
}
\email{gummadi@mpi-sws.org}

\author{Isabel Valera}
\affiliation{
 \institution{Saarland University}
   \city{}
  \country{}
 }
\affiliation{%
  \institution{MPI for Software Systems}
    \city{Saarbrücken}
  \country{Germany}
}
\email{ivalera@cs.uni-saarland.de}


\begin{abstract}
Decision making algorithms, in practice, are often trained on data that exhibits a variety of biases.
Decision-makers often aim to take decisions based on some ground-truth target that is assumed or expected to be unbiased, i.e., equally distributed across socially salient groups.
In many practical settings, the ground-truth cannot be directly observed, and instead, we have to rely on a biased proxy measure of the ground-truth, i.e., \emph{biased labels}, in the data.
In addition, data is often \emph{selectively labeled}, i.e., even the biased labels are only observed for a small fraction of the data that received a positive decision.
To overcome label and selection biases, recent work proposes to learn stochastic, exploring decision policies via i) online training of new policies at each time-step and ii) enforcing fairness as a constraint on performance.
However, the existing approach uses only labeled data, disregarding a large amount of unlabeled data, and thereby suffers from high instability and variance in the learned decision policies at different times.
In this paper, we propose a novel method based on a variational autoencoder for practical fair decision-making. 
Our method learns an unbiased data representation leveraging both labeled and unlabeled data and uses the representations to learn a policy in an online process.
Using synthetic data, we empirically validate that our method converges to the \emph{optimal (fair) policy} according to the ground-truth with low variance.
In real-world experiments, we further show that our training approach not only offers a more stable learning process but also yields policies with higher fairness as well as utility than previous approaches.
\end{abstract}



\maketitle

\section{Introduction} 

The extensive literature on fair machine learning has focused primarily on studying the fairness of the predictions by classification models~\cite{dwork2012fairness,hardt2016equality,zafar2017bfairness,agarwal2018reductions,zafar2019fairness} deployed in critical decision-making scenarios.
Consider a university admissions process where the goal is to admit students based on their \emph{true potential}, which may not be directly observable.
{We assume that a student's ground-truth potential is independent of and unbiased by, their assignment to various socially salient groups - defined by sensitive characteristics (race, gender) protected by anti-discrimination laws~\cite{barocas2016big}}. 
That is, we hold it as self-evident that students of different socially salient groups are endowed with a similar (equal) distribution of potential.
{In practice, ground-truth potential cannot be measured directly and remains unobserved.
Instead, we rely on \emph{proxy labels}, which we assume to contain information about the ground truth.}
However, due to structural discrimination, these proxy labels are often biased measures of ground truth. 
{For example, prevailing societal discrimination may result in students with similar ground-truth potential but different assigned genders having a very different distribution of university grades (proxy labels).}
This phenomenon is termed \emph{label bias}~\cite{wick2019unlocking} and has been studied extensively in ~\cite{zemel2013learning,hardt2016equality,zafar2017bfairness,agarwal2018reductions}.
{These fairness studies assume that independent and identically distributed (i.i.d.) labeled data is available for training.}
However, in decision-making scenarios, {data may} also suffer from \emph{selection bias}~\cite{lakkaraju2017selective}.
That is, we only observe the labels of a small fraction of the data which received positive decisions.
{For example,} we observe (biased) university grades of only the admitted students, {resulting in} biased, non-i.i.d. labeled data.

In decision-making scenarios affected by \emph{both label bias and selection bias}, \citet{kilbertus2020fair} show that to learn the optimal policy, it is  necessary to
move from {learning} fair predictions (e.g., predicting grades) to learning
fair decisions (e.g., deciding to admit students). 
To tackle label bias, the authors introduce fairness constraints in the optimization problem. 
To address {selection bias}, they propose to learn \emph{stochastic, exploring} decision policies in an online learning process, where a new decision policy is {learned} at each time-step. 
To get unbiased loss estimates from non-i.i.d. labels, the authors further rely on inverse propensity scoring (IPS)~\cite{horvitz1952IPS}. 

{However, the approach ignores 
unlabeled data. 
A large fraction of data in the learning process may remain unlabeled due to receiving a negative decision (e.g., students denied admission).}
Using only labeled data, the approach suffers from high instability and
variance in the learning process. 
{In particular, i) the method may give very different outcomes to the same individual, depending on the random initializations of the learning process, and ii) the method may give the same individual very different outcomes at different points in time.}

In this paper, we propose a novel online learning process for fair decision-making that leverages both labeled and unlabeled data. 
Our method learns fair representations of all data using latent variable models in an attempt to capture the unobserved and unbiased ground truth information. 
In turn, these representations are used to learn a policy that approximates the optimal fair policy (according to the unobserved ground truth). 
Importantly, as shown in our experiments, our approach leads to a stable, fair learning process, achieving decision policies with similar utility and fairness measures 
across time and training initializations.

Our primary contributions in this paper are listed below: 
\begin{enumerate}
    \item We propose a novel two-phase decision-making framework that utilizes both labeled and unlabeled data to learn a policy that converges to the optimal (fair) policy with respect to the unobserved and unbiased ground truth.
    \item We present a novel policy learning framework, \name\ that relies  on a VAE (a latent variable model) architecture to significantly reduce the need for bias correction of selective labeling.
    \item Through theoretical analyses and empirical evaluation on synthetic data, we show that the VAE from our \name\  framework is able to learn an unbiased data representation that captures information from the ground truth.
    \item Through extensive evaluations based on real-world data, we show that \name, compared to prior work, offers a significantly more effective and stable learning process, achieving higher utility and fairness.
\end{enumerate}

\subsection{Related Work}\label{sec:related_work}

\paragraph{Fair Classification}
There exists a variety of approaches for fair classification to tackle biased labels. 
\emph{In-processing methods} optimize for correct predictions under additional fairness constraints~\cite{dwork2012fairness,zafar2017bfairness,agarwal2018reductions,zafar2019fairness}. 
This requires formulating differentiable fairness constraints and often lead to unstable training \cite{cotter2019two}.
\emph{Pre-processing methods} instead utilize representation learning first to learn a fair data representation. 
This representation is then used for downstream predictive tasks \cite{zemel2013learning}.
Different methods for fair representation learning have been brought forward, including variational autoencoders (VAEs) \cite{louizos2015variational, moyer2018invariant}, normalizing flows \cite{balunovic2021fair}, and generative adversarial networks \cite{xu2018fairgan}. To enforce independence between the learned representation and the sensitive attribute, some methods condition deep generative models on the sensitive features~\cite{creager2019flexibly, madras2018adversarially-learning, moyer2018invariant, grari2020adversarial}, revert to disentanglement~\cite{creager2019flexibly}, perform adversarial training~\cite{madras2018adversarially-learning, song2019learning, grari2020adversarial} or add regularization, like Maximum-Mean-Discrepancy~\cite{louizos2015variational, grari2020adversarial}. 
While most work on fair representation learning focuses on satisfying group fairness notions \cite{louizos2015variational, creager2019flexibly}, some have also considered individual fairness \cite{ruoss2020learning} and counterfactual fairness \cite{grari2020adversarial}. 
Recently, contrastive learning for fair representations has attracted much attention \citep{park2022fair}. However, it requires the definition of a similarity measure and meaningful data augmentations. {This is non-trivial, especially for tabular data. While some recent work~\citep{bahri2021scarf} exists, further research is needed.}
 
Although all of the works above use fully labeled training data, some have studied fair classification in the presence of partially labeled data~\cite{louizos2015variational, zhu2021rich, zhang2020fairness}. 
Further, all of these works assume access to a \emph{biased proxy label} and not the ground truth.
\citet{zafar2017fairness} considered analyzing fairness notions separately, assuming access to a ground-truth label.
Further, the notions of biased observed proxies and unbiased, unobserved ground-truth (denoted as construct spaces) were discussed in~\cite{friedler2016possibility,dutta2020there}.
Note that all of these studies also assume i.i.d. data. 
But, in most real-world scenarios, the semi-labeled data is not i.i.d. (selection bias \cite{lakkaraju2017selective}).
\citet{wick2019unlocking} perform an initial fairness-accuracy analysis of classifiers with respect to label and selection bias.
Our work, similar to~\cite{kilbertus2020fair} aims to tackle both label and selection bias while transitioning from a static classification setting to online decision-making.

\paragraph{Fair Online Decision Making}
Recent works \cite{kilbertus2020fair, bechavod2019equal} have started exploring fairness in online decision-learning processes in the presence of partially labeled non-i.i.d. data.
In such settings, convergence to the optimal policy requires exploration and stochastic policies \cite{kilbertus2020fair}.
\citet{kilbertus2020fair} use an extra fairness constraint in the loss to trade-off between utility and fairness.
Additionally, they correct for selection bias in training using \textit{inverse propensity scoring} (IPS) \cite{horvitz1952IPS} on the entire loss function, which, unfortunately, can introduce additional variance. 
\citet{bechavod2019equal} derive an oracle-efficient bandit algorithm to learn an accurate policy while explicitly controlling the exploration-exploitation trade-off, and thus the variance. 
However, both approaches disregard a major portion of the data that receives the negative decision and remain unobserved.
We show how this data contain useful information about the underlying data distribution. 
In our approach, we posit utilizing this unlabeled data to reduce the need for IPS.
We empirically validate how this helps in faster convergence to an optimal decision policy while providing high utility and fairness during training.\footnote{{In 
Section~\ref{sec:evaluation} we compare our method to \cite{kilbertus2020fair}. Note that \cite{bechavod2019equal} is a theoretical work that provides neither experimental results nor an implementation and thus prevents us from comparing to them as the baseline.}}
\begin{figure*}[t]
  \centering
     \begin{subfigure}[t]{0.45\textwidth}
         \centering
              \input{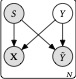}
          \caption{Ground truth generative process.}
          \label{fig:generative}
     \end{subfigure}
    \hspace{1 cm}
          \begin{subfigure}[t]{0.45\textwidth}
         \centering
              \input{our_model}
          \caption{\name\ latent variable model. 
          }
          \label{fig:our_model}
     \end{subfigure}
     \caption{(a) Ground truth data generative process and (b) our \name\ generative {(solid)} and inference ({dashed}) models. Sensitive attribute $S$, non-sensitive attribute $\X$, proxy label $\Yprox$, ground truth label $\Y$ and decision $D$. Observed (unobserved) random variables grey (white). }
    \Description{This figure contains two graphical models: The first figure (A) depicts the ground truth data generative process, and the second figure (B) shows our FairAll latent variable model. The first figure (A) contains four random variables. Of those are three observed: the sensitive attribute S, non-sensitive attributes X, and proxy label. The fourth variable, the ground truth target Y, remains unobserved. Directed arrows indicate the generative model. X is generated by S and Y, and the proxy label is generated by S and Y. The second figure (B) shows our FairAll latent variable model. The figure contains four random variables, and two fixed variables. The random variables are: the observed non-sensitive attribute X, the observed proxy utility, the unobserved latent variables Z, and the semi-observed proxy label. The fixed and observed variables are: sensitive attribute S, and decision D. Solid directed arrows indicate the generative model. X is generated by S and Z, the proxy utility is generated by S, Z, D, and the proxy label. Dashed directed arrows indicate the inference model. Z is inferred from S, X, and proxy utility. Proxy utility is inferred from D, S, and X. }
\end{figure*}

\section{Background and Problem Setting}\label{sec:background}
Let us consider a university admission decision-making process inspired by \cite{Kusner2017-ln}, which we will use as a running example. We use uppercase letters for random variables and lowercase letters for their assignments. With $p$, we optionally refer to a probability distribution or a probability mass function.
Let $S$ be a random variable indicating a \emph{sensitive attribute} of an individual describing their membership in a socially salient group (e.g., gender). 
For simplicity we assume binary $S \in \{-1,1\}$.
Let $\X \in \real^n$ be a set of $n$ \emph{non-sensitive features} that are observed {(e.g., high school grades)},
and may be influenced by $S$. 
 The  university aims to take an admission decision $D \in \{0, 1\}$ based on a {\emph{ground truth target} }
 $\Y$ {(e.g., intellectual potential)}~\cite{friedler2016possibility,dutta2020there}.
For simplicity we assume $\Y \in \{0,1\}$. Importantly, throughout this paper, we assume that 
$Y \indep S$ (e.g., {potential}
is equally distributed across social groups), such that an optimal policy
decides $D \indep S$.\footnote{Note, depending on which label $\Y$ refers to, $\Y \indep S$ may not always hold in practice. See Section \ref{sec:discussion} for a discussion of this assumption.}

\subsection{{Label Bias and Selection Bias}}
\paragraph{Label Bias} 
In practice, the ground truth $\Y$ often remains unobserved (as it cannot be directly measured). 
Instead, as shown in Figure~\ref{fig:generative}, we observe a different label $\Yprox$ (e.g., {semester grades}) that is assumed to contain information on $\Y$ along with measurement noise.
We {refer to} this label as the \emph{proxy {label}}.
For simplicity, we assume $\Yprox \in \{0,1\}$. A data-generative process exhibits \emph{label bias}, if $\Yprox \dep S$, i.e., the proxy target is biased by the sensitive attribute \cite{wick2019unlocking}. For example, the same {potential}
may result in higher grades for one demographic group over another due to existing structural discrimination. Figure~\ref{fig:generative} highlights our assumed {data} generative process with biased labels.
Recall that we aim to take decisions according to $\Y \indep S$. 
However, in the biased label scenario, both $\X$ and $\Yprox$ are biased by $S$.
A policy that maps $\X$ (and potentially $S$) to $\Yprox$ will thus -- in the absence of fairness constraints --  take biased decisions.

\paragraph{Selection Bias} 
In practice, algorithms often also need to learn from partially labeled data, where labels $\Yprox$ are observed only for a particular (usually positive) decision. 
This is called the \emph{selective labels problem} \cite{lakkaraju2017selective}. 
For example, a university only knows whether a student gets good semester grades if it accepts the student in the first place. 
Let these decisions be taken according to policy $\pi$, which may be biased and not optimal. 
For example, for an individual with features $(\x, s)$, a decision $d$ may be taken according to $d \sim \pi(\x, s)$.\footnote{A policy always takes as input features of an individual. Here, we assume the features to be $(\X, S)$. However, they could also be only $\X$ or some feature representation.} 
Then a {labeled} data point $(\x, s, \yprox)$ observed under a policy $\pi$ is \emph{not an i.i.d. sample} from the true distribution $p(\X, S, \Yprox)$.  
Instead, the data is sampled from the distribution {induced by the probability of a positive decision $\pi(d=1 \mid \x, s)$ such that}:
$p_{\pi}(\X, S, \Yprox) \propto p(\Yprox \mid \X, S) \pi(D=1 \mid \X, S) p(\X, S)$ \cite{kilbertus2020fair}. 

\citet{corbett2017algorithmic} have shown that deterministic decisions (e.g. taken by thresholding) are optimal {with respect to $\Yprox$} for i.i.d. data.
However, \citet{kilbertus2020fair} demonstrated that, if labels are not i.i.d.,
we require \emph{exploration}, i.e., stochastic decision policies. 
Such policies map features to a strictly positive distribution over $D$. This implies that the probability of making a positive decision for any individual is never zero.
Exploring policies are trained in an \emph{online} fashion, where the policy is updated at each time step $t$ as $\pi_t$.
{Moreover, we typically learn a policy from labeled data by minimizing a loss that is a function of the revealed labels.}
However, if labels are non-i.i.d., it is necessary to perform bias correction to get an unbiased loss estimate.
A common technique for such bias correction
is \textit{inverse propensity score} (IPS) weighting~\cite{horvitz1952IPS}. 
It divides the loss for each labeled datum by the {probability
with} which it was labeled, i.e., received a positive decision under policy $\pi$.
However, this bias correction may lead to high variance in the learning process, when {this probability}
is small.

\subsection{Measures of Interest: Utility, Fairness, and Their Temporal Stability}
\label{subsec:fair_opt_pols}
Assuming an incurred {cost $c$} for every positive decision (e.g., university personnel and facility costs)~\cite{corbett2017algorithmic}, the decision-maker aims to \emph{maximize its profit} (revenue $-$ costs), which we call \emph{utility}. 
We define utility $U = D(\Y - c)$ as a random variable that can take on three values {$U \in \{-c, 1-c, 0\}$} depending on decision $D$. 
A correct positive decision results in a positive profit of $1-c$
(admitting students with high potential leads to more success and funding),
an incorrect positive decision results in a negative profit $-c$ (sunk facility costs), and a negative decision (rejecting students) in zero profit. 
{The utility of a policy is then defined as the expected utility $U$ with respect to population $p(\X, S, \Y)$ and policy $\pi$:} 

\begin{definition}[Utility of a policy \cite{kilbertus2020fair}]\label{eq:utility}
Given utility as a random variable $U=D(\Y-c)$, we define the utility \UT\ of a policy $\pi$ as the expected overall utility $\operatorname{UT}(\pi):=\mathbb{E}_{\x, s, y \sim p(\X, S, \Y)}[\pi(D=1 \mid \x, s) \allowbreak  \left(\Y   -c\right)], $
where decision and label are $D, \Y \in \{0,1\}$, and $c \in (0,1)$ is a problem specific cost of taking a positive decision.
\end{definition}

{Note, we defined \UT\ with respect to {ground truth target}
$\Y$. However, as mentioned above, in most practical settings, we only observe proxy $\Yprox$ and can thus only report \UTprox, i.e., the expected utility measured with respect to proxy $\Yprox$.}

As detailed above, we are interested in taking decisions according to $\Y$, where $\Y \indep S$ ({e.g., potential}
is equally distributed across sensitive groups). 
A policy that takes decisions based on $\Y$ satisfies \emph{counterfactual fairness} \cite{Kusner2017-ln} and \emph{demographic parity}  \cite{dwork2012fairness} .
This follows directly from the fact that $\Y$ is a non-descendant of $S$. Any policy $\pi$ that is a function of the non-descendants of $S$ (namely $\Y$) is demographic parity and counterfactually fair \cite{Kusner2017-ln}.\footnote{Since $\Y \indep S$, a policy that decides according to $\Y$ also satisfies equal opportunity \cite{Kusner2017-ln}.} 
The notion of DP fairness for a policy $\pi$ requires the proportion of decision $d$ to be the same across all social groups {:} 

\begin{definition}[Demographic Parity Unfairness of a Policy \cite{dwork2012fairness}]\label{eq:dp} We define the demographic parity unfairness (\DPU) of a policy $\pi$ with respect to sensitive attribute $S\in \{-1, 1\}$ and decision $D \in \{0,1\}$: 
\begin{displaymath}
\begin{split}
    \operatorname{DPU}(\pi) & = \big | \mathbb{E}_{\x \sim p(\X \mid S=1)}[\pi(D=1 \mid \x, S=1)] \\
    & - \mathbb{E}_{\boldsymbol{x}\sim p(\X \mid S=-1)}[\pi(D=1 \mid \x, S=-1)] \big |
\end{split}
\end{displaymath}
\end{definition}
Correspondingly, a policy is counterfactually fair if it assigns the same decision to an individual in the observed (or, factual) world as well as in a \emph{counterfactual world}, in which the individual belongs to a different sensitive group\footnote{{For example, {where the individual had been growing up with a different sensitive identity like gender.}}}.
\begin{definition}[Counterfactual Unfairness of a Policy \cite{Kusner2017-ln}] \label{eq:CFU} The counterfactual unfairness (\CFU) of policy $\pi$ with respect to a factual individual belonging to $S = s$ with features $\xf$ and the decision $D \in \{0, 1\}$ can be defined as:
%
\begin{displaymath}
    \begin{split}
            \operatorname{CFU}(\pi) = & \mathbb{E}_{{(\xf, s) \sim p(\X, S), {\xcf \sim p(\X \mid \xf, d o\left(S=s'\right) )}}} \\ 
            & \left| \pi(D=1 \mid \xf, s) - \pi(D=1 \mid \xcf, s') \right|
    \end{split}
\end{displaymath}
Here, $\xf$ refers to the non-sensitive features of an individual in the factual world with sensitive attribute $s$, and $\xcf$ refers to the non-sensitive features of the same individual in a counterfactual world, where its sensitive attribute is $s'$ with $s' \neq s$.
\end{definition}
Note, from~\cite{Kusner2017-ln} that satisfying counterfactual fairness implies satisfying demographic parity but not vice-versa.
Further, counterfactual analysis requires hypothetical interventions on $S$ and exact knowledge of the causal generation process.
While estimation techniques for real-world data exist \cite{khemakhem2021causal, sanchez2021vaca}, in this paper, we only analyze for synthetic data (with access to the true exogenous variables and the structural equations). See Appendix~C.4
for more details.

The above metrics allow assessing the performance of one particular policy. 
However, the online policy learning process outputs,
over $T$ training steps, the set of policies ${\Pi_{t=1}^{T}:= \{\pi_{1} \dots \pi_{T}\}}$.
{Assume we wish to stop the learning process from time $t_1$. Can we reliably deploy any policy $\pi_{t \geq t_1}$? 
Inspired by prior work on temporal fairness~\cite{celis2018algorithmic, gupta2018temporal}, we propose a new notion of \textit{temporal variance} (\TV) for a policy learning process. 
\TV\ indicates how much a metric $M$ (e.g., utility, fairness) varies for the set of policies across some time interval $[t_1, t_2]$. }
\begin{definition}[Temporal Variance of a Policy Learning Process]\label{def:TV}
We define the temporal variance (\TV) of {the outcome of} a policy learning process ${\Pi_{t_1}^{t_2} := \{\pi_{t_1} \dots \pi_{t_2}\}}$ in time interval $[t_1, t_2]$ with respect to metric $\operatorname{M}$ as:  
\begin{displaymath}
    \operatorname{TV}_M(\Pi_{t_1}^{t_2}) = \sqrt{\frac{1}{t_2-t_1}\sum_{t=t_1}^{t_2}\left[(\operatorname{M}({\pi_t})-\mu_M)^{2}\right]}
\end{displaymath} 
where $\mu_{M} = \frac{1}{t_2-t_1}\sum_{t=t_1}^{t_2} \operatorname{\operatorname{M}}(\pi_t)$ denotes the temporal average for the metric $M$ over the time interval $[t_1, t_2]$. 
\end{definition} 
{High \TV\ denotes an unstable learning process, where policies of different time steps achieve different utility and fairness levels for a fixed group of people.
For example, policy $\pi_{t_1}$ may treat the same group of individuals very different compared to $\pi_{t_1+1}$.
A low \TV\ on the other hand indicates a stable learning process, where policies of different time steps achieve similar utility and fairness levels. Hence, it is safe to stop the learning process any time after $t_1$.}
Note, $\mu_{M}$ measures the average metric value (e.g. utility, fairness) over the time interval $t = [t_1, t_2]$.

\subsection{Variational Autoencoder}
Deep generative models (DGMs), like Variational Autoencoders (VAEs) \cite{kingma2013auto, rezende2014stochastic}, Normalizing Flows, \cite{rezende2015variational} and Generative Adversarial Networks \cite{goodfellow2014generative} are latent variable models (LVMs) that \emph{estimate complex data distributions} by capturing hidden structures in the latent space $Z$.

VAE is one of the most prominent DGMs. It jointly learns a probabilistic generative model $p_{\theta}(\X \mid \Z)$ (\emph{decoder}) and an approximate posterior estimator $q_{\phi}(\Z \mid \X)$ (\emph{encoder}). Encoder and decoder are parameterized by neural networks.
As the marginal likelihood $p(\x) = \int p(\x, \z) dz$ is intractable, a VAE is trained by maximizing the evidence lower bound (ELBO) of the observations $\x \sim p(\X)$, consisting of the expected log-likelihood and the posterior-to-prior {KL} divergence:

\begin{displaymath}
    \begin{split}
        \log p(\x) & \geq  \underbrace{\mathbb{E}_{z \sim q_{\phi}(\Z \mid \x)}\left[\log p_{\theta}(\x \mid \z)\right]}_{\text{Exp. log-likelihood}}-\underbrace{\operatorname{K L}\left(q_{\phi}(\Z\mid \x) \| p(\Z)\right)}_{\text{KL divergence}} \\
        & = \operatorname{ELBO}(\theta, \phi; \x)
    \end{split}
\end{displaymath}

\section{Learning to decide fair}
\label{sec:learn_decide_fair}

Let us assume that the data $p(\X, S, \Y)$ has a generative process as shown in Figure~\ref{fig:generative}\footnote{Note, Figure~\ref{fig:generative} is the same as the causal model presented as the \textit{Scenario 3: University success} in \cite{Kusner2017-ln}.}.
Recall that a decision-maker ideally aims to take decisions $d$ according to 
the unbiased ground truth $\Y$, i.e., 
$d\sim p(\Y)$\footnote{
{Note, this is formulated abusing notation for simplicity. Here, decision $D$ is a deterministic function of $\Y$, i.e., for a data point $d=\y$, where $\y \sim \pi(\Y)$.}
} \cite{kilbertus2020fair}.
However, in practice, $\Y$ remains unobserved.
{Consider for now a setting {with label bias but no selection bias.} We have access to i.i.d. samples from the underlying distribution and the observable label is a biased proxy $\Yprox$. }
As per Figure~\ref{fig:generative}, observed features $\X$ and proxy labels $\Yprox$ both contain information about $\Y$, but are biased by $S$ {(label bias)}, i.e., $\X \dep S$, $\Yprox \dep S$.
Hence, a policy that takes decisions $d_{\text{prox}} \sim p(\Yprox \mid \X,S)$ from such biased observed data  is unfair.

Assuming access to only biased data, we posit using a conditional latent variable model for fair decision making.
As we theoretically show, with the help of a conditional latent variable model (LVM), it is possible to learn a latent representation $\Z$ that: i) is independent of the sensitive $S$, i.e.,  $\Z \indep S$ and ii) captures the information contained in $\Y$, up to the noise of the observed features and the approximation error of the LVM.

We assume observed features $\X$ and proxy labels $\Yprox$ are generated by $\Y, S$ and an independent noise variable $\mathcal{E} \in \mathbb{R}$.
%
%
\begin{lemma}\label{lemma:lemmaone}
{Assume the observed $\{\X, \Yprox\}$ is a bijective function of the ground-truth $\Y$, sensitive $S$ and noise $\Ecal$ with $S, \Y, \Ecal$ being pairwise independent. Then, the conditional data entropy is  
$H(\X, \Yprox \mid S) = H(\Y) + H(\X, \Yprox \mid S, \Y)= H(\Y) + H(\Ecal)$.}
\end{lemma}
So, the conditional data distribution $p(\X, \Yprox \mid S)$ captures the information of the unobserved ground-truth $\Y$, up to the extent of noise $\Ecal$.
Next, we consider approximating the underlying data distribution with LVMs.
 %
%
\begin{lemma}\label{lemma:lemmatwo}
Given a latent variable model conditional on $S$ and input data $\{\X,\Yprox\}$, having encoder $q_{\phi}(\Z\mid \X,\Yprox, S)$ and decoder $p_{\theta}(\X,\Yprox \mid \Z, S$), the mutual information between latent variable $\Z$ and the conditional data distribution $p(\X, \Yprox \mid S)$ is ${I( \Z ; \X, \Yprox \mid S)} = {H(\X, \Yprox \mid S) - \Delta}$ with approximation error $\Delta$.
\end{lemma}
Hence, the information captured by the latent $\Z$ reduces the uncertainty about the conditional distribution $p(\X,\Yprox \mid S)$ up to the error.
We refer to Appendix~A.2 
for the detailed proof following~\cite{alemi2018information}.
Combining  the two lemmas, we get:
\begin{equation}
\label{eq:lemmaboth}
    I(\Z;{\X,\Yprox}\mid S) = \underbrace{H(\Y)}_{{\text{information of }\Y}} + \underbrace{H(\Ecal)}_{\text{noise}} -  \underbrace{\vphantom{- H(\Ecal)} \Delta}_{\vphantom{noise of \V}\text{approx. error}}
\end{equation}
The two lemmas together show that using a conditional LVM to model the observed data $p(\X,\Yprox \mid S)$ allows us to learn a latent variable $\Z$ that captures the information of the unobserved ground-truth $\Y$, up to the extent of noise (note that $\Delta$ is also dependent on $\Ecal$).
Consequently, a policy that learns to make decisions using the latent $\Z$ with respect to the proxy $\Yprox$ would, in fact, make decisions based on the information contained in $\Y$ (up to the effect of noise $H(\Ecal)$).

{As pointed out in Section~\ref{sec:background}, following \cite{Kusner2017-ln}, a policy deciding based on $\Y$ satisfies \emph{counterfactual fairness} and \emph{demographic parity}.}
Hence, a policy $\pi$ mapping from $\Z$ to $\Yprox$ {tackles label bias and} satisfies both fairness notions without the need for additional constraints (up to the distortion due to $H(\Ecal)$ and $\Delta$).
Following, in Section~\ref{sec:our-approach}, we propose a pipeline to learn a fair policy using unbiased representations $\Z$ from \emph{non-i.i.d.} data that suffer from both label and selection bias.

\section{Our Approach}\label{sec:our-approach}
{In this section, we propose a novel online fair policy learning framework \name\ for tackling both \emph{biased} and \emph{selective} labels.}
 Our \name \ framework consists of: i) a fair representation learning step that relies on a VAE-based model 
 (illustrated in Figure~\ref{fig:our_model}) trained on  both labeled and unlabeled data and; ii) a policy learning approach that leverages the learned fair representations to approximate the optimal fair policy according to the ground truth $\Y$. 
 {Both steps of our framework, i.e., the VAE and the policy, are {continually} optimized as more data becomes available through the development of previous policies (i.e., \emph{in an online manner}).}
 {Note, in taking decisions based on a fair representation, our policy mitigates \emph{label bias}; in learning a stochastic policy in an online manner, we allow for exploration during training, which mitigates \emph{selection bias}~\cite{kilbertus2020fair}.} 
{Specifically, we correct \emph{label bias} by conditioning the VAE on sensitive $S$, while we correct \emph{selection bias} by weighting our online learning loss with IPS.}

In the following, we first detail how to {use both labeled and unlabeled data to} learn a fair representation and then describe decision learning with policy $\pi$. Lastly, we present an overview of our entire fair online policy learning pipeline and propose a method to further exploit unlabeled population information.

\subsection{Learning a Fair Representation}
\label{sec:semiVAE}
Following the result in Eq.~\ref{eq:lemmaboth}, we aim to learn latent $\Z$ that is both informative of $\Y$ and independent of the sensitive attribute $S$, i.e., $\Z \indep S$. 
Consider an online setting in which we have access to a dataset {$\A_t$} (\emph{applicants}) at each time step $t$.
The partitioning of {$\A_t$} 
into labeled data {$\ALt$} 
(\emph{accepted applicants}) and unlabeled {$\AULt$}
(\emph{rejected applicants}) is invoked by policy {$\pi_t$.}
{To ease notation, we will, in the following, consider a particular time step $t$ and omit the subscript.}

Let us recall that for each data observation, we only observe  the proxy label $\Yprox$ if the previous policy made the positive decision $D=1$ (labeled data), and the actual value of the ground truth remains unobserved.
However, we can leverage the fact that the utility with respect to the proxy label, $\Uprox = D(\Yprox - c)$, is always observed.\footnote{As per Section~\ref{subsec:fair_opt_pols}, utility U can take three values dependent on the decision, hence providing a value for accepted and rejected applicants.} 
{This allows us to learn an unbiased latent representation $\Z$ from both labeled and unlabeled data.}

\paragraph{VAE-Based Fair Representation Learning}
Specifically, we build on previous work on semi-supervised and conditional VAEs \cite{kingma2014semi, rezende2015variational, sohn2015learning} to approximate the conditional distribution  $p_\theta(\X,\Uprox|S,D)=  \int p_\theta(\X,\Uprox, \Z|S,D) dZ$ (generative model), {and}
the posterior over the fair latent representation 
\[{q_{\omega, \phi}(\Z|\X,S, D=1) = \int q_\phi (\Z|\X, S, \tilde{u}, D=1) q_\omega (\tilde{u}|\X,S, D=1) d\tilde{u}}.\]
The inference model contains an encoder $q_\phi$ and a separate classifier model $q_\omega$. {Note, we condition the inference model on $d=1$ (see Appendix~C.1 
for an overview) and thus introduce the classifier to predict the label for any unlabeled data point.}
We optimize the model parameters ($\theta,\phi$, and $\omega$) by minimizing the following objective function:
\begin{equation}
\begin{split}
    &J(\theta, \phi, \omega) = \alpha \underbrace{\E_{(\x,\tilde{u}, s)\sim \mathbf{A}^L} [\mathcal{R}(\omega; \x, s, \tilde{u}, \pi)]}_{\text{IPS-weighted classification loss}} \\
    &- \underbrace{\E_{(\x,\tilde{u},s)\sim \mathbf{A}^L, d=1} [\mathcal{L}(\theta, \phi; \x, s, \tilde{u})] - \E_{\x,s\sim \mathbf{A}^{UL}, d=0} [\mathcal{U}(\theta, \phi; \x, s)]}_{\text{ELBO}}
\end{split}
\end{equation} \label{eq:overall_loss}

where the latter terms corresponds to the ELBO capturing the goodness of fit of the VAE; and the first term measures the accuracy of the classifier model that estimates the utility of labeled data. 
The hyperparameter $\alpha$ balances the classification loss term relative to ELBO. 
Notice that defining our model with respect to the completely observed $\Uprox$ allows us to compute the ELBO on both labeled data and unlabeled data, removing the need for IPS.\footnote{Previous work \cite{sohn2015learning} defined a semi-supervised VAE with respect to partially observed label $\Yprox$. In this case, one would need to apply IPS on the ELBO, which may introduce high variance.}

\paragraph{Evidence Lower Bound (ELBO)}
More specifically, 
for all accepted applicants ($d=1$), we compute the ELBO of $\log p_{\theta}(\X,\Uprox|S,D=1)$ as:
\begin{equation} \label{eq:elbo_sup}
\begin{split}
    \mathcal{L}(\theta, & \phi;  \x,s, \tilde{u}) = \E_{\z \sim q_\phi (\Z|\x,s,\tilde{u},D=1)} [\log p_{\theta}(\x|\z,s) \\ 
    & + \log p_{\theta}(\tilde{u}|\z,s,D=1)] - KL (q_{\phi}(\Z|\x,\tilde{u},s, D=1)||p(\Z)) 
\end{split}
\end{equation}
For all rejected applicants, which received $d=0$, utility is a constant $\Uprox=0$, such that $\log p(\Uprox=0|\Z,D=0)$ = $\log 1 = 0$. Thus, the ELBO for unlabeled data is given by:
\begin{equation} \label{eq:elbo_unsup}
\begin{split}
    \mathcal{U}(\theta, \phi; \x,s) &= \E_{\tilde{u} \sim q_\omega (\Uprox|\x,d=1)} \E_{\z \sim q_\phi (\z|\x,\tilde{u},d=1)}  [\log p_{\theta}(\x|\z,s)] \\
    &-  KL [\int q_\phi (\z|\x,\tilde{u},s, d=1) q_\omega (\tilde{u}|\x,s, d=1) d\tilde{u}||p(\Z))
\end{split}
\end{equation}

\paragraph{Classification Loss} 
We utilize labeled data and cost-sensitive cross-entropy loss to train the classifier $q_\omega(\Uprox|\X,S,D=1)$.
Based on decision costs, false negatives are weighed by the lost profit $(1-c)$, and false positives by $c$.
Note for labeled data, we have a binary prediction task as $\Uprox \in \{-c,1-c\}$, which in this context can be taken as $\{0,1\}.$ 
As we learn only on labeled data, we apply IPS weights based on the policy $\pi$ to correct for selective bias as:
\begin{equation}
    \label{eq:cross-entropy}
    \begin{split}
        \mathcal{R}(\omega; \x, s, & \tilde{u}, \pi) =  -\Big[  c (1-\tilde{u}) \log (1 -  q_{\omega}(\tilde{u}=1|\x,s, d=1)) \\ 
        &+ (1-c) \tilde{u}\log q_{\omega}(\tilde{u}=1|\x,s, d=1)\Big] \underbrace{\frac{1}{\pi(d=1|\x, s)}}_{\text{IPS}}
    \end{split}
\end{equation}
In our implementation, we learn functions
$p_\theta(\X|\Z, S)$, $p_\theta(\Uprox | \Z, S, D=1)$, $q_\omega(\Uprox|\X, S, D=1)$, $q_\phi(\Z|\Uprox,S,\X, D=1)$ with fully-connected (deep) neural networks. 
See Appendix~C 
and~D
for practical considerations and training setups. 

\subsection{Learning a Fair Policy}
At each time step after improving our representation learning model, we update the policy $\pi$ that maps the fair representation $\Z$ into a distribution over decisions $D$. 
For each data point in the complete dataset $\mathbf{A} = \mathbf{A}^L \cup \mathbf{A}^{UL}$ we sample $\z$ using the  \emph{encoder}  $q_\phi(\Z|\tilde{u},s,\x, D=1)$, and use
the decoder to get an estimated proxy utility, $\tilde{u} \sim p_\theta(\Uprox | \Z, S, D=1)$.
Note, for $D=1$, the proxy utility is binary.
We thus train the policy by minimizing the cross-entropy binary loss: $(1-\Uprox)\log \pi(\Z) + \Uprox*\log(1-\pi(\Z))$.
We describe other options for the policy model for our approach in Appendix~D.4.

\subsection{Exploiting Fully Unlabeled Data} 
In real-world settings, a decision-maker often has {prior} access to large unlabeled datasets containing i.i.d. samples from the population of interest. For example, in the case of university admissions, a university may have access to a database of all students who passed their high school diploma to enter university -- including those who did not apply at that particular university. 
{Such  dataset contains features $\X, S$, but no labels $\Yprox$, i.e., it is fully unlabeled.} 
{However, as we show, such unlabeled data can significantly improve the policy learning process.}
In particular, the fully unlabeled dataset can help learn fair representations $\Z$ by approximating the conditional $p(\X\mid S)$. That is, we can learn a VAE optimized as: 
%
\begin{equation}
\begin{split}
    \mathcal{K}(\theta', \phi' ; x, s)& =\mathbb{E}_{z \sim q_{\phi'}(\z\mid \x, s)}\left[\log p_{\theta'}(\x \mid \z, s)\right] \\ & - \operatorname{KL}\left(q_{\phi'}(\Z \mid \x, s) \| p({\Z})\right)
\end{split}
\end{equation} \label{eq:phaseone_loss}
The resulting unsupervised VAE model can then be used to initialize the parameters of the semisupervised VAE proposed in Section~\ref{sec:semiVAE} via transfer learning. 
Transfer learning is typically studied in supervised learning~\cite{yosinski2014transferable,sharif2014cnn}, where models are pre-trained on large datasets of one domain and then transferred to a different domain with fewer data.
In our proposed \emph{two-phase} approach, we utilize the unlabeled data in a \phaseone\ to initialize our semi-supervised VAE model of the online decision-making \phasetwo.
We initialize parameters $\phi$ and $\theta$ of semi-supervised VAE with the trained parameters $\phi'$ and $\theta'$ of the unsupervised VAE. 
Note from Eq.~\ref{eq:overall_loss} that the semi-supervised VAE encoder additionally takes $\Uprox$ as input and that the decoder also outputs $\Uprox$. 
To account for this, we add \emph{new neural connections}.
At the encoder, we add connections from input $\Uprox$ to each neuron in the first hidden layer. 
At the decoder's output, we add a new head to output $\Uprox$.
The new connections are initialized randomly at the start of \phasetwo.

\begin{figure*}[t] 
 \centering
\includegraphics[width=0.9\linewidth]{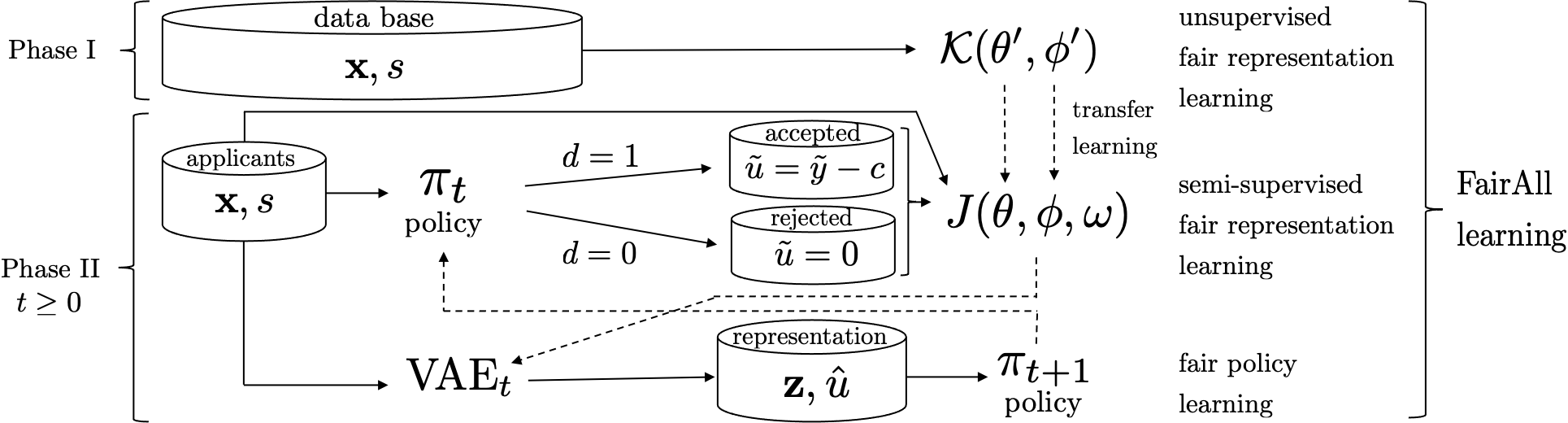}
 \caption{Pipeline of our approach \name.
In \phaseone, we pre-train the \nameone\ using a large pool of unlabeled data with features $\x, s$. We transfer the trained parameters $\phi'$ and $\theta'$ of the VAE at the start of \phasetwo. Next, at each time step $t$, decisions $d$ for a new batch of data are drawn from the current policy $\pi_t$. In case of acceptance ($d=1$), labels $\yprox$ are revealed. The VAE is updated via loss $J$ using both labeled and unlabeled data. 
Subsequently, the updated VAE outputs fair representations $\z$ and estimated utility $\hat{u}$ for all applicants. The policy is updated with these fair representations.  In this way, \name\ learns to decide by using all available data.
 }
 \Description{A figure showing the overall pipeline of our approach FairAll, which we sometimes also call FairAll I+II. The figure shows two phases of learning. In the first phase, a database of features X and sensitive attribute S is input to the unsupervised fair representation learning loss function K. In the second phase, for each time step greater than or equal to zero, a pool of applicants with features X and sensitive attribute S are input to the existing policy pi t. The policy decides to accept some applicants and reject others. The accepted applicants have a proxy utility computed as proxy label minus costs c. The rejected applicants have a proxy utility of zero. The accepted and rejected applicants' features are then input to the semi-supervised fair representation learning loss J. This loss J corresponds to the V.A.E at the time step t. A dashed arrow from J to V.A.E shows this. Dashed arrows show how the parameters from the phase 1 are used to initialize the learning in phase 2 through transfer learning. Representations Z and utility values U hat are output from V.A.E. The Z and U hat are used to train a new policy pi t+1. The old policy is updated with the new one. A dashed arrow from the new policy to the old one shows this. This is the fair policy learning. The unsupervised fair representation learning, semi-supervised fair representation learning, and the fair policy learning together comprise the pipeline FairAll learning.}
 \label{fig:pipeline}
\end{figure*}

\subsection{\name{} Overview}
Our \name{} learning framework is illustrated in Figure~\ref{fig:pipeline} and  consists of two phases. 
{In the first step (\phaseone), we  learn a fair representation using an unsupervised VAE trained in an \emph{offline} manner using only unlabeled data.
We then use the resulting model to  initialize the parameters of the semisupervised VAE via transfer learning. In a second step (\phasetwo), we enter the \emph{online} decision-learning process, where at each time step, we first update our semisupervised VAE using both labeled and unlabeled data, and then update the decision policy $\pi$.}

\section{Experimental Results} \label{sec:evaluation}
In this section, we evaluate our fair policy learning framework with regard to i) its convergence to the optimal policy; ii) its training effectiveness until convergence; and iii) its deployment performance after convergence. 
{We can evaluate (i) on synthetic data only, and evaluate (ii) and (iii) on real-world datasets.}
\paragraph{Baseline and Reference Models.} We perform rigorous empirical comparisons among the following learning frameworks:

\begin{itemize}
    \item \textbf{\nameonetwo}: \emph{Our complete proposed} learning framework  including  \phaseone\ (i.e., offline unsupervised representation learning) and \phasetwo\ (online semisupervised representation and policy learning). 
    \item \textbf{\nametwo}:
    \emph{Baseline} approach that make use of  only  \phasetwo\ of the proposed \name. This approach allows us to evaluate the impact of \phaseone, i.e., fully unlabeled data. 
    \item \textbf{\ipsvae}: 
    \emph{Baseline} approach that consist of unsupervised \phaseone\ and a \emph{fully supervised} \phasetwo\, using only the IPS-weighted ELBO on labeled data. It allows us to evaluate the importance of unlabeled data in \phasetwo. 
     \item \textbf{\nikifair}~\cite{kilbertus2020fair}: \emph{Competing approach} that minimizes the IPS weighted cross entropy (Eq.~\ref{eq:cross-entropy}),
     denoted by $\mathcal{L}^{\text{\niki}}$, with a Lagrange fairness constraint, i.e.,  $\mathcal{L}^{\text{\niki}}+ \lambda * \text{DPU}$ with \DPU\  as defined in Def.~\ref{eq:dp}.
    \item \textbf{\niki}~\cite{kilbertus2020fair}: \emph{Unfair reference} model, corresponding to \nikifair\ without fairness constraint, i.e., $\lambda = 0$. 
    It allows us to measure the \emph{cost of fairness}. 
\end{itemize}

We refer to the Appendix for a detailed description of baselines and competing methods (Appendix~D.5),
%
details on the hyperparameter selection (Appendix~D.2)
and other practical considerations (Appendix~C).

\paragraph{Metrics.} 
We measure \emph{observed proxy} \emph{utility} \UTprox\ (Def.~\ref{eq:utility} {w.r.t. $\Yprox$}) and \emph{demographic parity unfairness} \DPU\ (Def.~\ref{eq:dp}) on {i.i.d.} test data on both synthetic and real-world datasets.
Further, on synthetic data with access to the ground truth generative process, we report {\emph{counterfactual unfairness}} \CFU\ (Def.~\ref{eq:CFU})\footnote{See Appendix~C.4 
for further details on generating counterfactuals on the synthetic dataset.} as well as the \emph{unobserved ground truth utility} \UT\  {(Def.~\ref{eq:utility} {w.r.t. $\Y$})}.
{For the real-world settings, we report \textit{effective} \UTprox\ \cite{kilbertus2020fair},  which is the average \UTprox\ accumulated by the decision-maker on the training data up to time $t$. Similarly, we report \textit{effective} \DPU.
{We also report the \emph{temporal {variance}} (Def.~\ref{def:TV}) of \UTprox\ and \DPU\ over time interval $t=[125, 200]$.}}

\paragraph{Datasets.}
We report results on one synthetic and three real-world datasets (more details in Appendix~B):
\begin{itemize}
    \item \SCB, where $\X$ contains 2 features, with Gender $S$ and \emph{grades after university admission} $\Yprox$ (note that the ground truth \emph{intellectual potential} $\Y$ is considered unobserved and only used for evaluation).
    \item \COMPAS~\cite{angwin2016machine,larson2016how}, where $\X$ contains 3 features, with Race $S$ and \emph{no recidivism} $\Yprox$.
    \item \CREDIT~\cite{dua2019}, where $\X$ contains 19 non-sensitive features, with Gender $S$ and \emph{credit score} $\Yprox$.
    \item \MEPS~\cite{meps2018}, where $\X$ contains 39 non-sensitive features, with Race $S$ and \emph{high healthcare utilization} $\Yprox$.
\end{itemize}

\paragraph{{Optimal Policies.}} {In our \SCB\ setting, where observed $\X \dep S$ and unobserved $K\indep S$, the optimal unfair policy (\optimalunfair) takes decisions $d \sim p(\Yprox \mid X, S)$ and the optimal fair policy (\optimalfair) decides $d \sim p(\Yprox \mid K)$.  \optimalunfair\ can be approximated with access to i.i.d. samples from the posterior distribution, while \optimalfair\ additionally requires access to unobserved $K$. See Appendix~C.3 
for details.}

\paragraph{Setup.}
{We assume access to the proxy labels.}
{Since labeled data is often scarce in practice,}
we assume an initial \HARSH\ policy 
which labels around 10-18\% of the data {that} we see {in \phasetwo\ at $t=0$.}
{We report} results {for a lenient} policy in Appendix~E.
%
For details on initial policies, see Appendix~D.1. 
%
{The decision cost is $c=0.1$ for \MEPS\ and  $c=0.5$ for all other datasets. See Appendix~E.4 
for a case study on the impact of the cost value. Wherever applicable, \phaseone\ was trained over a large number of epochs. \phasetwo\ was trained for 200 time steps with the same number of candidates in each step. All policy training were done over 10 independent random initializations. For a full description of the experimental setup, see Appendix~D.
Our code is publicly available\footnote{\url{https://github.com/ayanmaj92/fairall}}.}

\begin{figure*}[t]
	\centering
    	\begin{subfigure}[b]{0.28\linewidth}
    		\centering
    		\includegraphics[width=2.36\linewidth]{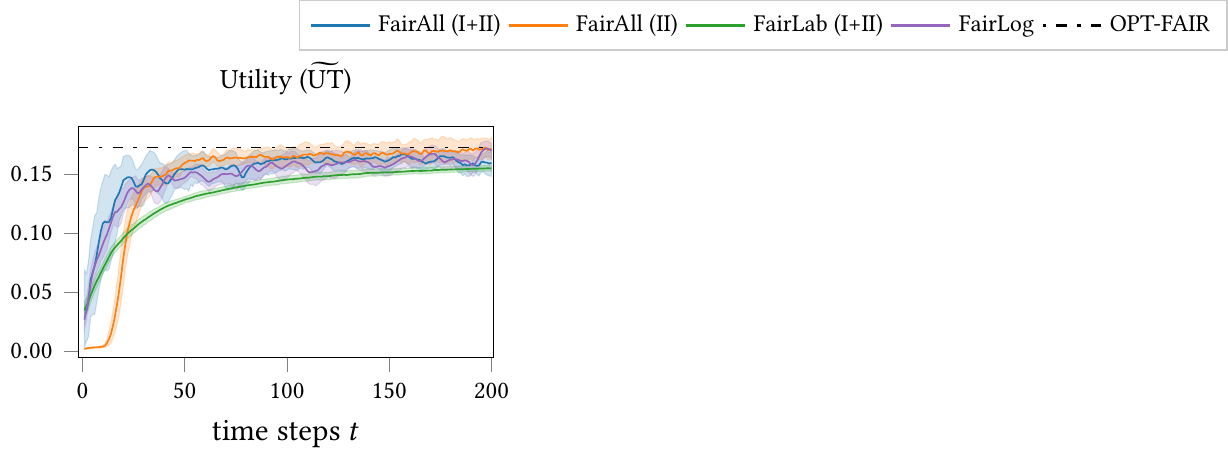}
    	\end{subfigure}
    	\begin{subfigure}[b]{0.28\linewidth}  
    		\centering 
    		\includegraphics[width=\linewidth]{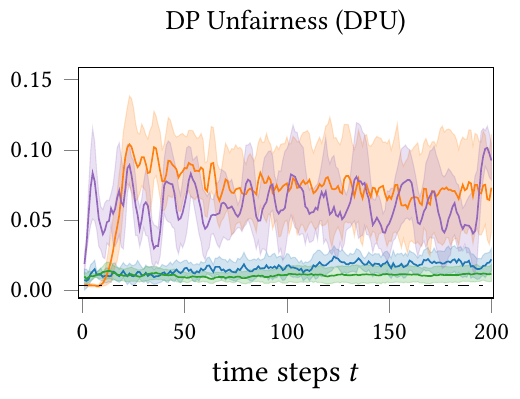}
    	\end{subfigure}
    	\begin{subfigure}[b]{0.28\linewidth}  
    		\centering 
    		\includegraphics[width=\linewidth]{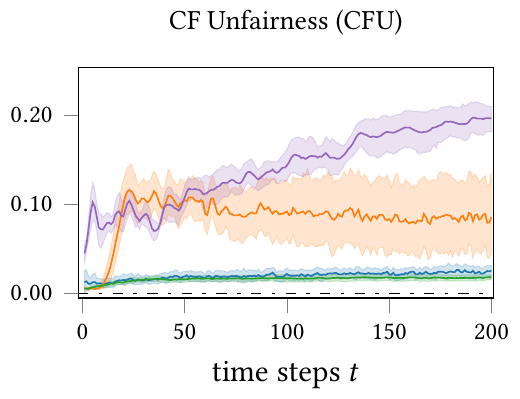}
    	\end{subfigure}
\caption{
{Utility (\UTprox) with respect to the proxy, demographic parity unfairness (DPU), and counterfactual unfairness (CFU) on the \SCB\ dataset. \nameonetwo\ converges to the optimal {fair} policy in utility and fairness while being more fair than \nikifair.}}
 \Description{The figure shows three line plots to compare the performance of different online learning models in terms of proxy utility, demographic parity unfairness and counterfactual unfairness. All three plots have time steps from 0 to 200 as x axis. All three plots show four lines, each for one of the methods and a fifth line for the optimal policy. 
 The first plot shows the values of proxy utility from 0 to 0.19 on the y axis.  For all methods utility increases rapidly in the first 25 time steps and then increases slightly with different slopes and different values at time step 200. FairAll I+II and FairLog show similar utility values across time, converging to the optimal fair utility of around 0.19 at time step 200. FairAll II converges slower, while FairLab I+II converges to utility value lower than the optimal fair value. 
 The second plot shows the values of demographic parity unfairness (DPU) on y-axis ranging from 0 to 0.15. FairAll I+II and FairLab I+II are almost constant and show near 0 unfairness. FairAll II and FairLog show higher mean unfairness, and also higher variance of unfairness, with values ranging from 0.05 to 0.1. 
 The third plot shows the values of counterfactual unfairness (CFU) on the y-axis ranging from 0 to above 0.2. FairAll I+II and FairLab I+II increase over time and at time step 200 shows very low CFU value of 0.025 with low variance. FairAll II increases rapidly in the first 25 time steps and then stays around the same value with a slight decline and at time step 200 shows a CFU value between 0.08 and 0.1 with very high variance. FairLog increases over time starting at time step 0 from around 0.1 and rising to near 0.2 at time step 200.}
\label{fig:scb_results}
\end{figure*}

\subsection{Can We Reach the Optimal Fair Policy?} 
\label{sec:eval_synth}
{In this section, we use synthetic data to evaluate if, given enough time-steps, the different learning methods yield policies that converge to the \emph{optimal (fair) policy} (\optimalfair) both in terms of proxy utility and fairness.} 

\paragraph{{Results.}} 
Figure~\ref{fig:scb_results} reports \UTprox, \DPU\ and \CFU\ on test data across time.
Recall that \ipsvae\ uses unlabeled data only in \phaseone, \nametwo\ only in \phasetwo , and \name\ (I+II) in both \phaseone\ and \phasetwo.
\nametwo\ starts convergence after approximately $20$ steps. \ipsvae\ starts at higher utility but then exhibits slower convergence behavior and has not fully converged at $t=200$. \name\ (I+II) instead starts at a higher utility and  at $t=200$ yields utility and fairness (both in demographic parity and counterfactual fairness) values close to \optimalfair. 
Regarding fairness, \nametwo\ does not converge to \optimalfair\ and exhibits high variance in DPU and CFU both across seeds and over time. Instead, \name\ (I+II) and \ipsvae\ both rely on unsupervised fair representation learning (Phase I) to  converge to a value very close to the optimal one. Thus, the fully unsupervised \phaseone\ appears crucial for convergence to the optimal fair policy with low variance.
Comparing \name\ to \nikifair,
we observe that both methods asymptotically converge to the optimal fair utility.
However, \nikifair\ suffers from higher and more noisy DPU while CFU \emph{increases over time}.
 This can be explained by the fact that \nikifair\ enforces \DPU\ constraints, 
 and, as shown empirically, \DPU\ \emph{does not imply} \CFU.
 In contrast, \nameonetwo\ learns a fair representation $\Z$ that achieves \CFU\ and, as a result, also \DPU\ (see Section~\ref{sec:learn_decide_fair}). 

This evaluation concludes that i) all fair models approximately converge to the optimal utility; ii) unlabeled data helps in faster convergence; iii) utilizing \phaseone\ leads to significantly lower unfairness; 
iv) our approach \name, compared to \nikifair, achieves convergence while satisfying both counterfactual and demographic parity fairness notions.

\begin{figure*}[t]
	\centering
	\begin{subfigure}[b]{0.28\linewidth}
		\centering
		\includegraphics[width=2.37\linewidth]{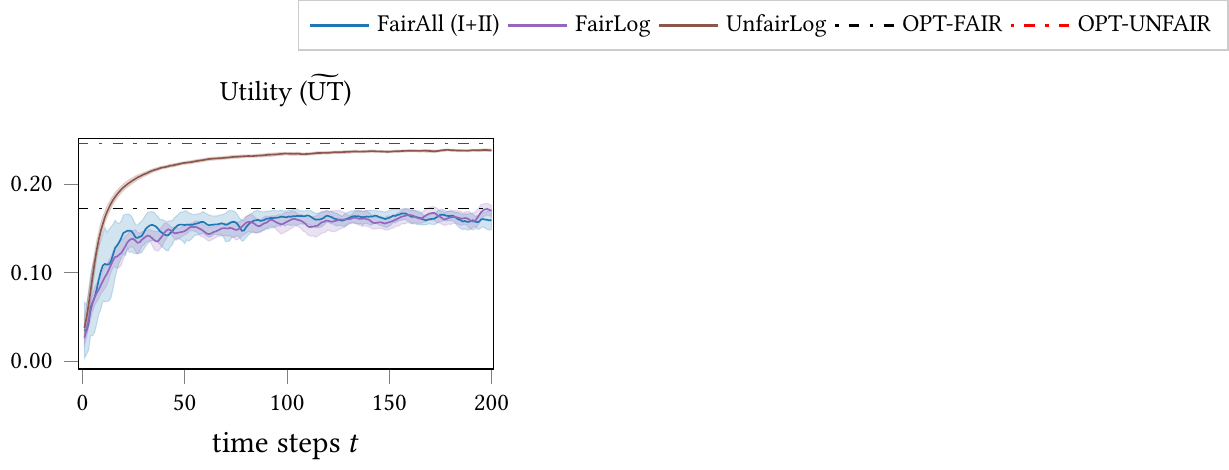}
	\end{subfigure}
	\begin{subfigure}[b]{0.28\linewidth}  
		\centering 
		\includegraphics[width=\linewidth]{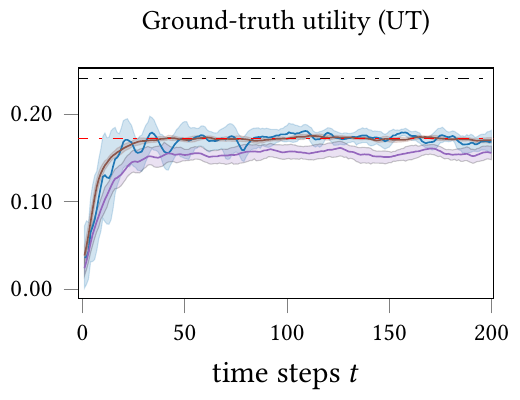}
	\end{subfigure}
	\begin{subfigure}[b]{0.28\linewidth}  
		\centering 
		\includegraphics[width=\linewidth]{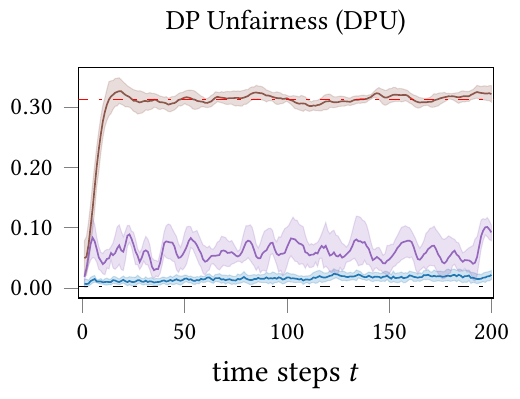}
	\end{subfigure}
\caption{
{Results on the \SCB\ dataset. For \nameonetwo\ and \niki\ a trade-off between fairness, i.e., demographic parity (DPU), and utility when measured with respect to proxy labels (\UTprox) -- but not when measured with respect to ground truth (\UT). } }
 \Description{The figure shows three line plots to compare the performance of the different models in terms of proxy utility, ground truth utility and demographic parity unfairness. All three plots show three lines, each for one of the methods FairAll I+II, FairLog and UnfairLog, and also two lines for the two optimal policies, OPT-FAIR and OPT-UNFAIR. The plots show that measuring utility to the ground truth may do away the utility-fairness tradeoff that we observe when measuring utility with respect to the proxy label. All the three plots have time steps from 0 to 200 as x axis. 
 The first plot shows utility measured with respect to the proxy label (proxy utility) on the y-axis, with values ranging from 0 to 0.24. As for the optimal policies OPT-UNFAIR has a value of 0.24, OPT-FAIR has a value of 0.17. UnfairLog converges to optimal unfair utility value of around 0.24 after about 50 time steps, while FairAll I+II and FairLog both converge to optimal fair utility value of around 0.16. 
 The second plot shows utility measured with respect to the ground truth target (ground truth utility) on the y-axis, ranging from 0 to 0.24. As for the optimal policies, OPT-FAIR has a value of 0.24, OPT-UNFAIR has a value of 0.17. Now, both unfair and fair methods, converge to the OPT-UNFAIR value.  While FairLog converges to a ground truth utility of 0.16, FairAll I+II and UnfairLog reach around 0.17. 
 The third plot shows demographic parity unfairness (DPU), ranging from 0 to 0.32 in the y-axis. As for the optimal policies, OPT-FAIR has a value of near 0, OPT-UNFAIR has a value of 0.31. While UnfairLog converges to DPU of 0.31, FairAll I+II has a DPU value of nearly 0. FairLog has a DPU value of around 0.05 to 0.06 with high variance.}
\label{fig:scb_util_gnd}
\end{figure*}

\subsection{Do We Actually Trade-off Utility for Fairness?}
In Figure~\ref{fig:scb_util_gnd}, we observe that \name\ yields  higher fairness but lower observed \emph{proxy utility} \UTprox\ compared to the unfair reference model \niki. This observation is often referred to as a \textit{fairness-accuracy trade-off} \cite{barocas19fairmlbook}, which assumes that fairness comes at the cost of utility (or accuracy in predictive settings). As pointed out in~\cite{wick2019unlocking, dutta2020there}, if utility is a function of biased labels, then the utility measurement is also biased. 
{Recall the assumption that a decision-maker aims to take decisions based on $\Y$ and aims to maximize \UT. For example, potential ($\Y=1$) drives a successful career, not high university grades ($\Yprox =1)$. In the synthetic setting, we can measure unbiased \emph{ground truth utility} \UT.}

\paragraph{Results.} 
{In Figure~\ref{fig:scb_util_gnd}, we observe that both \name\ and \niki\ achieve a similar level of ground truth utility \UT, while \name\ reports significantly less discrimination (\DPU). This suggests that a decision-maker \emph{may not} {actually} trade-off fairness and {(true)} utility, although we observe lower proxy utility. 
Note that despite being fair, \name\ (and \nikifair) do not reach the \UT\ of the optimal (fair) policy \optimalfair.
This could be due to the noise $\Ecal$ in the dataset, which is known to the optimal policy, but is naturally not captured by the models.}
An in-depth discussion of the phenomenon is outside the scope of this paper, and we refer the reader to the literature \cite{wick2019unlocking, cooper2021emergent, dutta2020there}.

\subsection{How Effective Is the Learning Process?}
\label{sec:eval_effective}
We have shown
that 
\name\ {asymptotically} outputs a policy that approximates the optimal. 
Now we investigate how much it costs the decision-maker in terms of utility and the society in terms of fairness to learn this policy. 
{We evaluate the \emph{effective} proxy \UTprox, and \DPU\ that the online learning process accumulates across time on \emph{real-world datasets}.}

{ 
\begin{table*}[t]
\caption{Accumulated utility w.r.t. observed proxy (Effect. \UTprox) and demographic parity unfairness (Effect. \DPU) measured during the policy learning process for different real-world datasets over time interval $t=[0,200]$. We report mean values over the same ten independent seeds. The numbers in the brackets show the standard deviation. All reported values are multiplied by 100 for readability.}
 \Description{Results}
  \label{tab:eff_util_fair_results}
 \small
\begin{tabular}{@{\extracolsep{4pt}}lcccccccccc}
\toprule
\multirow{2}{*}{\textbf{Model}} &
  \multicolumn{2}{c}{COMPAS} &
  \multicolumn{2}{c}{\CREDIT} &
  \multicolumn{2}{c}{\MEPS} \\
  \cline{2-3} \cline{4-5} \cline{6-7}
& Effect. \UTprox ($\uparrow$) & Effect. \DPU ($\downarrow$) & Effect. \UTprox ($\uparrow$) & Effect. \DPU ($\downarrow$) & Effect. \UTprox ($\uparrow$) & Effect.\DPU ($\downarrow$) \\ 
\midrule
\nameonetwo &  {6.2 (0.8)} & {10.4 (0.7)} & {20.3 (0.5)} & {8.2 (1.6)} & {8.0 (0.4)} & {10.1 (1.4)} \\
\nametwo & 5.1 (0.6) & {10.6 (0.7)} & 19.8 (1.0) & {7.8 (1.9)} & 7.7 (0.2) & {9.4 (1.8)}\\
\ipsvae & 2.3 (0.9) & 10.7 (0.9) & 16.3 (0.7) & 10.0 (1.9) & 5.6 (0.5) & {10.3 (0.4)} \\
\nikifair & 3.6 (0.4) & 10.8 (0.9) & 19.4 (1.0) & 10.2 (1.6) & 6.8 (0.5) & 11.0 (1.2) \\
\niki & 4.6 (0.7) & 14.6 (1.5) & {20.4 (0.5)} & 10.8 (1.9) & 7.6 (0.3) & 19.1 (2.3) \\
\bottomrule
\end{tabular}
 \end{table*}
}

\paragraph{Results.} 
{Table~\ref{tab:eff_util_fair_results} summarizes the results for several real-world datasets.
\name\ (I+II) consistently accumulates more utility and less unfairness during the online learning process compared to the other approaches.  Note that \ipsvae, which uses only labeled data in \phasetwo, accumulates less utility and more unfairness than \nametwo, which skips \phaseone\ but uses both labeled and unlabeled data in \phasetwo. This suggests that joint training on labeled and unlabeled data in \phasetwo\ significantly improves learning of both a fair representation and policy. Furthermore, \name~(I+II)  outperforms \nametwo, suggesting that using unlabeled data in \phaseone\ also improves the process. In summary, our results show that \emph{unlabeled data available at any learning stage should not be thrown away, but should be used to learn fair and profitable policies}.
Importantly, our method also outperforms the competing approach \nikifair\ both in terms of utility and fairness.
Moreover, even if compared to  the unfair approach \niki, \nameonetwo\ accumulates comparable or higher utility with significantly lower unfairness. This suggests that we \emph{we may not observe a trade-off between observed utility and fairness in the real world, assuming an unbiased ground truth exists.}
Note that without access to the ground truth label, we cannot comment on the performance with respect to the ground truth utility.}

\begin{figure*}[t]
	\centering
	\begin{subfigure}[t]{0.23\textwidth}
		\centering
		\includegraphics[width=3.12\textwidth]{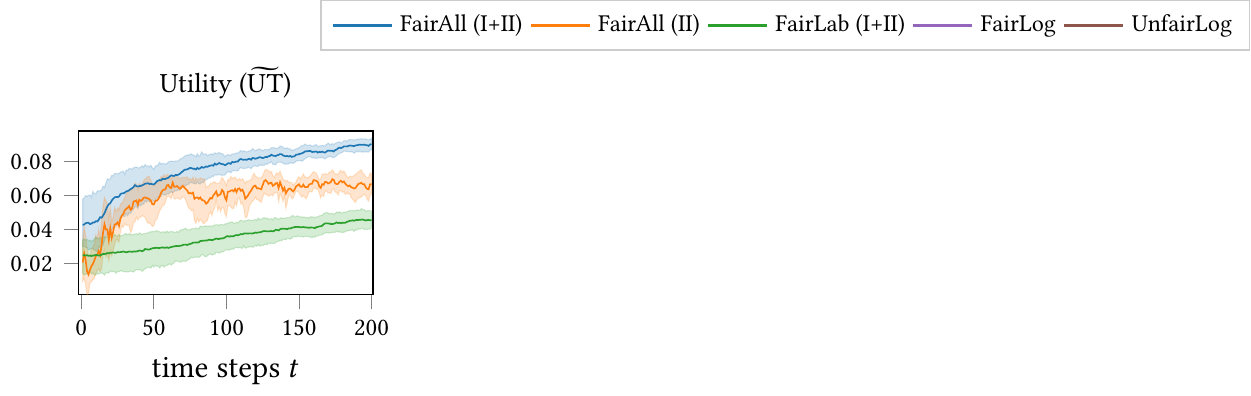}
	\end{subfigure}
	\begin{subfigure}[t]{0.23\textwidth}  
		\centering 
		\includegraphics[width=\textwidth]{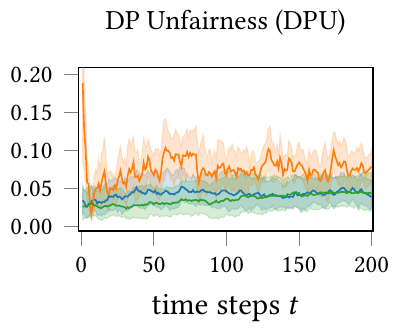}
	\end{subfigure}
	\begin{subfigure}[t]{0.23\textwidth}  
		\centering 
		\includegraphics[width=\textwidth]{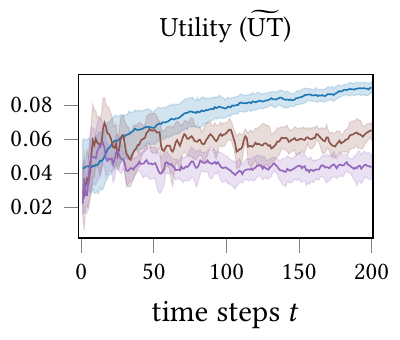}
	\end{subfigure}	
	\begin{subfigure}[t]{0.23\textwidth}   
		\centering 
		\includegraphics[width=\textwidth]{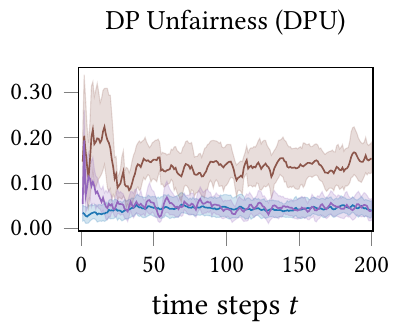}
	\end{subfigure}
	\caption{Proxy utility (\UTprox) and demographic parity unfairness (DPU) of policy models deployed on test-data for COMPAS dataset at each time-step. Left two plots show the benefit of using unlabeled data in both phases, comparing \name \ against \nametwo\ and \ipsvae. Right two plots show the better performance of \name\ compared to competing methods \nikifair, \niki.
	}
	 \Description{
	The figure shows four line plots to compare the performance of the different models in terms of proxy utility and demographic parity unfairness (DPU) on the real world dataset COMPAS. For visibility reasons - as lines would overlap - there are two plots of proxy utility and two plots of DPU. In the following we describe them together. For both proxy and DP utility we show five lines, each for one of the methods FairAll I+II, FairAll II, FairLab I+II, FairLog, and UnfairLog. There are no lines for the two optimal policies, because results are for a real world dataset. The plots show the benefit of using unlabeled data in both phases, comparing FairAll against FairAll (II), and FairLab (I+II) and also the better performance of FairAll compared to competing methods FairLog, and UnfairLog. All the plots have time steps from 0 to 200 as x axis. 
    Proxy utility is shown in plots with the utility measured with respect to the proxy label on the vertical y-axis, with values ranging from 0 to 0.1. 
    The proxy utility for our FairAll I+II starts at around 0.04 at time step 0 and increases until above 0.09 at time step 200. The variance decreases over time. In comparison, FairAll (II), and FairLab (I+II) both start at around 0.02 and also both increase over time. FairAll (II) reaches around 0.04 at time step 200 and the variance decreases over time. FairLab (I+II) reaches around 0.06 at time step 200 with higher variance both over seeds and over mean value. The competing methods FairLog, and UnfairLog both increase in utility in the first 25 time steps and then stay around the same value with comparably high variance in the mean value compared to our FairALL I+II.
    Demographic parity unfairness (DPU) is shown in plots with DPU on the vertical y-axis with values ranging from 0 to 0.2. 
    DPU for our FairAll I+II starts at around 0.03 at time step 0 without significant change of value until time step 200. The same holds for FairLab I+II. FairAll (II) starts with a high value of 0.2 DPU in time step 0 and then decreases with a high variance of the mean ranging between 0.05 and 0.1 and at time step 200 it has a value of around 0.06. For the competing methods FairLog has a higher DPU of up to 0.2 in the first time steps, and rapidly decreases to the same value as FairAll I+II, but with a higher variance of the mean. UnfairLog has a high DPU value of above 0.2 in the first time steps, from around time step 50 it stays around 0.13 with a high variance of the mean.
    }
	\label{fig:compas_test_results}
\end{figure*}

\subsection{How Do the Learned Policies Perform During Deployment?}
\label{sec:eval_real_test}

{In this section, we analyze how a given strategy, when applied to the population of interest, is expected to perform in the long run. To this end, we compare the performance of the resulting strategy at each time step using an i.i.d. test set.}

\paragraph{Result.}
{Figure~\ref{fig:compas_test_results} shows how utility \UTprox\ and unfairness \DPU\ evolve over time for \COMPAS.
Results on the other real-world datasets are in Appendix~E.
%
{\nametwo\ achieves both higher utility and higher unfairness compared to \ipsvae\ with significantly higher variance.}
{This suggests that unlabeled data in \phaseone\ benefit fairness, while unlabeled data in \phasetwo\ benefit utility.}
\name~(I+II) provides significantly higher utility and lower unfairness than the other learning models after approximately 50 time steps. 
Moreover, \name~(I+II) even provides higher utility than even the unfair reference model \niki\ while being as fair as \nikifair. 
 This empirically confirms the importance of unlabeled data \emph{in  both \phaseone\ and \phasetwo} to achieve high test utility and fairness in real-world scenarios.}

\subsubsection{Can We Reliably Stop Learning at Any Time?}
{{Assume that we want to stop the policy learning process from time $t_1$.} 
 {{Can we stop the learning process and deploy the policy at any time $t \geq t_1$?}}
We study this by measuring how much utility (fairness) of the output policies vary over time interval $[t_1, t]$ for a \emph{fixed test dataset}.
A low temporal variance (\TV) indicates stable behavior, such that it is safe to terminate the learning process at any time. However, when the variance is high, the decision-maker must carefully select the best stopping point. High \TV\
leads to unstable policies that may lead to low utility and/or high fairness.
In addition, we measure the temporal average $\mu$ of utility (unfairness). It is desirable for a learning process to have both low variance and high (low) $\mu$ for utility (unfairness).}

\paragraph{Results.} 
{In Table~\ref{tab:temporal_unfair}, we report $\tvdp,\mudp$ for unfairness and $\tvut,\muut$ for utility (see Def.~\ref{def:TV}). 
Compared to \ipsvae\ and \nametwo, \name~(I+II) provides similar \TV\ values. However, \name~(I+II) exhibits a better temporal average utility and fairness.
For \CREDIT, for example, note that \ipsvae\ has a lower $\tvdp$ and a higher average unfairness level $\mudp$ than \name~(I+II).
Compared to \nikifair, our method \nameonetwo\ is much more stable. It has lower $\tvdp$ and $\tvut$ as well as a higher average utility $\muut$ and lower unfairness $\mudp$.}

{ 
\renewcommand{\arraystretch}{1.2}
\begin{table*}[t]
\caption{Temporal variance and means of utility ($\tvut,\muut$) and demographic parity unfairness ($\tvdp,\mudp$). Metrics are measured on the time interval $t=[125, 200]$ on real-world datasets. We report the mean over ten runs with the standard deviation in brackets. For \TV, lower values are better. For $\mu$ higher (lower) is better for \UTprox\ (\DPU). Values multiplied by 100 for readability.}
 \Description{Results}
\label{tab:temporal_unfair}
\resizebox{\textwidth}{!}{%
\begin{tabular}{@{\extracolsep{2pt}}lcccccc}
\toprule
\multirow{2}{*}{Model} &
  \multicolumn{2}{c}{\COMPAS} &
  \multicolumn{2}{c}{\CREDIT} &
  \multicolumn{2}{c}{\MEPS} \\ 
  \cline{2-3} \cline{4-5} \cline{6-7}
 &
  $\tvdp$($\downarrow$) |
  $\mudp$ ($\downarrow$)&
  $\tvut$ ($\downarrow$) |
  $\muut$ ($\uparrow$) &
  $\tvdp$ ($\downarrow$) |
  $\mudp$ ($\downarrow$) &
  $\tvut$ ($\downarrow$) |
  $\muut$ ($\uparrow$) &
  $\tvdp$ ($\downarrow$) |
  $\mudp$ ($\downarrow$) &
  $\tvut$ ($\downarrow$) |
  $\muut$ ($\uparrow$) \\ 
  \midrule
  \nameonetwo &
  {1.0 (0.7)} |
  {4.3 (3.1)} &
  {0.4 (0.3)} |
  {8.6 (0.7)} &
  3.0 (2.0) |
  {5.0 (5.0)} &
  1.0 (0.7) |
  {18.8 (1.3)} &
  1.8 (1.6) |
  5.1 (3.1) &
  {0.2 (0.1)} |
  {7.9 (0.5)} \\ 
  \nametwo &
  2.8 (1.7) |
  7.8 (4.0) &
  0.8 (0.5) |
  6.6 (1.0) &
  2.7 (2.4) |
  {4.6 (3.0)} &
  0.7 (0.6) |
  18.2 (1.5) &
  3.8 (2.9) |
  6.4 (4.8) &
  0.3 (0.2) |
  7.4 (0.5) \\
\ipsvae &
  {0.7 (0.6)} |
  {4.3 (3.1)} &
  {0.3 (0.3)} |
  4.2 (0.9) &
  {1.7 (1.3)} |
  9.1 (5.7) &
  {0.5 (0.4)} |
  14.9 (1.4) &
  {0.7 (0.6)} |
  {4.0 (2.3)} &
  {0.3 (0.2)} |
  5.9 (0.7) \\
\nikifair &
  1.8 (1.4) |
  4.5 (3.8) &
  0.5 (0.4) |
  4.3 (1.1) &
  3.6 (2.0) |
  7.6 (4.4) &
  1.3 (1.1) |
  17.1 (2.3) &
  2.6 (2.8) |
  5.4 (3.3) &
  0.5 (0.4) |
  6.4 (0.7) \\
  \bottomrule
\end{tabular}%
}
\end{table*}
}

\section{Discussion of Assumptions and Outlook}\label{sec:discussion}


{In this section, we discuss the main assumptions, limitations, and potential consequences of our proposed framework.}

\paragraph{Assumptions on the Data Generation Process.}
{In this work, we assume that the true data generation process follows Figure~\ref{fig:generative} and that $S$ is a social construct like \textit{gender} or \textit{race}. This follows the understanding that discrimination is based on social constructs and not biological differences \cite{issa_sex_2020}. 
{We assume that each individual can be assigned to {a} social group within a social 
{construct}
(e.g., \textit{gender})}
%
and that $S$ is a root node in the graphical model. 
While being
common 
~\cite{Kusner2017-ln}, this is a debated modeling assumption~\cite{barocas19fairmlbook, issa_sex_2020}. 
Furthermore, we assume that
we 
observe a biased proxy variable $\Yprox$, 
{and} that
an {unobserved} unbiased ground truth $Y$ exists 
{that}
is independent of 
$S$. 
This means that there are no innate differences between social groups with respect to $\Y$. 
%
{However, see Appendix~F.1
for examples where $Y \dep S$. }
We 
{advise} any practitioner using our pipeline to carefully assess whether these assumptions hold in their case.}

\paragraph{Assumptions of Our Policy Learning Pipeline}
%
First, we assume access to a large unlabeled dataset of i.i.d. samples {from} the population of interest {for} our pipeline (Phase I). {For example, a university may have access to the grades of all students who graduated from high school in a given time period. We also assume access to sensitive information, which, in the real world, may conflict with privacy regulations (e.g., the principle of data minimization). In this paper, we show that access to a large unlabeled dataset of sensitive information not only increases the utility of the decision-maker but also fairness. We hope that this contributes to the debate between fair algorithmic decision-making and privacy.}

Second, we assume a decision-maker has an unlimited budget at each time step, i.e., it can take as many positive decisions as  desired. 
{A related line of work~\cite{kleinberg2018selection,khalili2021fair}, deals with fairness in selection problems, where candidates compete for a limited number of positions, or 
with pipeline settings~\cite{dwork2020individual}, where candidates enter the decision process one at a time.}
This is an interesting direction for future research.

Third, we assume that the underlying data distribution does not change over time and thus is not affected by the decisions. This assumption does not necessarily hold in the real world. While it exceeds the scope of this paper, it would be interesting to extend our pipeline to address \emph{distribution shift} {as a consequence of the decision-making process.}

{Lastly, in \phasetwo\ we learn a stochastic policy at each time step and use it to collect new data. We follow \citet{kilbertus2020fair} in their call for a general discussion about the ethics of non-deterministic decision-making.}

\paragraph{Assumptions on Fairness Metrics} 
{While the ethical evaluation of the applicability of a particular fairness notion in a specific context lies outside of our scope, we 
give an overview of when the use of our pipeline may be helpful in practice.}
{We evaluate fairness} based on the demographic parity (DP)
notion \cite{dwork2012fairness}.
\citet{heidari2019moral} map DP to Rawl's moral understanding, 
{according to which unequal treatment may only be justified on the basis of innate potential or ambition, not gender or race.}
Within this framework, the underlying assumption of DP is that individuals should receive utility solely based on the factors for which they are accountable.
In this paper, such factors are assumed to be captured by the unobserved ground truth label $\Y$. 
\citet{hertweck2021moral} show that one should enforce DP not only if socio-demographic groups have similar innate potential at birth, but in some cases even if unjust life biases lead to differences in realized abilities. \citet{wachter2020bias} similarly argue that DP and counterfactual fairness are bias transforming metrics {that} acknowledge historical inequalities and {assume that certain groups have} a worse starting point than others. 
{However, they 
also 
warn that, e.g., 
giving an individual a loan that they cannot repay can exacerbate inequalities. }

\section{Conclusion}
In this paper, we considered the problem of learning optimal fair decision policies in the presence of label bias and selection bias.
Prior work that attempted to solve the problem by learning stochastic {exploratory} decision policies in an online process {neglects} a large amount of unlabeled data and suffers from high variance in the learned policies over time.
In this work, we proposed a novel two-phase framework that leverages both labeled and unlabeled data to learn stable, fair decision policies.
We introduced a practical sequential decision-making pipeline \name\ that uses the latent representations of a variational autoencoder to learn, over time, the \emph{optimal fair policy according to the unobserved ground truth}. 
In line with our assumptions, {the} decision policies learned by \name\ satisfy {both the notion of} counterfactual fairness and demographic parity without {requiring} additional fairness constraints.
Through theoretical analysis and experiments with synthetic data, we validate that \name\ converges to the optimal (fair) policy with respect to the unobserved ground truth both in terms of utility and fairness.
Compared to the prior work, we show how our modeling approach helps us {to be} counterfactual and demographic parity fair.
{On real-world data,} we show how \name\ provides not only a significantly more effective learning method, but also higher utility, higher fairness, and a more stable learning process than the existing approach.
In comparison to baseline models, we also show the importance of {using} unlabeled data in both phases to achieve a more accurate, fair, and stable decision learning process.

\begin{acks}
{The authors would like to thank Adrián Javaloy Bornás for his guidance, discussion and for providing an initial codebase for working on VAEs with heterogenous data.
The authors also thank Pablo Sánchez-Martín, Batuhan Koyuncu for providing valuable feedback.
Special thanks goes to Diego Baptista Theuerkauf for help with formalizing proofs and notation.
Finally, the authors thank the anonymous reviewers for their detailed feedback and comments.}

{Miriam Rateike is supported by the German Federal Ministry of Education and Research (BMBF): Tübingen AI Center, FKZ: 01IS18039B.
Ayan Majumdar and Krishna P. Gummadi are supported by the ERC Advanced
Grant ``Foundations for Fair Social Computing'' (no. 789373).
Olga Mineeva is supported by the Max Planck ETH Center for Learning Systems.
The authors appreciate the generous funding support.}
\end{acks}

\nocite{singh2019understanding}
\nocite{pearl2016primer}
\nocite{aif360}

\bibliographystyle{ACM-Reference-Format}
\bibliography{references}


\begin{thebibliography}{60}


\ifx \showCODEN    \undefined \def \showCODEN     #1{\unskip}     \fi
\ifx \showDOI      \undefined \def \showDOI       #1{#1}\fi
\ifx \showISBNx    \undefined \def \showISBNx     #1{\unskip}     \fi
\ifx \showISBNxiii \undefined \def \showISBNxiii  #1{\unskip}     \fi
\ifx \showISSN     \undefined \def \showISSN      #1{\unskip}     \fi
\ifx \showLCCN     \undefined \def \showLCCN      #1{\unskip}     \fi
\ifx \shownote     \undefined \def \shownote      #1{#1}          \fi
\ifx \showarticletitle \undefined \def \showarticletitle #1{#1}   \fi
\ifx \showURL      \undefined \def \showURL       {\relax}        \fi
\providecommand\bibfield[2]{#2}
\providecommand\bibinfo[2]{#2}
\providecommand\natexlab[1]{#1}
\providecommand\showeprint[2][]{arXiv:#2}

\bibitem[Agarwal et~al\mbox{.}(2018)]%
        {agarwal2018reductions}
\bibfield{author}{\bibinfo{person}{Alekh Agarwal}, \bibinfo{person}{Alina
  Beygelzimer}, \bibinfo{person}{Miroslav Dudik}, \bibinfo{person}{John
  Langford}, {and} \bibinfo{person}{Hanna Wallach}.}
  \bibinfo{year}{2018}\natexlab{}.
\newblock \showarticletitle{A Reductions Approach to Fair Classification}
  \emph{(\bibinfo{series}{Proceedings of Machine Learning Research},
  Vol.~\bibinfo{volume}{80})}, \bibfield{editor}{\bibinfo{person}{Jennifer Dy}
  {and} \bibinfo{person}{Andreas Krause}} (Eds.). \bibinfo{publisher}{PMLR},
  \bibinfo{address}{Stockholmsmässan, Stockholm Sweden},
  \bibinfo{pages}{60--69}.
\newblock
\urldef\tempurl%
\url{http://proceedings.mlr.press/v80/agarwal18a.html}
\showURL{%
\tempurl}


\bibitem[Alemi et~al\mbox{.}(2017)]%
        {alemi2018information}
\bibfield{author}{\bibinfo{person}{Alexander~A. Alemi}, \bibinfo{person}{Ben
  Poole}, \bibinfo{person}{Ian Fischer}, \bibinfo{person}{Joshua~V. Dillon},
  \bibinfo{person}{Rif~A. Saurous}, {and} \bibinfo{person}{Kevin Murphy}.}
  \bibinfo{year}{2017}\natexlab{}.
\newblock \showarticletitle{An Information-Theoretic Analysis of Deep
  Latent-Variable Models}.
\newblock  (\bibinfo{year}{2017}).
\newblock
\showeprint[arXiv]{1711.00464}


\bibitem[Angwin et~al\mbox{.}(2016)]%
        {angwin2016machine}
\bibfield{author}{\bibinfo{person}{Julia Angwin}, \bibinfo{person}{Jeff
  Larson}, \bibinfo{person}{Surya Mattu}, {and} \bibinfo{person}{Lauren
  Kirchner}.} \bibinfo{year}{2016}\natexlab{}.
\newblock \showarticletitle{Machine bias: There's software used across the
  country to predict future criminals and it's biased against blacks}.
\newblock \bibinfo{journal}{\emph{ProPublica}}  \bibinfo{volume}{23}
  (\bibinfo{year}{2016}).
\newblock


\bibitem[Bahri et~al\mbox{.}(2022)]%
        {bahri2021scarf}
\bibfield{author}{\bibinfo{person}{Dara Bahri}, \bibinfo{person}{Heinrich
  Jiang}, \bibinfo{person}{Yi Tay}, {and} \bibinfo{person}{Donald Metzler}.}
  \bibinfo{year}{2022}\natexlab{}.
\newblock \showarticletitle{Scarf: Self-Supervised Contrastive Learning using
  Random Feature Corruption}. In \bibinfo{booktitle}{\emph{International
  Conference on Learning Representations}}.
\newblock
\urldef\tempurl%
\url{https://openreview.net/forum?id=CuV_qYkmKb3}
\showURL{%
\tempurl}


\bibitem[Balunovic et~al\mbox{.}(2021)]%
        {balunovic2021fair}
\bibfield{author}{\bibinfo{person}{Mislav Balunovic}, \bibinfo{person}{Anian
  Ruoss}, {and} \bibinfo{person}{Martin Vechev}.}
  \bibinfo{year}{2021}\natexlab{}.
\newblock \showarticletitle{Fair Normalizing Flows}. In
  \bibinfo{booktitle}{\emph{International Conference on Learning
  Representations}}.
\newblock


\bibitem[Barocas et~al\mbox{.}(2019)]%
        {barocas19fairmlbook}
\bibfield{author}{\bibinfo{person}{Solon Barocas}, \bibinfo{person}{Moritz
  Hardt}, {and} \bibinfo{person}{Arvind Narayanan}.}
  \bibinfo{year}{2019}\natexlab{}.
\newblock \bibinfo{booktitle}{\emph{Fairness and Machine Learning}}.
\newblock \bibinfo{publisher}{fairmlbook.org}.
\newblock
\newblock
\shownote{\url{http://www.fairmlbook.org}}.


\bibitem[Barocas and Selbst(2016)]%
        {barocas2016big}
\bibfield{author}{\bibinfo{person}{Solon Barocas} {and}
  \bibinfo{person}{Andrew~D Selbst}.} \bibinfo{year}{2016}\natexlab{}.
\newblock \showarticletitle{Big Data’s Disparate Impact}.
\newblock \bibinfo{journal}{\emph{California Law Review}}
  \bibinfo{volume}{104} (\bibinfo{year}{2016}), \bibinfo{pages}{671}.
\newblock


\bibitem[Bechavod et~al\mbox{.}(2019)]%
        {bechavod2019equal}
\bibfield{author}{\bibinfo{person}{Yahav Bechavod}, \bibinfo{person}{Katrina
  Ligett}, \bibinfo{person}{Aaron Roth}, \bibinfo{person}{Bo Waggoner}, {and}
  \bibinfo{person}{Steven~Z. Wu}.} \bibinfo{year}{2019}\natexlab{}.
\newblock \showarticletitle{Equal Opportunity in Online Classification with
  Partial Feedback}. In \bibinfo{booktitle}{\emph{Advances in Neural
  Information Processing Systems}},
  \bibfield{editor}{\bibinfo{person}{H.~Wallach},
  \bibinfo{person}{H.~Larochelle}, \bibinfo{person}{A.~Beygelzimer},
  \bibinfo{person}{F.~d\textquotesingle Alch\'{e}-Buc},
  \bibinfo{person}{E.~Fox}, {and} \bibinfo{person}{R.~Garnett}} (Eds.),
  Vol.~\bibinfo{volume}{32}. \bibinfo{publisher}{Curran Associates, Inc.}
\newblock
\urldef\tempurl%
\url{https://proceedings.neurips.cc/paper/2019/file/084afd913ab1e6ea58b8ca73f6cb41a6-Paper.pdf}
\showURL{%
\tempurl}


\bibitem[Bellamy et~al\mbox{.}(2018)]%
        {aif360}
\bibfield{author}{\bibinfo{person}{Rachel K.~E. Bellamy},
  \bibinfo{person}{Kuntal Dey}, \bibinfo{person}{Michael Hind},
  \bibinfo{person}{Samuel~C. Hoffman}, \bibinfo{person}{Stephanie Houde},
  \bibinfo{person}{Kalapriya Kannan}, \bibinfo{person}{Pranay Lohia},
  \bibinfo{person}{Jacquelyn Martino}, \bibinfo{person}{Sameep Mehta},
  \bibinfo{person}{Aleksandra Mojsilovic}, \bibinfo{person}{Seema Nagar},
  \bibinfo{person}{Karthikeyan~Natesan Ramamurthy}, \bibinfo{person}{John
  Richards}, \bibinfo{person}{Diptikalyan Saha}, \bibinfo{person}{Prasanna
  Sattigeri}, \bibinfo{person}{Moninder Singh}, \bibinfo{person}{Kush~R.
  Varshney}, {and} \bibinfo{person}{Yunfeng Zhang}.}
  \bibinfo{year}{2018}\natexlab{}.
\newblock \bibinfo{title}{{AI Fairness} 360: An Extensible Toolkit for
  Detecting, Understanding, and Mitigating Unwanted Algorithmic Bias}.
\newblock
\newblock
\urldef\tempurl%
\url{https://arxiv.org/abs/1810.01943}
\showURL{%
\tempurl}


\bibitem[Celis et~al\mbox{.}(2018)]%
        {celis2018algorithmic}
\bibfield{author}{\bibinfo{person}{L~Elisa Celis}, \bibinfo{person}{Sayash
  Kapoor}, \bibinfo{person}{Farnood Salehi}, {and} \bibinfo{person}{Nisheeth~K
  Vishnoi}.} \bibinfo{year}{2018}\natexlab{}.
\newblock \showarticletitle{An algorithmic framework to control bias in
  bandit-based personalization}.
\newblock \bibinfo{journal}{\emph{arXiv preprint arXiv:1802.08674}}
  (\bibinfo{year}{2018}).
\newblock


\bibitem[Cooper et~al\mbox{.}(2021)]%
        {cooper2021emergent}
\bibfield{author}{\bibinfo{person}{A.~Feder Cooper}, \bibinfo{person}{Ellen
  Abrams}, {and} \bibinfo{person}{NA NA}.} \bibinfo{year}{2021}\natexlab{}.
\newblock \showarticletitle{Emergent Unfairness in Algorithmic
  Fairness-Accuracy Trade-Off Research}. In
  \bibinfo{booktitle}{\emph{Proceedings of the 2021 AAAI/ACM Conference on AI,
  Ethics, and Society}}. \bibinfo{publisher}{Association for Computing
  Machinery}, \bibinfo{address}{New York, NY, USA}, \bibinfo{pages}{46–54}.
\newblock
\showISBNx{9781450384735}
\urldef\tempurl%
\url{https://doi.org/10.1145/3461702.3462519}
\showURL{%
\tempurl}


\bibitem[Corbett-Davies et~al\mbox{.}(2017)]%
        {corbett2017algorithmic}
\bibfield{author}{\bibinfo{person}{Sam Corbett-Davies}, \bibinfo{person}{Emma
  Pierson}, \bibinfo{person}{Avi Feller}, \bibinfo{person}{Sharad Goel}, {and}
  \bibinfo{person}{Aziz Huq}.} \bibinfo{year}{2017}\natexlab{}.
\newblock \showarticletitle{Algorithmic Decision Making and the Cost of
  Fairness}. In \bibinfo{booktitle}{\emph{Proceedings of the 23rd ACM SIGKDD
  International Conference on Knowledge Discovery and Data Mining}} (Halifax,
  NS, Canada) \emph{(\bibinfo{series}{KDD '17})}.
  \bibinfo{publisher}{Association for Computing Machinery},
  \bibinfo{address}{New York, NY, USA}, \bibinfo{pages}{797–806}.
\newblock
\showISBNx{9781450348874}
\urldef\tempurl%
\url{https://doi.org/10.1145/3097983.3098095}
\showDOI{\tempurl}


\bibitem[Cotter et~al\mbox{.}(2019)]%
        {cotter2019two}
\bibfield{author}{\bibinfo{person}{Andrew Cotter}, \bibinfo{person}{Heinrich
  Jiang}, {and} \bibinfo{person}{Karthik Sridharan}.}
  \bibinfo{year}{2019}\natexlab{}.
\newblock \showarticletitle{Two-Player Games for Efficient Non-Convex
  Constrained Optimization}. In \bibinfo{booktitle}{\emph{Proceedings of the
  30th International Conference on Algorithmic Learning Theory}}
  \emph{(\bibinfo{series}{Proceedings of Machine Learning Research},
  Vol.~\bibinfo{volume}{98})}, \bibfield{editor}{\bibinfo{person}{Aurélien
  Garivier} {and} \bibinfo{person}{Satyen Kale}} (Eds.).
  \bibinfo{publisher}{PMLR}, \bibinfo{pages}{300--332}.
\newblock
\urldef\tempurl%
\url{https://proceedings.mlr.press/v98/cotter19a.html}
\showURL{%
\tempurl}


\bibitem[Creager et~al\mbox{.}(2019)]%
        {creager2019flexibly}
\bibfield{author}{\bibinfo{person}{Elliot Creager}, \bibinfo{person}{David
  Madras}, \bibinfo{person}{Joern-Henrik Jacobsen}, \bibinfo{person}{Marissa
  Weis}, \bibinfo{person}{Kevin Swersky}, \bibinfo{person}{Toniann Pitassi},
  {and} \bibinfo{person}{Richard Zemel}.} \bibinfo{year}{2019}\natexlab{}.
\newblock \showarticletitle{Flexibly Fair Representation Learning by
  Disentanglement}. In \bibinfo{booktitle}{\emph{Proceedings of the 36th
  International Conference on Machine Learning}}
  \emph{(\bibinfo{series}{Proceedings of Machine Learning Research},
  Vol.~\bibinfo{volume}{97})}, \bibfield{editor}{\bibinfo{person}{Kamalika
  Chaudhuri} {and} \bibinfo{person}{Ruslan Salakhutdinov}} (Eds.).
  \bibinfo{publisher}{PMLR}, \bibinfo{pages}{1436--1445}.
\newblock
\urldef\tempurl%
\url{https://proceedings.mlr.press/v97/creager19a.html}
\showURL{%
\tempurl}


\bibitem[Dua and Graff(2017)]%
        {dua2019}
\bibfield{author}{\bibinfo{person}{Dheeru Dua} {and} \bibinfo{person}{Casey
  Graff}.} \bibinfo{year}{2017}\natexlab{}.
\newblock \bibinfo{title}{{UCI} Machine Learning Repository}.
\newblock
\newblock
\urldef\tempurl%
\url{http://archive.ics.uci.edu/ml}
\showURL{%
\tempurl}


\bibitem[Dutta et~al\mbox{.}(2020)]%
        {dutta2020there}
\bibfield{author}{\bibinfo{person}{Sanghamitra Dutta}, \bibinfo{person}{Dennis
  Wei}, \bibinfo{person}{Hazar Yueksel}, \bibinfo{person}{Pin-Yu Chen},
  \bibinfo{person}{Sijia Liu}, {and} \bibinfo{person}{Kush Varshney}.}
  \bibinfo{year}{2020}\natexlab{}.
\newblock \showarticletitle{Is There a Trade-Off Between Fairness and Accuracy?
  {A} Perspective Using Mismatched Hypothesis Testing}. In
  \bibinfo{booktitle}{\emph{Proceedings of the 37th International Conference on
  Machine Learning}} \emph{(\bibinfo{series}{Proceedings of Machine Learning
  Research}, Vol.~\bibinfo{volume}{119})},
  \bibfield{editor}{\bibinfo{person}{Hal~Daumé III} {and}
  \bibinfo{person}{Aarti Singh}} (Eds.). \bibinfo{publisher}{PMLR},
  \bibinfo{pages}{2803--2813}.
\newblock
\urldef\tempurl%
\url{https://proceedings.mlr.press/v119/dutta20a.html}
\showURL{%
\tempurl}


\bibitem[Dwork et~al\mbox{.}(2012)]%
        {dwork2012fairness}
\bibfield{author}{\bibinfo{person}{Cynthia Dwork}, \bibinfo{person}{Moritz
  Hardt}, \bibinfo{person}{Toniann Pitassi}, \bibinfo{person}{Omer Reingold},
  {and} \bibinfo{person}{Richard Zemel}.} \bibinfo{year}{2012}\natexlab{}.
\newblock \showarticletitle{Fairness through Awareness}. In
  \bibinfo{booktitle}{\emph{Proceedings of the 3rd Innovations in Theoretical
  Computer Science Conference}} (Cambridge, Massachusetts)
  \emph{(\bibinfo{series}{ITCS '12})}. \bibinfo{publisher}{Association for
  Computing Machinery}, \bibinfo{address}{New York, NY, USA},
  \bibinfo{pages}{214–226}.
\newblock
\showISBNx{9781450311151}
\urldef\tempurl%
\url{https://doi.org/10.1145/2090236.2090255}
\showDOI{\tempurl}


\bibitem[Dwork et~al\mbox{.}(2020)]%
        {dwork2020individual}
\bibfield{author}{\bibinfo{person}{Cynthia Dwork}, \bibinfo{person}{Christina
  Ilvento}, {and} \bibinfo{person}{Meena Jagadeesan}.}
  \bibinfo{year}{2020}\natexlab{}.
\newblock \showarticletitle{Individual Fairness in Pipelines}. In
  \bibinfo{booktitle}{\emph{1st Symposium on Foundations of Responsible
  Computing (FORC 2020)}}. Schloss Dagstuhl-Leibniz-Zentrum f{\"u}r Informatik.
\newblock


\bibitem[Friedler et~al\mbox{.}(2016)]%
        {friedler2016possibility}
\bibfield{author}{\bibinfo{person}{Sorelle~A Friedler}, \bibinfo{person}{Carlos
  Scheidegger}, {and} \bibinfo{person}{Suresh Venkatasubramanian}.}
  \bibinfo{year}{2016}\natexlab{}.
\newblock \showarticletitle{On the (im) possibility of fairness}.
\newblock \bibinfo{journal}{\emph{arXiv preprint arXiv:1609.07236}}
  (\bibinfo{year}{2016}).
\newblock


\bibitem[Goodfellow et~al\mbox{.}(2014)]%
        {goodfellow2014generative}
\bibfield{author}{\bibinfo{person}{Ian Goodfellow}, \bibinfo{person}{Jean
  Pouget-Abadie}, \bibinfo{person}{Mehdi Mirza}, \bibinfo{person}{Bing Xu},
  \bibinfo{person}{David Warde-Farley}, \bibinfo{person}{Sherjil Ozair},
  \bibinfo{person}{Aaron Courville}, {and} \bibinfo{person}{Yoshua Bengio}.}
  \bibinfo{year}{2014}\natexlab{}.
\newblock \showarticletitle{Generative Adversarial Nets}. In
  \bibinfo{booktitle}{\emph{Advances in Neural Information Processing
  Systems}}, \bibfield{editor}{\bibinfo{person}{Z.~Ghahramani},
  \bibinfo{person}{M.~Welling}, \bibinfo{person}{C.~Cortes},
  \bibinfo{person}{N.~Lawrence}, {and} \bibinfo{person}{K.Q. Weinberger}}
  (Eds.), Vol.~\bibinfo{volume}{27}. \bibinfo{publisher}{Curran Associates,
  Inc.}
\newblock
\urldef\tempurl%
\url{https://proceedings.neurips.cc/paper/2014/file/5ca3e9b122f61f8f06494c97b1afccf3-Paper.pdf}
\showURL{%
\tempurl}


\bibitem[Grari et~al\mbox{.}(2020)]%
        {grari2020adversarial}
\bibfield{author}{\bibinfo{person}{Vincent Grari}, \bibinfo{person}{Sylvain
  Lamprier}, {and} \bibinfo{person}{Marcin Detyniecki}.}
  \bibinfo{year}{2020}\natexlab{}.
\newblock \showarticletitle{Adversarial learning for counterfactual fairness}.
\newblock \bibinfo{journal}{\emph{arXiv preprint arXiv:2008.13122}}
  (\bibinfo{year}{2020}).
\newblock


\bibitem[Gupta and Kamble(2021)]%
        {gupta2018temporal}
\bibfield{author}{\bibinfo{person}{Swati Gupta} {and} \bibinfo{person}{Vijay
  Kamble}.} \bibinfo{year}{2021}\natexlab{}.
\newblock \showarticletitle{Individual Fairness in Hindsight}.
\newblock \bibinfo{journal}{\emph{Journal of Machine Learning Research}}
  \bibinfo{volume}{22}, \bibinfo{number}{144} (\bibinfo{year}{2021}),
  \bibinfo{pages}{1--35}.
\newblock
\urldef\tempurl%
\url{http://jmlr.org/papers/v22/19-658.html}
\showURL{%
\tempurl}


\bibitem[Hardt et~al\mbox{.}(2016)]%
        {hardt2016equality}
\bibfield{author}{\bibinfo{person}{Moritz Hardt}, \bibinfo{person}{Eric Price},
  \bibinfo{person}{Eric Price}, {and} \bibinfo{person}{Nati Srebro}.}
  \bibinfo{year}{2016}\natexlab{}.
\newblock \showarticletitle{Equality of Opportunity in Supervised Learning}. In
  \bibinfo{booktitle}{\emph{Advances in Neural Information Processing
  Systems}}, \bibfield{editor}{\bibinfo{person}{D.~Lee},
  \bibinfo{person}{M.~Sugiyama}, \bibinfo{person}{U.~Luxburg},
  \bibinfo{person}{I.~Guyon}, {and} \bibinfo{person}{R.~Garnett}} (Eds.),
  Vol.~\bibinfo{volume}{29}. \bibinfo{publisher}{Curran Associates, Inc.}
\newblock
\urldef\tempurl%
\url{https://proceedings.neurips.cc/paper/2016/file/9d2682367c3935defcb1f9e247a97c0d-Paper.pdf}
\showURL{%
\tempurl}


\bibitem[Healthcare Research \&~Quality(2018)]%
        {meps2018}
\bibfield{author}{\bibinfo{person}{Agency~for Healthcare Research \&~Quality}.}
  \bibinfo{year}{2018}\natexlab{}.
\newblock \bibinfo{title}{Medical Expenditure Panel Survey (MEPS)}.
\newblock
\newblock
\urldef\tempurl%
\url{https://www.ahrq.gov/data/meps.html}
\showURL{%
\tempurl}


\bibitem[Heidari et~al\mbox{.}(2019)]%
        {heidari2019moral}
\bibfield{author}{\bibinfo{person}{Hoda Heidari}, \bibinfo{person}{Michele
  Loi}, \bibinfo{person}{Krishna~P. Gummadi}, {and} \bibinfo{person}{Andreas
  Krause}.} \bibinfo{year}{2019}\natexlab{}.
\newblock \showarticletitle{A Moral Framework for Understanding Fair ML through
  Economic Models of Equality of Opportunity}. In
  \bibinfo{booktitle}{\emph{Proceedings of the Conference on Fairness,
  Accountability, and Transparency}} (Atlanta, GA, USA)
  \emph{(\bibinfo{series}{FAT* '19})}. \bibinfo{publisher}{Association for
  Computing Machinery}, \bibinfo{address}{New York, NY, USA},
  \bibinfo{pages}{181–190}.
\newblock
\showISBNx{9781450361255}
\urldef\tempurl%
\url{https://doi.org/10.1145/3287560.3287584}
\showDOI{\tempurl}


\bibitem[Hertweck et~al\mbox{.}(2021)]%
        {hertweck2021moral}
\bibfield{author}{\bibinfo{person}{Corinna Hertweck},
  \bibinfo{person}{Christoph Heitz}, {and} \bibinfo{person}{Michele Loi}.}
  \bibinfo{year}{2021}\natexlab{}.
\newblock \showarticletitle{On the Moral Justification of Statistical Parity}.
  In \bibinfo{booktitle}{\emph{Proceedings of the 2021 ACM Conference on
  Fairness, Accountability, and Transparency}} (Virtual Event, Canada)
  \emph{(\bibinfo{series}{FAccT '21})}. \bibinfo{publisher}{Association for
  Computing Machinery}, \bibinfo{address}{New York, NY, USA},
  \bibinfo{pages}{747–757}.
\newblock
\showISBNx{9781450383097}
\urldef\tempurl%
\url{https://doi.org/10.1145/3442188.3445936}
\showDOI{\tempurl}


\bibitem[Horvitz and Thompson(1952)]%
        {horvitz1952IPS}
\bibfield{author}{\bibinfo{person}{Daniel~G Horvitz} {and}
  \bibinfo{person}{Donovan~J Thompson}.} \bibinfo{year}{1952}\natexlab{}.
\newblock \showarticletitle{A generalization of sampling without replacement
  from a finite universe}.
\newblock \bibinfo{journal}{\emph{Journal of the American statistical
  Association}} \bibinfo{volume}{47}, \bibinfo{number}{260}
  (\bibinfo{year}{1952}), \bibinfo{pages}{663--685}.
\newblock


\bibitem[Hu and Kohler-Hausmann(2020)]%
        {issa_sex_2020}
\bibfield{author}{\bibinfo{person}{Lily Hu} {and} \bibinfo{person}{Issa
  Kohler-Hausmann}.} \bibinfo{year}{2020}\natexlab{}.
\newblock \showarticletitle{What's Sex Got to Do with Machine Learning?}. In
  \bibinfo{booktitle}{\emph{Proceedings of the 2020 Conference on Fairness,
  Accountability, and Transparency}} (Barcelona, Spain)
  \emph{(\bibinfo{series}{FAT* '20})}. \bibinfo{publisher}{Association for
  Computing Machinery}, \bibinfo{address}{New York, NY, USA},
  \bibinfo{pages}{513}.
\newblock
\showISBNx{9781450369367}
\urldef\tempurl%
\url{https://doi.org/10.1145/3351095.3375674}
\showDOI{\tempurl}


\bibitem[Khalili et~al\mbox{.}(2021)]%
        {khalili2021fair}
\bibfield{author}{\bibinfo{person}{Mohammad~Mahdi Khalili},
  \bibinfo{person}{Xueru Zhang}, {and} \bibinfo{person}{Mahed Abroshan}.}
  \bibinfo{year}{2021}\natexlab{}.
\newblock \showarticletitle{Fair Sequential Selection Using Supervised Learning
  Models}. In \bibinfo{booktitle}{\emph{Advances in Neural Information
  Processing Systems}}, \bibfield{editor}{\bibinfo{person}{M.~Ranzato},
  \bibinfo{person}{A.~Beygelzimer}, \bibinfo{person}{Y.~Dauphin},
  \bibinfo{person}{P.S. Liang}, {and} \bibinfo{person}{J.~Wortman Vaughan}}
  (Eds.), Vol.~\bibinfo{volume}{34}. \bibinfo{publisher}{Curran Associates,
  Inc.}, \bibinfo{pages}{28144--28155}.
\newblock
\urldef\tempurl%
\url{https://proceedings.neurips.cc/paper/2021/file/ed277964a8959e72a0d987e598dfbe72-Paper.pdf}
\showURL{%
\tempurl}


\bibitem[Khemakhem et~al\mbox{.}(2021)]%
        {khemakhem2021causal}
\bibfield{author}{\bibinfo{person}{Ilyes Khemakhem}, \bibinfo{person}{Ricardo
  Monti}, \bibinfo{person}{Robert Leech}, {and} \bibinfo{person}{Aapo
  Hyvarinen}.} \bibinfo{year}{2021}\natexlab{}.
\newblock \showarticletitle{Causal Autoregressive Flows}. In
  \bibinfo{booktitle}{\emph{Proceedings of The 24th International Conference on
  Artificial Intelligence and Statistics}} \emph{(\bibinfo{series}{Proceedings
  of Machine Learning Research}, Vol.~\bibinfo{volume}{130})},
  \bibfield{editor}{\bibinfo{person}{Arindam Banerjee} {and}
  \bibinfo{person}{Kenji Fukumizu}} (Eds.). \bibinfo{publisher}{PMLR},
  \bibinfo{pages}{3520--3528}.
\newblock
\urldef\tempurl%
\url{https://proceedings.mlr.press/v130/khemakhem21a.html}
\showURL{%
\tempurl}


\bibitem[Kilbertus et~al\mbox{.}(2020)]%
        {kilbertus2020fair}
\bibfield{author}{\bibinfo{person}{Niki Kilbertus},
  \bibinfo{person}{Manuel~Gomez Rodriguez}, \bibinfo{person}{Bernhard
  Sch\"olkopf}, \bibinfo{person}{Krikamol Muandet}, {and}
  \bibinfo{person}{Isabel Valera}.} \bibinfo{year}{2020}\natexlab{}.
\newblock \showarticletitle{Fair Decisions Despite Imperfect Predictions}. In
  \bibinfo{booktitle}{\emph{Proceedings of the Twenty Third International
  Conference on Artificial Intelligence and Statistics}}
  \emph{(\bibinfo{series}{Proceedings of Machine Learning Research},
  Vol.~\bibinfo{volume}{108})}, \bibfield{editor}{\bibinfo{person}{Silvia
  Chiappa} {and} \bibinfo{person}{Roberto Calandra}} (Eds.).
  \bibinfo{publisher}{PMLR}, \bibinfo{pages}{277--287}.
\newblock
\urldef\tempurl%
\url{https://proceedings.mlr.press/v108/kilbertus20a.html}
\showURL{%
\tempurl}


\bibitem[Kingma et~al\mbox{.}(2014)]%
        {kingma2014semi}
\bibfield{author}{\bibinfo{person}{Durk~P Kingma}, \bibinfo{person}{Shakir
  Mohamed}, \bibinfo{person}{Danilo Jimenez~Rezende}, {and}
  \bibinfo{person}{Max Welling}.} \bibinfo{year}{2014}\natexlab{}.
\newblock \showarticletitle{Semi-supervised Learning with Deep Generative
  Models}. In \bibinfo{booktitle}{\emph{Advances in Neural Information
  Processing Systems}}, \bibfield{editor}{\bibinfo{person}{Z.~Ghahramani},
  \bibinfo{person}{M.~Welling}, \bibinfo{person}{C.~Cortes},
  \bibinfo{person}{N.~Lawrence}, {and} \bibinfo{person}{K.Q. Weinberger}}
  (Eds.), Vol.~\bibinfo{volume}{27}. \bibinfo{publisher}{Curran Associates,
  Inc.}
\newblock
\urldef\tempurl%
\url{https://proceedings.neurips.cc/paper/2014/file/d523773c6b194f37b938d340d5d02232-Paper.pdf}
\showURL{%
\tempurl}


\bibitem[Kingma and Welling(2014)]%
        {kingma2013auto}
\bibfield{author}{\bibinfo{person}{Diederik~P Kingma} {and}
  \bibinfo{person}{Max Welling}.} \bibinfo{year}{2014}\natexlab{}.
\newblock \showarticletitle{Auto-encoding Variational Bayes}. In
  \bibinfo{booktitle}{\emph{2nd International Conference on Learning
  Representations}}, \bibfield{editor}{\bibinfo{person}{Yoshua Bengio} {and}
  \bibinfo{person}{Yann LeCun}} (Eds.), Vol.~\bibinfo{volume}{2}.
\newblock


\bibitem[Kleinberg and Raghavan(2018)]%
        {kleinberg2018selection}
\bibfield{author}{\bibinfo{person}{Jon Kleinberg} {and} \bibinfo{person}{Manish
  Raghavan}.} \bibinfo{year}{2018}\natexlab{}.
\newblock \showarticletitle{Selection Problems in the Presence of Implicit
  Bias}. In \bibinfo{booktitle}{\emph{9th Innovations in Theoretical Computer
  Science Conference (ITCS 2018)}}. Schloss Dagstuhl-Leibniz-Zentrum fuer
  Informatik.
\newblock


\bibitem[Kusner et~al\mbox{.}(2017)]%
        {Kusner2017-ln}
\bibfield{author}{\bibinfo{person}{Matt~J Kusner}, \bibinfo{person}{Joshua
  Loftus}, \bibinfo{person}{Chris Russell}, {and} \bibinfo{person}{Ricardo
  Silva}.} \bibinfo{year}{2017}\natexlab{}.
\newblock \showarticletitle{Counterfactual Fairness}. In
  \bibinfo{booktitle}{\emph{Advances in Neural Information Processing
  Systems}}, \bibfield{editor}{\bibinfo{person}{I.~Guyon},
  \bibinfo{person}{U.~Von Luxburg}, \bibinfo{person}{S.~Bengio},
  \bibinfo{person}{H.~Wallach}, \bibinfo{person}{R.~Fergus},
  \bibinfo{person}{S.~Vishwanathan}, {and} \bibinfo{person}{R.~Garnett}}
  (Eds.), Vol.~\bibinfo{volume}{30}. \bibinfo{publisher}{Curran Associates,
  Inc.}
\newblock
\urldef\tempurl%
\url{https://proceedings.neurips.cc/paper/2017/file/a486cd07e4ac3d270571622f4f316ec5-Paper.pdf}
\showURL{%
\tempurl}


\bibitem[Lakkaraju et~al\mbox{.}(2017)]%
        {lakkaraju2017selective}
\bibfield{author}{\bibinfo{person}{Himabindu Lakkaraju}, \bibinfo{person}{Jon
  Kleinberg}, \bibinfo{person}{Jure Leskovec}, \bibinfo{person}{Jens Ludwig},
  {and} \bibinfo{person}{Sendhil Mullainathan}.}
  \bibinfo{year}{2017}\natexlab{}.
\newblock \showarticletitle{The Selective Labels Problem: Evaluating
  Algorithmic Predictions in the Presence of Unobservables}. In
  \bibinfo{booktitle}{\emph{Proceedings of the 23rd ACM SIGKDD International
  Conference on Knowledge Discovery and Data Mining}} (Halifax, NS, Canada)
  \emph{(\bibinfo{series}{KDD '17})}. \bibinfo{publisher}{Association for
  Computing Machinery}, \bibinfo{address}{New York, NY, USA},
  \bibinfo{pages}{275–284}.
\newblock
\showISBNx{9781450348874}
\urldef\tempurl%
\url{https://doi.org/10.1145/3097983.3098066}
\showDOI{\tempurl}


\bibitem[Larson et~al\mbox{.}(2016)]%
        {larson2016how}
\bibfield{author}{\bibinfo{person}{Jeff Larson}, \bibinfo{person}{Surya Mattu},
  \bibinfo{person}{Lauren Kirchner}, {and} \bibinfo{person}{Julia Angwin}.}
  \bibinfo{year}{2016}\natexlab{}.
\newblock \bibinfo{title}{How We Analyzed the COMPAS Recidivism Algorithm}.
\newblock
\newblock
\urldef\tempurl%
\url{https://www.propublica.org/article/how-we-analyzed-the-compas-recidivism-algorithm}
\showURL{%
\tempurl}


\bibitem[Louizos et~al\mbox{.}(2016)]%
        {louizos2015variational}
\bibfield{author}{\bibinfo{person}{Christos Louizos}, \bibinfo{person}{Kevin
  Swersky}, \bibinfo{person}{Yujia Li}, \bibinfo{person}{Max Welling}, {and}
  \bibinfo{person}{Richard~S. Zemel}.} \bibinfo{year}{2016}\natexlab{}.
\newblock \showarticletitle{The Variational Fair Autoencoder}. In
  \bibinfo{booktitle}{\emph{4th International Conference on Learning
  Representations, {ICLR} 2016, San Juan, Puerto Rico, May 2-4, 2016,
  Conference Track Proceedings}}, \bibfield{editor}{\bibinfo{person}{Yoshua
  Bengio} {and} \bibinfo{person}{Yann LeCun}} (Eds.).
\newblock
\urldef\tempurl%
\url{http://arxiv.org/abs/1511.00830}
\showURL{%
\tempurl}


\bibitem[Madras et~al\mbox{.}(2018)]%
        {madras2018adversarially-learning}
\bibfield{author}{\bibinfo{person}{David Madras}, \bibinfo{person}{Elliot
  Creager}, \bibinfo{person}{Toniann Pitassi}, {and} \bibinfo{person}{Richard
  Zemel}.} \bibinfo{year}{2018}\natexlab{}.
\newblock \showarticletitle{Learning Adversarially Fair and Transferable
  Representations}. In \bibinfo{booktitle}{\emph{Proceedings of the 35th
  International Conference on Machine Learning}}
  \emph{(\bibinfo{series}{Proceedings of Machine Learning Research},
  Vol.~\bibinfo{volume}{80})}, \bibfield{editor}{\bibinfo{person}{Jennifer Dy}
  {and} \bibinfo{person}{Andreas Krause}} (Eds.). \bibinfo{publisher}{PMLR},
  \bibinfo{pages}{3384--3393}.
\newblock
\urldef\tempurl%
\url{https://proceedings.mlr.press/v80/madras18a.html}
\showURL{%
\tempurl}


\bibitem[Moyer et~al\mbox{.}(2018)]%
        {moyer2018invariant}
\bibfield{author}{\bibinfo{person}{Daniel Moyer}, \bibinfo{person}{Shuyang
  Gao}, \bibinfo{person}{Rob Brekelmans}, \bibinfo{person}{Aram Galstyan},
  {and} \bibinfo{person}{Greg Ver~Steeg}.} \bibinfo{year}{2018}\natexlab{}.
\newblock \showarticletitle{Invariant Representations without Adversarial
  Training}. In \bibinfo{booktitle}{\emph{Advances in Neural Information
  Processing Systems}}, \bibfield{editor}{\bibinfo{person}{S.~Bengio},
  \bibinfo{person}{H.~Wallach}, \bibinfo{person}{H.~Larochelle},
  \bibinfo{person}{K.~Grauman}, \bibinfo{person}{N.~Cesa-Bianchi}, {and}
  \bibinfo{person}{R.~Garnett}} (Eds.), Vol.~\bibinfo{volume}{31}.
  \bibinfo{publisher}{Curran Associates, Inc.}
\newblock
\urldef\tempurl%
\url{https://proceedings.neurips.cc/paper/2018/file/415185ea244ea2b2bedeb0449b926802-Paper.pdf}
\showURL{%
\tempurl}


\bibitem[Park et~al\mbox{.}(2022)]%
        {park2022fair}
\bibfield{author}{\bibinfo{person}{Sungho Park}, \bibinfo{person}{Jewook Lee},
  \bibinfo{person}{Pilhyeon Lee}, \bibinfo{person}{Sunhee Hwang},
  \bibinfo{person}{Dohyung Kim}, {and} \bibinfo{person}{Hyeran Byun}.}
  \bibinfo{year}{2022}\natexlab{}.
\newblock \showarticletitle{Fair Contrastive Learning for Facial Attribute
  Classification}.
\newblock \bibinfo{journal}{\emph{arXiv preprint arXiv:2203.16209}}
  (\bibinfo{year}{2022}).
\newblock


\bibitem[Pearl et~al\mbox{.}(2016)]%
        {pearl2016primer}
\bibfield{author}{\bibinfo{person}{Judea Pearl}, \bibinfo{person}{Madelyn
  Glymour}, {and} \bibinfo{person}{Nicholas~P Jewell}.}
  \bibinfo{year}{2016}\natexlab{}.
\newblock \bibinfo{booktitle}{\emph{Causal inference in statistics: A primer}}.
\newblock \bibinfo{publisher}{John Wiley \& Sons}.
\newblock


\bibitem[Razavian et~al\mbox{.}(2014)]%
        {sharif2014cnn}
\bibfield{author}{\bibinfo{person}{Ali~Sharif Razavian},
  \bibinfo{person}{Hossein Azizpour}, \bibinfo{person}{Josephine Sullivan},
  {and} \bibinfo{person}{Stefan Carlsson}.} \bibinfo{year}{2014}\natexlab{}.
\newblock \showarticletitle{CNN Features Off-the-Shelf: An Astounding Baseline
  for Recognition}. In \bibinfo{booktitle}{\emph{2014 IEEE Conference on
  Computer Vision and Pattern Recognition Workshops}}.
  \bibinfo{pages}{512--519}.
\newblock
\urldef\tempurl%
\url{https://doi.org/10.1109/CVPRW.2014.131}
\showDOI{\tempurl}


\bibitem[Rezende and Mohamed(2015)]%
        {rezende2015variational}
\bibfield{author}{\bibinfo{person}{Danilo Rezende} {and}
  \bibinfo{person}{Shakir Mohamed}.} \bibinfo{year}{2015}\natexlab{}.
\newblock \showarticletitle{Variational Inference with Normalizing Flows}. In
  \bibinfo{booktitle}{\emph{Proceedings of the 32nd International Conference on
  Machine Learning}} \emph{(\bibinfo{series}{Proceedings of Machine Learning
  Research}, Vol.~\bibinfo{volume}{37})},
  \bibfield{editor}{\bibinfo{person}{Francis Bach} {and} \bibinfo{person}{David
  Blei}} (Eds.). \bibinfo{publisher}{PMLR}, \bibinfo{address}{Lille, France},
  \bibinfo{pages}{1530--1538}.
\newblock
\urldef\tempurl%
\url{https://proceedings.mlr.press/v37/rezende15.html}
\showURL{%
\tempurl}


\bibitem[Rezende et~al\mbox{.}(2014)]%
        {rezende2014stochastic}
\bibfield{author}{\bibinfo{person}{Danilo~Jimenez Rezende},
  \bibinfo{person}{Shakir Mohamed}, {and} \bibinfo{person}{Daan Wierstra}.}
  \bibinfo{year}{2014}\natexlab{}.
\newblock \showarticletitle{Stochastic Backpropagation and Approximate
  Inference in Deep Generative Models}. In
  \bibinfo{booktitle}{\emph{Proceedings of the 31st International Conference on
  Machine Learning}} \emph{(\bibinfo{series}{Proceedings of Machine Learning
  Research}, Vol.~\bibinfo{volume}{32})},
  \bibfield{editor}{\bibinfo{person}{Eric~P. Xing} {and} \bibinfo{person}{Tony
  Jebara}} (Eds.). \bibinfo{publisher}{PMLR}, \bibinfo{address}{Bejing, China},
  \bibinfo{pages}{1278--1286}.
\newblock
\urldef\tempurl%
\url{https://proceedings.mlr.press/v32/rezende14.html}
\showURL{%
\tempurl}


\bibitem[Ruoss et~al\mbox{.}(2020)]%
        {ruoss2020learning}
\bibfield{author}{\bibinfo{person}{Anian Ruoss}, \bibinfo{person}{Mislav
  Balunovic}, \bibinfo{person}{Marc Fischer}, {and} \bibinfo{person}{Martin
  Vechev}.} \bibinfo{year}{2020}\natexlab{}.
\newblock \showarticletitle{Learning Certified Individually Fair
  Representations}. In \bibinfo{booktitle}{\emph{Advances in Neural Information
  Processing Systems}}, \bibfield{editor}{\bibinfo{person}{H.~Larochelle},
  \bibinfo{person}{M.~Ranzato}, \bibinfo{person}{R.~Hadsell},
  \bibinfo{person}{M.F. Balcan}, {and} \bibinfo{person}{H.~Lin}} (Eds.),
  Vol.~\bibinfo{volume}{33}. \bibinfo{publisher}{Curran Associates, Inc.},
  \bibinfo{pages}{7584--7596}.
\newblock
\urldef\tempurl%
\url{https://proceedings.neurips.cc/paper/2020/file/55d491cf951b1b920900684d71419282-Paper.pdf}
\showURL{%
\tempurl}


\bibitem[Sanchez-Martin et~al\mbox{.}(2021)]%
        {sanchez2021vaca}
\bibfield{author}{\bibinfo{person}{Pablo Sanchez-Martin},
  \bibinfo{person}{Miriam Rateike}, {and} \bibinfo{person}{Isabel Valera}.}
  \bibinfo{year}{2021}\natexlab{}.
\newblock \showarticletitle{VACA: Design of Variational Graph Autoencoders for
  Interventional and Counterfactual Queries}.
\newblock \bibinfo{journal}{\emph{arXiv preprint arXiv:2110.14690}}
  (\bibinfo{year}{2021}).
\newblock


\bibitem[Singh and Ramamurthy(2019)]%
        {singh2019understanding}
\bibfield{author}{\bibinfo{person}{Moninder Singh} {and}
  \bibinfo{person}{Karthikeyan~Natesan Ramamurthy}.}
  \bibinfo{year}{2019}\natexlab{}.
\newblock \showarticletitle{Understanding racial bias in health using the
  Medical Expenditure Panel Survey data}.
\newblock \bibinfo{journal}{\emph{arXiv preprint arXiv:1911.01509}}
  (\bibinfo{year}{2019}).
\newblock


\bibitem[Sohn et~al\mbox{.}(2015)]%
        {sohn2015learning}
\bibfield{author}{\bibinfo{person}{Kihyuk Sohn}, \bibinfo{person}{Honglak Lee},
  {and} \bibinfo{person}{Xinchen Yan}.} \bibinfo{year}{2015}\natexlab{}.
\newblock \showarticletitle{Learning Structured Output Representation using
  Deep Conditional Generative Models}. In \bibinfo{booktitle}{\emph{Advances in
  Neural Information Processing Systems}},
  \bibfield{editor}{\bibinfo{person}{C.~Cortes}, \bibinfo{person}{N.~Lawrence},
  \bibinfo{person}{D.~Lee}, \bibinfo{person}{M.~Sugiyama}, {and}
  \bibinfo{person}{R.~Garnett}} (Eds.), Vol.~\bibinfo{volume}{28}.
  \bibinfo{publisher}{Curran Associates, Inc.}
\newblock
\urldef\tempurl%
\url{https://proceedings.neurips.cc/paper/2015/file/8d55a249e6baa5c06772297520da2051-Paper.pdf}
\showURL{%
\tempurl}


\bibitem[Song et~al\mbox{.}(2019)]%
        {song2019learning}
\bibfield{author}{\bibinfo{person}{Jiaming Song}, \bibinfo{person}{Pratyusha
  Kalluri}, \bibinfo{person}{Aditya Grover}, \bibinfo{person}{Shengjia Zhao},
  {and} \bibinfo{person}{Stefano Ermon}.} \bibinfo{year}{2019}\natexlab{}.
\newblock \showarticletitle{Learning Controllable Fair Representations}. In
  \bibinfo{booktitle}{\emph{Proceedings of the Twenty-Second International
  Conference on Artificial Intelligence and Statistics}}
  \emph{(\bibinfo{series}{Proceedings of Machine Learning Research},
  Vol.~\bibinfo{volume}{89})}, \bibfield{editor}{\bibinfo{person}{Kamalika
  Chaudhuri} {and} \bibinfo{person}{Masashi Sugiyama}} (Eds.).
  \bibinfo{publisher}{PMLR}, \bibinfo{pages}{2164--2173}.
\newblock
\urldef\tempurl%
\url{https://proceedings.mlr.press/v89/song19a.html}
\showURL{%
\tempurl}


\bibitem[Wachter et~al\mbox{.}(2021)]%
        {wachter2020bias}
\bibfield{author}{\bibinfo{person}{Sandra Wachter}, \bibinfo{person}{Brent
  Mittelstadt}, {and} \bibinfo{person}{Chris Russell}.}
  \bibinfo{year}{2021}\natexlab{}.
\newblock \showarticletitle{Bias preservation in machine learning: the legality
  of fairness metrics under EU non-discrimination law}.
\newblock \bibinfo{journal}{\emph{West Virginia Law Review}}
  \bibinfo{volume}{123}, \bibinfo{number}{3} (\bibinfo{year}{2021}).
\newblock


\bibitem[Wick et~al\mbox{.}(2019)]%
        {wick2019unlocking}
\bibfield{author}{\bibinfo{person}{Michael Wick}, \bibinfo{person}{swetasudha
  panda}, {and} \bibinfo{person}{Jean-Baptiste Tristan}.}
  \bibinfo{year}{2019}\natexlab{}.
\newblock \showarticletitle{Unlocking Fairness: a Trade-off Revisited}. In
  \bibinfo{booktitle}{\emph{Advances in Neural Information Processing
  Systems}}, \bibfield{editor}{\bibinfo{person}{H.~Wallach},
  \bibinfo{person}{H.~Larochelle}, \bibinfo{person}{A.~Beygelzimer},
  \bibinfo{person}{F.~d\textquotesingle Alch\'{e}-Buc},
  \bibinfo{person}{E.~Fox}, {and} \bibinfo{person}{R.~Garnett}} (Eds.),
  Vol.~\bibinfo{volume}{32}. \bibinfo{publisher}{Curran Associates, Inc.}
\newblock
\urldef\tempurl%
\url{https://proceedings.neurips.cc/paper/2019/file/373e4c5d8edfa8b74fd4b6791d0cf6dc-Paper.pdf}
\showURL{%
\tempurl}


\bibitem[Xu et~al\mbox{.}(2018)]%
        {xu2018fairgan}
\bibfield{author}{\bibinfo{person}{Depeng Xu}, \bibinfo{person}{Shuhan Yuan},
  \bibinfo{person}{Lu Zhang}, {and} \bibinfo{person}{Xintao Wu}.}
  \bibinfo{year}{2018}\natexlab{}.
\newblock \showarticletitle{FairGAN: Fairness-aware Generative Adversarial
  Networks}. In \bibinfo{booktitle}{\emph{2018 IEEE International Conference on
  Big Data (Big Data)}}. \bibinfo{pages}{570--575}.
\newblock
\urldef\tempurl%
\url{https://doi.org/10.1109/BigData.2018.8622525}
\showDOI{\tempurl}


\bibitem[Yosinski et~al\mbox{.}(2014)]%
        {yosinski2014transferable}
\bibfield{author}{\bibinfo{person}{Jason Yosinski}, \bibinfo{person}{Jeff
  Clune}, \bibinfo{person}{Yoshua Bengio}, {and} \bibinfo{person}{Hod Lipson}.}
  \bibinfo{year}{2014}\natexlab{}.
\newblock \showarticletitle{How transferable are features in deep neural
  networks?}. In \bibinfo{booktitle}{\emph{Advances in Neural Information
  Processing Systems}}, \bibfield{editor}{\bibinfo{person}{Z.~Ghahramani},
  \bibinfo{person}{M.~Welling}, \bibinfo{person}{C.~Cortes},
  \bibinfo{person}{N.~Lawrence}, {and} \bibinfo{person}{K.Q. Weinberger}}
  (Eds.), Vol.~\bibinfo{volume}{27}. \bibinfo{publisher}{Curran Associates,
  Inc.}
\newblock
\urldef\tempurl%
\url{https://proceedings.neurips.cc/paper/2014/file/375c71349b295fbe2dcdca9206f20a06-Paper.pdf}
\showURL{%
\tempurl}


\bibitem[Zafar et~al\mbox{.}(2017a)]%
        {zafar2017fairness}
\bibfield{author}{\bibinfo{person}{Muhammad~Bilal Zafar},
  \bibinfo{person}{Isabel Valera}, \bibinfo{person}{Manuel Gomez~Rodriguez},
  {and} \bibinfo{person}{Krishna~P. Gummadi}.}
  \bibinfo{year}{2017}\natexlab{a}.
\newblock \showarticletitle{Fairness Beyond Disparate Treatment \&; Disparate
  Impact: Learning Classification without Disparate Mistreatment}. In
  \bibinfo{booktitle}{\emph{Proceedings of the 26th International Conference on
  World Wide Web}} (Perth, Australia) \emph{(\bibinfo{series}{WWW '17})}.
  \bibinfo{publisher}{International World Wide Web Conferences Steering
  Committee}, \bibinfo{address}{Republic and Canton of Geneva, CHE},
  \bibinfo{pages}{1171–1180}.
\newblock
\showISBNx{9781450349130}
\urldef\tempurl%
\url{https://doi.org/10.1145/3038912.3052660}
\showDOI{\tempurl}


\bibitem[Zafar et~al\mbox{.}(2019)]%
        {zafar2019fairness}
\bibfield{author}{\bibinfo{person}{Muhammad~Bilal Zafar},
  \bibinfo{person}{Isabel Valera}, \bibinfo{person}{Manuel Gomez-Rodriguez},
  {and} \bibinfo{person}{Krishna~P. Gummadi}.} \bibinfo{year}{2019}\natexlab{}.
\newblock \showarticletitle{Fairness Constraints: A Flexible Approach for Fair
  Classification}.
\newblock \bibinfo{journal}{\emph{Journal of Machine Learning Research}}
  \bibinfo{volume}{20}, \bibinfo{number}{75} (\bibinfo{year}{2019}),
  \bibinfo{pages}{1--42}.
\newblock
\urldef\tempurl%
\url{http://jmlr.org/papers/v20/18-262.html}
\showURL{%
\tempurl}


\bibitem[Zafar et~al\mbox{.}(2017b)]%
        {zafar2017bfairness}
\bibfield{author}{\bibinfo{person}{Muhammad~Bilal Zafar},
  \bibinfo{person}{Isabel Valera}, \bibinfo{person}{Manuel~Gomez Rogriguez},
  {and} \bibinfo{person}{Krishna~P. Gummadi}.}
  \bibinfo{year}{2017}\natexlab{b}.
\newblock \showarticletitle{{Fairness Constraints: Mechanisms for Fair
  Classification}}. In \bibinfo{booktitle}{\emph{Proceedings of the 20th
  International Conference on Artificial Intelligence and Statistics}}
  \emph{(\bibinfo{series}{Proceedings of Machine Learning Research},
  Vol.~\bibinfo{volume}{54})}, \bibfield{editor}{\bibinfo{person}{Aarti Singh}
  {and} \bibinfo{person}{Jerry Zhu}} (Eds.). \bibinfo{publisher}{PMLR},
  \bibinfo{pages}{962--970}.
\newblock
\urldef\tempurl%
\url{https://proceedings.mlr.press/v54/zafar17a.html}
\showURL{%
\tempurl}


\bibitem[Zemel et~al\mbox{.}(2013)]%
        {zemel2013learning}
\bibfield{author}{\bibinfo{person}{Rich Zemel}, \bibinfo{person}{Yu Wu},
  \bibinfo{person}{Kevin Swersky}, \bibinfo{person}{Toni Pitassi}, {and}
  \bibinfo{person}{Cynthia Dwork}.} \bibinfo{year}{2013}\natexlab{}.
\newblock \showarticletitle{Learning Fair Representations}. In
  \bibinfo{booktitle}{\emph{Proceedings of the 30th International Conference on
  Machine Learning}} \emph{(\bibinfo{series}{Proceedings of Machine Learning
  Research}, Vol.~\bibinfo{volume}{28})},
  \bibfield{editor}{\bibinfo{person}{Sanjoy Dasgupta} {and}
  \bibinfo{person}{David McAllester}} (Eds.). \bibinfo{publisher}{PMLR},
  \bibinfo{address}{Atlanta, Georgia, USA}, \bibinfo{pages}{325--333}.
\newblock
\urldef\tempurl%
\url{https://proceedings.mlr.press/v28/zemel13.html}
\showURL{%
\tempurl}


\bibitem[Zhang et~al\mbox{.}(2022)]%
        {zhang2020fairness}
\bibfield{author}{\bibinfo{person}{Tao Zhang}, \bibinfo{person}{Tianqing Zhu},
  \bibinfo{person}{Jing Li}, \bibinfo{person}{Mengde Han},
  \bibinfo{person}{Wanlei Zhou}, {and} \bibinfo{person}{Philip~S. Yu}.}
  \bibinfo{year}{2022}\natexlab{}.
\newblock \showarticletitle{Fairness in Semi-Supervised Learning: Unlabeled
  Data Help to Reduce Discrimination}.
\newblock \bibinfo{journal}{\emph{IEEE Transactions on Knowledge and Data
  Engineering}} \bibinfo{volume}{34}, \bibinfo{number}{4}
  (\bibinfo{year}{2022}), \bibinfo{pages}{1763--1774}.
\newblock
\urldef\tempurl%
\url{https://doi.org/10.1109/TKDE.2020.3002567}
\showDOI{\tempurl}


\bibitem[Zhu et~al\mbox{.}(2022)]%
        {zhu2021rich}
\bibfield{author}{\bibinfo{person}{Zhaowei Zhu}, \bibinfo{person}{Tianyi Luo},
  {and} \bibinfo{person}{Yang Liu}.} \bibinfo{year}{2022}\natexlab{}.
\newblock \showarticletitle{The Rich Get Richer: Disparate Impact of
  Semi-Supervised Learning}. In \bibinfo{booktitle}{\emph{International
  Conference on Learning Representations}}.
\newblock
\urldef\tempurl%
\url{https://openreview.net/forum?id=DXPftn5kjQK}
\showURL{%
\tempurl}


\end{thebibliography}

\clearpage
\appendix
\newpage
\section{Proofs}\label{apx:proofs}

\subsection{Proof Lemma 1}\label{apx:proof_entropy}

\begin{proof}
Let 
$V = \{\X , \Yprox\}$. Considering $V=f(S, \Y,  \Ecal)$, {where $S,\Y$ and noise $\Ecal$ are pairwise independent}. We want to determine ${H(V \mid S)}$:
\begin{equation} \label{eq:lemma1-first}
    I(\V ; \Y \mid S)=H(V \mid S)-H(\V \mid \Y, S)
\end{equation}
\begin{equation}\label{eq:lemma1-second}
    I(\V ; \Y \mid S)=H(\Y \mid S)-H(\Y \mid \X, S)
\end{equation}
Equating (\ref{eq:lemma1-first}) and (\ref{eq:lemma1-second}) gives:
\begin{equation*}
\begin{split}
    H(\V \mid S) - H(V \mid \Y, S) & = H(\Y \mid S) - H(\Y \mid \X, S) \\
H(\V \mid S) & = H(\Y \mid S) - H(\Y \mid \V, S) + H(V \mid \Y, S) \\
&=H(\Y) - H(\Y \mid \V) + H(\V \mid \Y, S)
\end{split}
\end{equation*}
Assuming $f$ is a bijective function with a joint probability mass function $p(\Y, S, \Ecal)$, where due to pairwise independence $p(\y, s, \epsilon) = p(\y)p(s)p(\epsilon)$, we get: 
\begin{align*}
\begin{split}
     & H(\Y \mid \V) = { H(\Y \mid f(S, \Y,  \Ecal))} \\
     & = -\sum_{y, s, \epsilon} p(\y, s, \epsilon)  \log \frac{p(\y, s, \epsilon)  }{p(\y, s, \epsilon)  } \\
     & = -\sum_{y, s, \epsilon} p(\y, s, \epsilon)  \log 1   = 0
\end{split}
\end{align*}
\begin{align*}
\begin{split}
     H(\V & \mid \Y, S) = H(f(S, \Y,  \Ecal) \mid \Y, S)\\
     &=   -\sum_{\y, s, \epsilon} p(\y, s,\epsilon)  \log \frac{p(\y, s,\epsilon)  }{p(\y, s)  }
      = - \sum_{\y, s, \epsilon} p(\y, s,\epsilon)  \log {p(\epsilon)  }  \\
     & = - \sum_{\y} p(\y) \sum_s p(s) \sum_\epsilon p(\epsilon)  \log {p(\epsilon)  }  = - \sum_\epsilon p(\epsilon)  \log {p(\epsilon)  } \\
     & =  H(\Ecal)
\end{split}
\end{align*}

This leads to:
\begin{align*}
\begin{split}
     H(\Y \mid \V) &=H(\Y) - H(\Y \mid \V) + H(\V \mid \Y, S)\\
     &=H(\Y) + H(\Ecal)
\end{split}
\end{align*}

\end{proof}

\subsection{Proof Lemma 2}\label{apx:proof_latent}
In this proof we follow \cite{alemi2018information}. {Let $V = \{\X , \Yprox\}$.}
Here $p$ denotes the original data distribution, $p_{\theta}$ refers to the distribution induced by the decoder, and $q_{\phi}$ refers to the distribution induced by the encoder of the latent variable model.
\begin{proof}

{\small
\begin{align*}
    \begin{split}
          &{I(\V; \Z \mid S)} \\
          & = \int_s \int_z \int_v p(v, z, s) \log\left(\frac{p(s)p(v, z, s)}{p(v, s)p(z, s)}\right) \\
          & = \int_s \int_z \int_v p(v \mid z, s)p(z,s) \log\left(\frac{p(s)p(v \mid z, s)p(z,s)}{p(v, s)p(z, s)}\right) \\
          & = \int_s \int_z \int_v  p(v \mid z, s)p(z,s) \log\left(\frac{p(s)p(v \mid z, s)}{p(v \mid s)p(s)}\right)\\
          & = \int_s \int_z \int_v  p(z,s) \left[p(v \mid z, s) \log\left(\frac{p(v \mid z, s)}{p(v \mid s)}\right)  \right]\\
          & = \int_s \int_z p(z,s) \left[  \int_v p(v \mid z, s) \log p(v \mid z, s)  -  \int_v p(v \mid z, s) \log {p(v, s)}\right] \\
          & \geq \int_s \int_z p(z,s) \left[  \int_v p(v \mid z, s) \log p_{\theta}(v \mid z, s)  -   \int_v p(v \mid z, s) \log {p(v, s)}\right] \\
          & = \int_s \int_z \int_v p(v, z, s) \log \left( \frac{p_{\theta}(v \mid z, s)}{p(v \mid s)} \right)\\
          & = \int_s \int_z \int_v p(z \mid v, s) p(v, s) \log \left( \frac{p_{\theta}(v \mid z, s)}{p(v \mid s)} \right)\\
          & = \int_s \int_z \int_v q_{\phi}(z \mid v, s) p(v, s) \log \left( \frac{p_{\theta}(v \mid z, s)}{p(v \mid s)} \right)\\
          & = \int_s \int_v p(v, s) \int_z q_{\phi}(z \mid v, s)  \log  {p_{\theta}(v \mid z, s)} \\
          & \quad - \int_s \int_v p(v, s) \int_z q_{\phi}(z \mid v, s) \log  p(v \mid s) \\
          & = \int_s \int_v p(v, s) \int_z q_{\phi}(z \mid v, s)  \log \left( {p_{\theta}(v \mid z, s)}\right) \\
          & \quad - \int_s \int_v p(v, s)  \log p(v \mid s) \\
          & = \underbrace{\left(- \int_s \int_v p(v, s)  \log p(v \mid s)\right)}_{H(V|S)}\\
          & \quad - \underbrace{\left(- \int_s \int_v p(v, s) \int_z q_{\phi}(z \mid v, s)  \log \left( {p_{\theta}(v \mid z, s)}\right) \right)}_{\Delta}\\
    \end{split}
\end{align*}
}

Thus $I({\V} \mid S ; \Z) = H({\V} \mid S) - \Delta$ with distortion $\Delta$ that measures the approximation error or reconstruction ability of the latent variable model.

Note that we obtain the inequality using: 
\begin{align*}
    \begin{split}
        KL(p(v \mid z, s) \mid \mid p_{\theta}(v \mid z, s)) & \geq 0\\
        \int_v p(v \mid z, s) \log p(v \mid z, s) &\geq \int p(v \mid z, s)\log  p_{\theta}(v \mid z, s)
    \end{split}
\end{align*}
\end{proof}

\begin{figure}[t]
    \centering
    \input{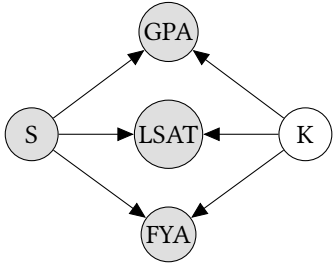}
    \caption{Synthetic data setting following \cite{Kusner2017-ln}, generative model.}
    \label{fig:scb}
\end{figure}

\section{Datasets}\label{apx:datasets}
\subsection{Synthetic Dataset}
Our synthetic dataset $\SCB$ follows~\cite{Kusner2017-ln} and models the Law school admission scenario. 
{It includes two continuous observed variables (test-scores LSAT, GPA), one sensitive variable (Gender S), and one unobserved confounding variable (hidden potential Y expressed as \emph{talent (knowledge) K}).}
The label $\Yprox$ indicates whether a student passes or fails the first year of Law school and is determined by thresholding the First-Year Average (FYA).
The {generative model} is shown in Figure~\ref{fig:scb}.

\begin{equation}
\label{eqn:single_conf_1a}
	\begin{split}
	    Y &= 2 \cdot (m \sim \text{Bern}(0.5)) - 1 \\
	    K &\sim \mathcal{N}(Y, 0.5) \\
		\text{LSAT} &\sim \mathcal{N}(4 K + 3.5 S, 0.1) \\
		\text{GPA} &\sim \mathcal{N}(0.75 K + S, 0.01) \\
		\tilde{Y} &= \mathbb{I}\left[ FYA \sim \mathcal{N}(1.3 K + S, 0.05) > 0\right]
	\end{split}
\end{equation}

\begin{table*}[t]
\caption{Summary of properties of real-world datasets used in our experimental section.}
\label{tab:dataset_info}
\small
%
\begin{tabular}{@{\extracolsep{1pt}}lccccccccc}
\toprule
\multirow{2}{*}{Dataset} &
  \multirow{2}{*}{\# data} &
  \multirow{2}{*}{\# features} &
  \multicolumn{4}{c}{Feature types} &
  \multirow{2}{*}{Protected feature} &
  \multirow{2}{*}{Proxy class} &
  \multirow{2}{*}{Decision task}
  \\ \cline{4-7}
            &       &    & real & count & binary & categorical &      &                             \\ \midrule
COMPAS      & 5278  & 3  & 0    & 1     & 1      & 1           & Race & No recidivism  &  Give bail  \\
\CREDIT      & 1000  & 19 & 5    & 2     & 2      & 10          & Gender  & Credit score   & Award loan             \\
\MEPS & 17570 & 39 & 4    & 0     & 1      & 34          & Race & High utilization & Special care\\
\bottomrule
\end{tabular}
\end{table*}

\subsection{Real-world Datasets}

{We select three real-world datasets for our analysis of fair decision-making. These datasets have different complexity in terms of heterogeneity of features and the total number of features. We show our results on the \COMPAS\ recidivism dataset~\cite{angwin2016machine,larson2016how}, \CREDIT\ (German) Dataset~\cite{dua2019} and \MEPS\ health-expenditure Dataset~\cite{meps2018} (the 2015 MEPS Panel 20 dataset~\cite{aif360}). The characteristics of these datasets are listed in table~\ref{tab:dataset_info}. We obtain all datasets using the AIF360 tool~\cite{aif360}. We describe a possible and simplified decision scenario for each dataset in the following.}

\paragraph{\COMPAS}

{For the \COMPAS\ recidivism dataset, we may aim to decide whether or not to offer an inmate a \emph{reintegration program}. A decision-maker may wish to offer integration support to inmates with high innate potential for reintegration. This potential is unobserved and assumed to be independent of the social construct (i.e., ground truth label $Y$ independent of race $S$). We observe if a person has committed a crime within two years after trial (recidivism). This is the observed proxy label $\Yprox$. We assume arrest to be dependent on the sensitive attribute. 
For example, members of a particular racial group may be profiled and arrested at a significantly higher rate. We may then aim to learn a decision-making algorithm, which decides according to \emph{potential} ($Y$) not \emph{arrest} ($\Yprox$).}

{As explained in the paper, we pre-processed the label $\Yprox$ of the \COMPAS\ dataset such that $\Yprox =1$ indicates \emph{no recidivism}. This is done in order to match our assumption that a person with a positive ground truth label should receive a positive decision.}

\paragraph{\CREDIT}

{For the \CREDIT\ dataset, we may aim to decide whether or not to give a credit to an individual. A decision-maker may like to provide credit to people who have a high \emph{willingness} to repay. This willingness is unobserved and assumed to be independent of the social construct (i.e., ground truth label $Y$ independent of gender $S$). Instead, we observe a \emph{credit risk score} (i.e., proxy label $\Yprox$) and reject individuals with poor credit risk scores. We assume that societal discrimination (e.g., gender wage gaps) causes risk scores to disadvantage certain genders $S$, even if an individual is honest and willing to repay. Our goal may then be to create a fair decision-making system that lends to individuals depending on their willingness to pay back ($Y$) not risk score ($\Yprox$).}

\paragraph{\MEPS}

{Care management often uses datasets such as $\MEPS$ Health~\cite{meps2018} to predict individuals who have increased health care needs. This helps ensure appropriate attention and care to reduce overall costs and improve the quality of services. A decision-maker may wish to make decisions based on the unobserved \emph{actual medical needs} of individuals (i.e., ground truth label $Y$ independent of race $S$). Instead, we may observe a person's number of hospital visits during a particular time period (i.e., proxy label $\Yprox$). We may assume that the more visits, the higher the likelihood that the person has a high need for medical care. We assume the number of visits to be dependent on the sensitive attribute, for example, due to unequal access to health care~~\cite{singh2019understanding}. Our goal may then decide to provide individuals with care management that have a high genuine medical needs ($\Y$) not a high number of hospital visits ($\Yprox$).}

\section{Practical considerations} \label{apx:implementation}

\subsection{Generative and Inference Model} \label{apx:generative_inference_model}

\paragraph{Generative model}
{Our generative model approximates features $X$ and utility $U$.
Note that for unlabeled data ($d=0$), their \emph{actual observed utility} is  0.
This enables us to easily train the model without any IPS correction.}
\begin{equation}
   p_\theta(x,\tilde{u},z|s,d)=p_\theta(x|z,s) p_\theta(\tilde{u}|z,s,d) p(z),
\end{equation}
where 
 \begin{equation}
 p_\theta(\tilde{u}|z,s,d) = (1-d)* \mathbb{I}\{\tilde{u}=0\} + d* p_\theta(\tilde{u} | z,s, d=1),
 \end{equation}
 
\paragraph{Inference model}
{
Note that our inference model is always conditioned on the positive decision ($d=1$).
This helps us encode useful information regarding which individual has \emph{actual positive/negative utility}.
}
{
For unlabeled data, we observe $u=0$, so we do not know the \emph{actual positive/negative utility}.
For estimating the same while avoiding the need for IPS bias correction, we utilize a separate classifier model.
This idea comes from the semi-supervised literature~\cite{louizos2015variational}.
The classifier is trained only with labeled data (with IPS) and approximates if an unlabeled datum \emph{would have positive or negative utility}. 
}
\begin{equation}
q_{\omega, \phi}(z|x,s, d=1) = q_\omega(\tilde{u}|x,s, d=1)q_\phi(z|x,\tilde{u},s, d=1)
\end{equation}
such that
\begin{equation}
\begin{split}
    & q_{\omega, \phi}(z|x,s, d=1) \\
    &= \int q_\phi (z|\tilde{u}, s, x,d=1) q_\omega (\tilde{u}|x,s, d=1) d\tilde{u} 
    \\
    &= \E_{q_\omega (\tilde{u}|x,s, d=1) }[q_\phi (z|\tilde{u}, x,s, d=1)]
\end{split}
\end{equation}
where
\begin{align}
    q_\phi(z|x,\tilde{u},s, d=1) &= \mathcal{N}(z|\mu_\phi(\tilde{u},x, s),diag(\sigma_\phi(\tilde{u},x, s))) \\
    q_\omega(\tilde{u}|x, s, d=1) &= Bern (\gamma_\omega(x, s))
\end{align}

\subsection{Monte-Carlo Estimation of KL Divergence} \label{apx:KL_MC}
In the binary case (which we assume), the KL-divergence for the unlabeled scenario in our approach as in Eq. \eqref{eq:elbo_unsup} can be computed as:

{\small
\begin{equation}
\begin{split}
& KL\Big[\int_{\tilde{u}} q_\phi (z|x,{\tilde{u}}) q_\phi ({\tilde{u}}|x) d{\tilde{u}}||p(z)\Big]\\
& =\int_z \log \left(\frac{\int q_\phi (z|x) q_\phi ({\tilde{u}}|x) d{\tilde{u}}}{p(z)}\right)  \int_{\tilde{u}} q_\phi (z|x,{\tilde{u}}) q_\phi ({\tilde{u}}|x) d{\tilde{u}} dz\\
& =\int_{\tilde{u}} \int_z \log \left(\frac{\int q_\phi (z|x) q_\phi ({\tilde{u}}|x) d{\tilde{u}}}{p(z)}\right)  q_\phi (z|x,{\tilde{u}}) dz q_\phi ({\tilde{u}}|x) d{\tilde{u}}\\
& =q_\phi ({\tilde{u}}=1|x)\int_z \log \left(\frac{\int q_\phi (z|x,{\tilde{u}}) q_\phi ({\tilde{u}}|x) d{\tilde{u}}}{p(z)}\right)  q_\phi (z|x,{\tilde{u}}=1) dz \\
& + q_\phi ({\tilde{u}}=0|x)\int_z \log \left(\frac{\int q_\phi (z|x,{\tilde{u}}) q_\phi ({\tilde{u}}|x) d{\tilde{u}}}{p(z)}\right)  q_\phi (z|x,{\tilde{u}}=0)dz \\
& =q_\phi ({\tilde{u}}=1|x)\\
& \quad \int_z \log \left(\frac{q_\phi ({\tilde{u}}=1|x) q_\phi (z|x,{\tilde{u}}=1)+ q_\phi ({\tilde{u}}=0|x) q_\phi (z|x,{\tilde{u}}=0)}{p(z)}\right)  \\
& \quad \quad q_\phi (z|x,{\tilde{u}}=1) dz + q_\phi ({\tilde{u}}=0|x)\\
& \quad \int_z \log \left(\frac{q_\phi ({\tilde{u}}=1|x) q_\phi (z|x,{\tilde{u}}=1)+ q_\phi ({\tilde{u}}=0|x) q_\phi (z|x,{\tilde{u}}=0)}{p(z)} \right)  \\
& \quad \quad q_\phi (z|x,{\tilde{u}}=0) dz
\end{split}
\end{equation}
}

Since there is no closed form solution, we approximate the KL-divergence using Monte-Carlo: 

{\small
\begin{equation} \label{eq:MC_KL}
\begin{split}
& q_\phi ({\tilde{u}}=1|x)  \frac{1}{K}  \sum_{z \sim q_\phi (z|x,{\tilde{u}}=1)} \\
& \log \left(\frac{q_\phi ({\tilde{u}}=1|x) q_\phi (z|x,{\tilde{u}}=1)+ q_\phi (u=0|x) q_\phi (z|x,{\tilde{u}}=0)}{p(z)}\right)  \\
& + q_\phi ({\tilde{u}}=0|x) \frac{1}{K} \sum_{z \sim q_\phi (z|x,{\tilde{u}}=0)} \\
& \quad \log \left(\frac{q_\phi ({\tilde{u}}=1|x) q_\phi (z|x,{\tilde{u}}=1)+ q_\phi ({\tilde{u}}=0|x) q_\phi (z|x,{\tilde{u}}=0)}{p(z)} \right) \\
& = q_\phi ({\tilde{u}}=1|x)  \frac{1}{K}  \sum_{z \sim q_\phi (z|x,{\tilde{u}}=1)} \log (q_\phi ({\tilde{u}}=1|x) q_\phi (z|x,{\tilde{u}}=1)\\
& + q_\phi ({\tilde{u}}=0|x) q_\phi (z|x,{\tilde{u}}=0)) - \log p(z) \\
& + q_\phi ({\tilde{u}}=0|x)  \frac{1}{K}  \sum_{z \sim q_\phi (z|x,{\tilde{u}}=0)} \log (q_\phi ({\tilde{u}}=1|x) q_\phi (z|x,{\tilde{u}}=1) \\
& + q_\phi ({\tilde{u}}=0|x) q_\phi (z|x,{\tilde{u}}=0))  - \log p(z)
\end{split}
\end{equation}
}

where we set $K=100$ as the number of samples. 

In the implementation we evaluate the log probability of the first term in Eq. \eqref{eq:MC_KL} as: 
\begin{equation}
\begin{split}
& \operatorname{logsumexp}(x)=\log ( \exp \left(\log(q_\phi ({\tilde{u}}=1|x) q_\phi (z|x,{\tilde{u}}=1)) \right) \\
& \quad + \exp \left(\log (q_\phi ({\tilde{u}}=0|x) q_\phi (z|x,{\tilde{u}}=0)) \right) )
\end{split}
\end{equation}

\subsection{Optimal Utilities for \SCB\ Data}
\label{apx:opt_util}

{In our $\SCB$ setting we have $\X \dep S$ and $K\indep S$ with observed features $X \coloneq (\text{LSAT}, \text{GPA})$ and the true hidden factor $K$, as in Eq. \ref{eqn:single_conf_1a}. Then we define the optimal unfair policy (\optimalunfair) to take decision $d$ according to the posterior distribution $d \sim p(\Yprox|X,S)$ and the optimal fair policy (\optimalfair) according to $d \sim p(\Yprox|K)$. While in our synthetic generation process, we do not have access to the posterior distribution, we can generate i.i.d. samples from it.} 
{We approximate the posterior distribution by training logistic regression models on these samples.}
{We approximate the unfair policy (\optimalunfair) by training the logistic model  $\Yprox = f\left(X,S\right)$ and the optimal fair policy (\optimalfair) by training the logistic model $\Yprox = g\left(K\right)$.}
{We compute the optimal utility and fairness measures on a held-out test set by taking the mean over five independent runs.}

\subsection{Counterfactual Generation on \SCB\ Data} \label{apx:counterfactuals}
{For a factual individual (with sensitive attribute $S=s$), we compute the counterfactual (\emph{had the sensitive attribute been $S=s'$} with $s'\neq s$) following the abduction-action-prediction steps by Pearl in~\cite{pearl2016primer}. We compute counterfactuals for the \SCB\ dataset with access to the generative process (Eq. \ref{eqn:single_conf_1a}) with the following steps:}

\begin{enumerate}
    \item {We sample} $Y$, correspondingly $K,S$ and exogenous noise-factors ${\epsilon}$ for each Gaussian distribution. Based on the values of $Y,K,{\epsilon}$, we compute the factual LSAT, GPA, and $\Yprox$.
    \item We \emph{intervene} by modifying the value of {$S=s$ to $S=s'$} (e.g. if for the factual $S=1$, then we set $S=0$).
    \item We use the same values of $Y,K,{\epsilon}$ and the modified $S=s'$ {and compute the \emph{counterfactual} LSAT and GPA.}
\end{enumerate}

\section{Experimental setup}\label{apx:experimental_setup}

{This chapter provides a complete description of the experiments presented in section~\ref{sec:evaluation}. We describe the initial policies (\ref{apx:initial_policies}), the training process, hyperparameter selection and coding environment (\ref{apx:validation}), metrics (\ref{apx:metrics}), policy model choices (\ref{apx:pol_models}), and baselines (\ref{apx:baselines}).}

\begin{table*}[t]
  \caption{{Acceptance rates (total and per sensitive group $S$) for different datasets and initial policies.}}
  \label{tab:freq}
  \small
  \begin{tabular}{lcccccc}
  \toprule
  Policy & \multicolumn{3}{c}{\LENI}  & \multicolumn{3}{c}{\HARSH}\\
    \hline
    Dataset & p(d=1) & p(d=1|S=1)  & p(d=1|S=0) & p(d=1) & p(d=1|S=1) & P(d=1|S=0)\\ 
    \midrule
    \SCB & \B{0.5468} & 0.7642 & 0.3297 & \B{0.128} & 0.1581 & 0.0979\\
    \COMPAS & \B{0.4995} & 0.7664 & 0.3274 & \B{0.1024} & 0.1519 & 0.0705\\ 
    \CREDIT & \B{0.4857} & 0.5846 & 0.2909 & \B{0.1776} & 0.2123 & 0.1091\\ 
    \MEPS & \B{0.4680} & 0.7826 & 0.2724 & \B{0.1119} & 0.1925 & 0.0617\\
  \bottomrule
\end{tabular}
\end{table*}
 
\subsection{Initial Policies}\label{apx:initial_policies}
{The initial policy is used at the beginning of our decision-making phase (\phasetwo at time $t=0$) before our policy models are trained. We consider two different types of initial policies: \HARSH, which provides a positive decision only to a small fraction of the data (thus resulting in a small number of labeled data points), and \LENI, which provides a larger fraction of the data with a positive decision. See also Figure~\ref{fig:pipeline}.
Table~\ref{tab:freq} shows the acceptance rates for the different policies. Note that because of existing biases in the data, the initial policies are also biased with respect to the sensitive characteristic.}

\subsection{Training and Validation}
\label{apx:validation}
\begin{table*}[t]
\caption{Hyperparameters for different types of models that were tuned for the best model selection. VAE models hyperparameters are used for \name, \ipsvae, \ipsvaelab. All combinations are tested on a separate held-out validation set.}
\label{tab:hyperparam_set}
\small
\begin{tabular}{@{\extracolsep{4pt}}lccccc}
\toprule
\multirow{2}{*}{Model} &
  \multicolumn{1}{l}{\multirow{2}{*}{Parameter}} &
  \multicolumn{4}{c}{Dataset} \\ \cline{3-6} 
 &
  \multicolumn{1}{l}{} &
  \SCB &
  \COMPAS &
  \CREDIT &
  \MEPS \\ \midrule
\multirow{5}{*}{VAE-\phaseone} &
  batch-size &
  64, 128, 256 &
  64, 128, 256 &
  64, 128, 256 &
  64, 128, 256 \\
 &
  learning-rate &
  1e-3, 5e-3, 1e-2 &
  1e-3, 5e-3, 1e-2 &
  1e-3, 5e-3, 1e-2 &
  1e-3,  5e-3, 1e-2 \\
 &
  vae-arch &
  \begin{tabular}[c]{@{}c@{}}(32x32), (32x32x32),\\ (64x64), (64x64x64)\end{tabular} &
  \begin{tabular}[c]{@{}c@{}}(32x32), (32x32x32),\\ (64x64), (64x64x64)\end{tabular} &
  \begin{tabular}[c]{@{}c@{}}(32x32), (32x32x32),\\ (64x64), (64x64x64)\end{tabular} &
  \begin{tabular}[c]{@{}c@{}}(64x64), (64x64x64),\\ (100x100), (100x100x100)\end{tabular} \\
 &
  latent-size &
  2 &
  2, 3 &
  10, 12, 14, 16 &
  20, 22, 25, 27, 30 \\
 &
  beta &
  0.7, 0.8, 0.9, 1.0 &
  0.7, 0.8, 0.9, 1.0 &
  0.7, 0.8, 0.9, 1.0 &
  0.7, 0.8, 0.9, 1.0 \\ \midrule
\multirow{5}{*}{VAE-\phasetwo} &
  learning-rate &
  1e-3, 1e-2 &
  1e-3, 1e-2 &
  1e-3, 1e-2 &
  1e-3, 1e-2 \\
 &
  clf-arch &
  \begin{tabular}[c]{@{}c@{}}(32x32), (32x32x32),\\ (64x64), (64x64x64)\end{tabular} &
  \begin{tabular}[c]{@{}c@{}}(32x32), (32x32x32),\\ (64x64), (64x64x64)\end{tabular} &
  \begin{tabular}[c]{@{}c@{}}(32x32), (32x32x32),\\ (64x64), (64x64x64)\end{tabular} &
  \begin{tabular}[c]{@{}c@{}}(64x64), (64x64x64),\\ (100x100), (100x100x100)\end{tabular} \\
 &
  clf-dropout &
  0.0, 0.1 &
  0.0, 0.1 &
  0.0, 0.1 &
  0.0, 0.1 \\
 &
  alpha &
  1, 5, 10, 15 &
  1, 5, 10, 15 &
  1, 5, 10, 15 &
  1, 7, 15 \\
 &
  beta &
  0.7, 0.85, 1.0 &
  0.7, 0.85, 1.0 &
  0.7, 0.85, 1.0 &
  0.7, 0.85, 1 \\ \midrule
\multirow{3}{*}{\niki} &
  \multicolumn{1}{c}{learning-rate} &
  1e-3, 1e-2 &
  1e-3, 1e-2 &
  1e-3, 1e-2 &
  1e-3, 5e-3, 1e-2 \\
 &
  \multicolumn{1}{c}{clf-arch} &
  \multicolumn{1}{l}{\begin{tabular}[c]{@{}l@{}}(32x32), (32x32x32),\\ (64x64), (64x64x64)\end{tabular}} &
  \multicolumn{1}{l}{\begin{tabular}[c]{@{}l@{}}(32x32), (32x32x32),\\ (64x64), (64x64x64)\end{tabular}} &
  \multicolumn{1}{l}{\begin{tabular}[c]{@{}l@{}}(32x32), (32x32x32),\\ (64x64), (64x64x64)\end{tabular}} &
  \multicolumn{1}{l}{\begin{tabular}[c]{@{}l@{}}(64x64), (64x64x64),\\ (100x100), (100x100x100)\end{tabular}} \\
 &
  \multicolumn{1}{c}{clf-dropout} &
  0.0, 0.1 &
  0.0, 0.1 &
  0.0, 0.1 &
  0.0, 0.1 \\ \midrule
\multirow{4}{*}{\nikifair} &
  learning-rate &
  1e-3, 1e-2 &
  1e-3, 1e-2 &
  1e-3, 1e-2 &
  1e-3, 5e-3, 1e-2 \\
 &
  clf-arch &
  \begin{tabular}[c]{@{}c@{}}(32x32), (32x32x32),\\ (64x64), (64x64x64)\end{tabular} &
  \begin{tabular}[c]{@{}c@{}}(32x32), (32x32x32),\\ (64x64), (64x64x64)\end{tabular} &
  \begin{tabular}[c]{@{}c@{}}(32x32), (32x32x32),\\ (64x64), (64x64x64)\end{tabular} &
  \begin{tabular}[c]{@{}c@{}}(64x64), (64x64x64),\\ (100x100), (100x100x100)\end{tabular} \\
 &
  clf-dropout &
  0.0, 0.1 &
  0.0, 0.1 &
  0.0, 0.1 &
  0.0, 0.1 \\
 &
  lambda &
  1--15 &
  1--15 &
  1--15 &
  1--15 \\ \bottomrule
\end{tabular}
\end{table*}

\begin{table*}[t]
\caption{Best selected hyperparameters for all the models, for each dataset. Note that for our methods, the policy model using latent $Z$ utilizes the same hyperparameters as the classifier model.}
\label{tab:hyperparam_best}

\small
\begin{tabular}{@{\extracolsep{6pt}}llcccc}
\toprule
\multirow{2}{*}{Model}                                                                  & \multirow{2}{*}{Parameter} & \multicolumn{4}{c}{Dataset}                \\ \cline{3-6} 
    &                            & Synthetic & COMPAS   & \CREDIT   & MEPS     \\ \midrule
\multirow{5}{*}{\begin{tabular}[c]{@{}l@{}}\nameone\\ (\phaseone)\end{tabular}}                                                          & batch-size                 & 64        & 256      & 128      & 256      \\
    & learning-rate              & 5e-3      & 5e-3     & 1e-3     & 1e-3     \\
    & vae-arch                   & 64x64     & 32x32    & 64x64    & 64x64    \\
    & latent-size                & 2         & 3        & 12       & 20       \\
    & beta                       & 0.8       & 0.8      & 0.8      & 0.7      \\ \midrule
\multirow{6}{*}{\begin{tabular}[c]{@{}l@{}}\nameonetwo\\ (\phasetwo)\end{tabular}}   & learning-rate              & 1e-2      & 1e-3     & 1e-2     & 1e-3     \\
    & vae-arch                   & 64x64     & 32x32    & 64x64    & 64x64    \\
    & clf-arch                   & 32x32x32  & 32x32x32 & 32x32x32 & 100x1000 \\
    & clf-dropout                & 0.0       & 0.1      & 0.1      & 0.1      \\
    & alpha                      & 5         & 1        & 5        & 1        \\
    & beta                       & 0.7       & 0.7      & 0.85     & 0.7      \\ \midrule
\multirow{7}{*}{\nametwo} & learning-rate              & 1e-2      & 1e-2     & 1e-2     & 1e-2     \\
    & vae-arch                   & 64x64     & 64x64x64 & 64x64    & 100x100  \\
    & clf-arch                   & 64x64     & 64x64x64 & 32x32    & 100x100  \\
    & clf-dropout                & 0.1       & 0.0      & 0.1      & 0.0      \\
    & latent-size                & 2         & 2        & 12       & 25       \\
    & alpha                      & 5         & 10       & 1        & 1        \\
    & beta                       & 0.85      & 1.0      & 0.7      & 0.7      \\ \midrule
\multirow{3}{*}{\niki}                                                                  & learning-rate              & 1e-2      & 1e-2     & 1e-3     & 5e-3     \\
    & clf-arch                   & 64x64x64  & 32x32x32 & 64x64    & 64x64    \\
    & clf-dropout                & 0.1       & 0.0      & 0.1      & 0.0      \\ \midrule
\multirow{4}{*}{\nikifair}                                                              & learning-rate              & 1e-2      & 1e-2     & 1e-2     & 1e-2     \\
    & clf-arch                   & 64x64x64  & 32x32x32 & 32x32x32 & 64x64    \\
    & clf-dropout                & 0.1       & 0.0      & 0.0      & 0.0      \\
    & lambda                     & 3         & 4        & 2        & 2        \\ \midrule
\multirow{5}{*}{\ipsvae} & learning-rate              & 1e-3      & 1e-3     & 1e-3     & 1e-3     \\
    & vae-arch                   & 64x64     & 32x32    & 64x64    & 64x64    \\
    & latent-size                & 2         & 3        & 12       & 20       \\
    & alpha                      & 1.0       & 1.0      & 1.0      & 1.0      \\
    & beta                       & 0.7       & 0.85     & 0.7      & 0.7      \\ \bottomrule
\end{tabular}
\end{table*}

\subsubsection{{Training Parameters}}

\paragraph{Dataset Size} 
For our synthetic data, we consider 5000 training samples for Phase 1.
In addition, we consider 2500 validation and 5000 test samples.
For \COMPAS, we split the data into 60-40-40 for training-validation-testing.
In the case of \CREDIT, we split the data into 70-15-15 for training-validation-testing.
Finally, for the \MEPS\ dataset, we consider 75-25-25 for training-validation-testing.
For each real-world dataset, we further consider 70\% of the training data in an \emph{unlabeled} fashion for Phase 1 pre-training.
At the beginning of the decision-making phase (Phase 2), we perform a warmup with 128 samples which are sampled using some initial policy (\HARSH, or \LENI).
Finally, in Phase 2, we consider we get 64 samples at \emph{each time-step}.

\paragraph{Training Epochs}
We train \phaseone\ models for 1200 epochs (for the larger \MEPS\ data, we perform 500 epochs instead) for cross-validation.
For final training in \phaseone, we train for 2000 epochs (500 epochs for \MEPS).
Before starting the online decision-learning \phasetwo, we perform a warmup learning where the data is labeled using some initial policy (explained prior).
In the warmup stage, we train each model for 50 steps.
Finally, in the online decision phase, we train for one epoch for each time step.
We perform each evaluation for 200 online decision time-steps.
During the decision stage, we ensure that each model has the same batch size. 
We determine the batch size by splitting the data such that we have 3 batches per epoch, per time-step during training.

\paragraph{Other Parameters}
For all training, we consider the Adam optimizer.
For each dataset, the real variables are standardized to have zero mean and unit variance.
In our semi-labeled training model, for the Monte-Carlo estimate of the KL divergence, we consider 100 samples.
We further consider 50 samples for ELBO computation for the unlabeled samples.

All models are deep neural networks with a fully-connected architecture.
For all neural network models, we consider the ReLU activation function in the hidden layers.
For weight initialization, we consider Xavier uniform weight initialization.
For our policy models that use the latent space of the VAE, we use the same neural architecture as the classifier models.

For the basic utility-fairness analyses, we consider a cost of 0.5 for the decisions for all datasets, except for \MEPS, where we consider 0.1 for positive decisions.
We show a cost-utility-fairness analysis separately.

\subsubsection{Hyperparameter Selection}

For each model in our evaluation section, we perform extensive hyperparameter selection across multiple parameters.
We list each hyperparameter, the combinations for each dataset and the model in Table~\ref{tab:hyperparam_set}.
We evaluate the models on the held-out validation data by training each model for 5 seeds.
We select the best models based on their performance.
For VAE based models we analyze the VAE reconstruction performance and the independence of the latent space with respect to the sensitive feature.
For \phasetwo models, we also test the classification performance.
Note that we do not perform separate validation for our distinct policy model trained from the latent space of VAE.
For each setup, we simply use the same neural architecture as is selected for the classifier model.
For the logistic baseline models (\niki, \nikifair) we test for classification performance and fairness.
The best parameters for each model are shown in Table~\ref{tab:hyperparam_best}.

\subsubsection{{Coding Environment}}

All evaluations are run on a cluster with Intel Xeon E5 family of processors using Linux OS.
Evaluation jobs are submitted to the cluster such that each experiment run is executed on one CPU.
Deep learning implementations are done on PyTorch v1.7 also utilizing the Ignite v0.3.0 library for training.
For \nikifair, the demographic parity constrained loss is applied using fairtorch v0.1.2.
All other machine learning classification models are trained using scikit-learn.
Real-world fairness datasets are loaded using the AIF360 library.

\subsection{Additional Metrics} \label{apx:metrics}

In addition to the metrics defined in Section~\ref{sec:background}, in our empirical evaluation in \ref{sec:evaluation} we report \textit{effective utility} and \textit{effective demographic parity}. %
{\textit{Effective utility} at time $t$ is the average utility that is accumulated by the decision-maker until time $t$ through the learning process. {We define effective utility with respect to ground truth $\y$.} However, in practical settings we often do not have access to $\y$. In this case, we compute effective proxy utility \emph{Effect.} \UTprox\ using proxy label $\yprox$.}

\begin{definition}[Effective Utility \cite{kilbertus2020fair}]\label{eq:utility_eff}
Effective utility: the utility realized during the learning process up to time $t$, i.e.,
$$
\text{Effect. UT}(t)=\frac{1}{N \cdot t} \sum_{t^{\prime} \leq t} \sum_{y_{i} \in \mathcal{D}^{t^{\prime}}}\left(y_{i}-c\right),
$$
where $\mathcal{D}^{t^{\prime}}$ is the data in which the policy $\pi_{t^{\prime}}$ took positive decisions $d_{i}=1$,  $N$ is the number of considered examples at each time step $t$ and $c$ the problem specific costs of a positive decision. 
\end{definition}

{\textit{Effective demographic parity} at time $t$ is the average unfairness accumulated by the decision-maker until time $t$ while learning the decision policy.}

\begin{definition}[Effective Demographic Parity \cite{kilbertus2020fair}]\label{eq:dp_eff}
The demographic parity level realized during the learning process up to time $t$, i.e.,
$$
\text{Effect. DPU}(t)=\frac{1}{t} \sum_{t^{\prime} \leq t} \left( \sum_{d_{i} \in \mathcal{D}_{+1}^{t^{\prime}}}\frac{d_{i}}{|\mathcal{D}_{+1}^{t^{\prime}}|} - \sum_{d_{j} \in \mathcal{D}_{-1}^{t^{\prime}}}\frac{d_{j}}{|\mathcal{D}_{-1}^{t^{\prime}}|} \right),
$$
where $\mathcal{D}_{-1}^{t^{\prime}}$ is the set of decisions received by the group of individuals with $s = -1$ at time $t\prime$ and $\mathcal{D}_{+1}^{t^{\prime}}$ is the set of decisions received by the group of individuals with $s = +1$.  This is the unfairness accumulated by the decision-maker while learning better policies.
\end{definition}

\begin{figure*}[t]
	\centering
	\begin{subfigure}[t]{0.22\textwidth}
		\centering
		\includegraphics{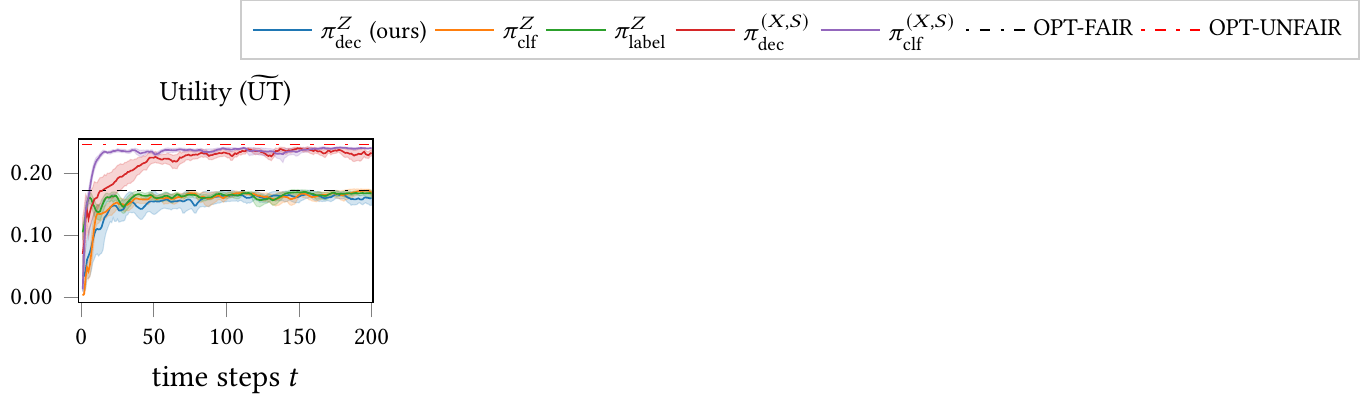}
		\caption{\SCB}
	\end{subfigure}
	\begin{subfigure}[t]{0.22\textwidth}  
		\centering 
		\includegraphics{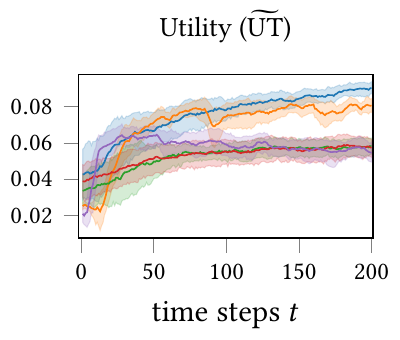}
		\caption{COMPAS}
	\end{subfigure}
	\begin{subfigure}[t]{0.22\textwidth}  
		\centering 
		\includegraphics{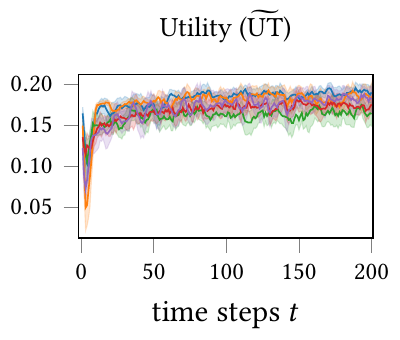}
		\caption{\CREDIT}
	\end{subfigure}	
	\begin{subfigure}[t]{0.22\textwidth}   
		\centering 
		\includegraphics{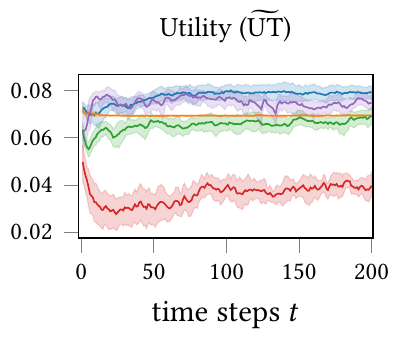}
		\caption{MEPS}
	\end{subfigure}
	\begin{subfigure}[t]{0.22\textwidth}
		\centering
		\includegraphics{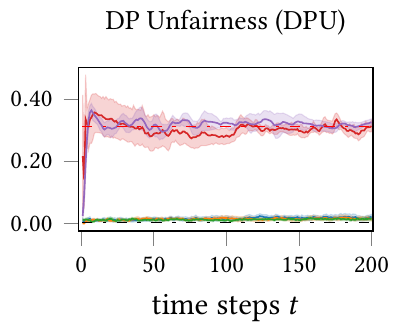}
		\caption{\SCB}
	\end{subfigure}
	\begin{subfigure}[t]{0.22\textwidth}  
		\centering 
		\includegraphics{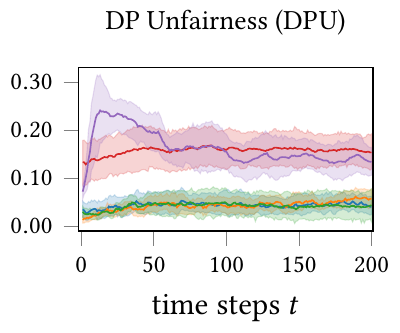}
		\caption{COMPAS}
	\end{subfigure}
	\begin{subfigure}[t]{0.22\textwidth}  
		\centering 
		\includegraphics{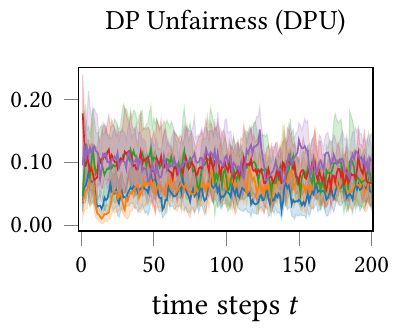}
		\caption{\CREDIT}
	\end{subfigure}	
	\begin{subfigure}[t]{0.22\textwidth}   
		\centering 
		\includegraphics{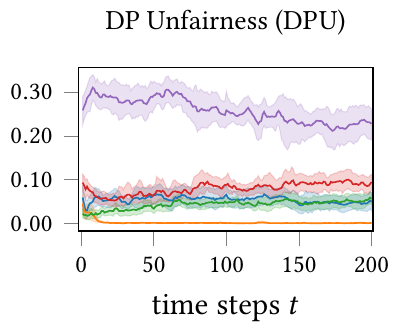}
		\caption{MEPS}
	\end{subfigure}
	\caption{Different policies in our method across different datasets. Harsh initial policy. Note in our evaluations we utilize $\pi^Z_\text{dec}$.}
	\Description{If you require a verbal description of this figure, please contact us, the authors. We are happy to provide you with a description.}
	\label{fig:ours_diff_pols}
\end{figure*}

\subsection{Policy Model Choices}
\label{apx:pol_models}

{Section~\ref{sec:our-approach} provides a detailed overview of our approach and modeling. We can propose several policy model options in $\phasetwo$ (i.e., for policy $\pi_{t+1}$ in Figure~\ref{fig:pipeline}). The policy model is trained at the end of the time step $t$, and then applied to the new set of applicants at time $t+1$. We explain the different policy options in the following, including the model $\pi^Z_\text{dec}$ that we chose for all our evaluations in this paper.}

\subsubsection{Classifier as Policy}

{One option is to use the classification model $\uprox \sim q_\omega(\uprox|x,s, d=1)$ to take decision $d=\uprox$.
One benefit is that this model is trained end-to-end while optimizing Eq. \ref{eq:overall_loss} and hence requires no separate training. 
At the same time, it is trained only on labeled data (and so trained via an IPS-weighted loss function potentially introducing high variance).
We denote this policy by $\policlf$. Note that the model is unfair as it directly utilizes the sensitive feature as input (without any fairness constraints).}

\subsubsection{Decoder as Policy}
{Another option is to deploy the decoder model of the trained VAE.}
{We denote this model as $\polidec$.
The decoder has been trained on labeled and unlabeled data. Receiving $\uprox \sim p_\theta(\tilde{u}|z,s,d)$ and taking decision $d=\uprox$ requires no separate training of a policy. }
{Note that the model is unfair as it directly utilizes the sensitive feature as input (without any fairness constraints).}

\subsubsection{Policy Using Latent $Z$}
{The third option is to take decisions based on the latent variable $Z$ of our conditional VAE model.
Note that following Section~\ref{sec:learn_decide_fair}, we assume $\Z$ to contain the information of Y and at the same time be independent of sensitive information $S$ up to an approximation error. Therefore, we assume taking decisions (only) based on $Z$ to be (approximately) DP and CF fair.
See proofs in Appendix~\ref{apx:proofs}.
We train a \emph{separate policy model} to take decisions only based on $\Z$, i.e., $\pi^{Z}:= \Z \rightarrow \Uprox$. There are different options for training this model, i.e., either on labeled data only or on \emph{all} data.}
\begin{itemize}
    \item {\emph{Use only labeled data:} We train the policy model only on labeled data and correct for selective labeling with IPS. We term this policy $\pi^Z_\text{label}$.}
   \item {\emph{Use the classifier:} We use the classifier model to label \emph{all} data (both labeled and unlabeled data). Training thus requires no IPS correction. We term this policy $\pi^Z_\text{clf}$. } 
     \item {\emph{Use the decoder:} We use the decoder model to label \emph{all} data (both labeled and unlabeled data). Training thus requires no IPS correction. We term this policy  $\pi^Z_\text{dec}$. We use this policy in our evaluations in the main paper, whenever referring to \name.}
\end{itemize}

\paragraph{Practical Considerations}
{Figure~\ref{fig:ours_diff_pols} shows a comparison of different policy models.}
{For any policy model $\pi^Z$, we train a deep fully-connected neural network with the \emph{same} architecture that we select for the classifier model (from Table~\ref{tab:hyperparam_best}).
Note that selecting hyperparameters for these policy models separately could be further explored in future work.}

\subsection{Baselines}
\label{apx:baselines}
{In our evaluations in Section~\ref{sec:evaluation} we compare our method to the baseline method \nikifair, and the reference model \niki.
Both are state-of-the-art methods as shown in~\cite{kilbertus2020fair}.
In addition, in order to show the importance of unlabeled data in the decision-making process, we modify our model to provide two comparative benchmark models, \ipsvae\ and \nametwo.}

\subsubsection{\niki\ Policy Model}
We train a deep neural model $f_\nu$ based on the logistic model from~\cite{kilbertus2020fair}.
We optimize the following cost-sensitive cross-entropy loss Eq. (\ref{eq:cross-entropy}), which we also refer to as $\mathcal{L}^{\text{\niki}}$.
{We train on labeled data only and correct for the selective labels bias with IPS.}

\subsubsection{\nikifair\ Policy Model}
Following~\cite{kilbertus2020fair}, we train a fair version of \niki\ by adding a fairness constraint with a Lagrangian hyperparameter. We call this {model} \nikifair.
This model is trained on labeled data only and uses IPS correction. {We train by minimizing the following loss:}
\begin{equation}
    \mathcal{L}^{\text{\nikifair}}(\nu; \x, s, u) = \mathcal{L}^{\text{\niki}}(\nu, \x, s, u) + \lambda \cdot \text{DPU}(\nu; \x, s)
\end{equation}

{where $\mathcal{L}^{\text{\niki}}$ refers to Eq.~\ref{eq:cross-entropy}, and $\text{DPU}$ to Eq.~\ref{eq:dp}.}
The Lagrangian multiplier $\lambda \geq 0$ is tuned for a utility-fairness trade-off. {Note, setting $\lambda = 0$ would reduce \nikifair\ to the \niki\ model.}

\subsubsection{\ipsvae\ Model}
{We design a comparative benchmark model to understand the importance of unlabeled data in \phasetwo. In this, we assume our VAE model would use \emph{only labeled} data in \phasetwo.
We can consider this model to be a \emph{natural extension} of the \nikifair\ model using VAEs (and \phaseone\ pre-training).
This reduces loss in Eq.~\ref{eq:overall_loss} to the labeled ELBO, but \emph{with IPS correction}:}
\begin{equation}
\begin{split}
    \mathcal{L}^{\text{\ipsvae}}& (\theta, \phi; x,s, u) 
    = -{\frac{1}{\pi(d=1|\x, s)}}\Big[\E_{z \sim q_\phi (\Z|x,u,d=1)} \\
    & [\log p_{\theta}(x|z,s) + \log p_{\theta}(u|z,s,d=1)]\\
    &  - KL (q_{\phi}(\Z|x,u,s, d=1)||p(\Z)) \Big] 
\end{split}
\end{equation}

{Note that this model \emph{does use} unlabeled data in the pre-training phase (\phaseone), but it does not use unlabeled data in \phasetwo.
It does so by training an \nameone\ model (using Eq.~\ref{eq:phaseone_loss}) and then applying transfer learning.
In terms of policy learning, as this model only utilizes labeled data in \phasetwo, we use $\pi^Z_\text{label}$.}

\subsubsection{\nametwo\ Model}
{We design another benchmark model to understand the role of pre-training with unlabeled data in \emph{\phaseone}.
We modify our model such that there is no pre-training in \phaseone. That is, we \emph{do not} train a \nameone\ model in \phaseone, instead we directly start with \phasetwo.
As such, this model optimizes the \phasetwo\ loss in Eq.~\ref{eq:overall_loss}.
Note that we \emph{do not} optimize any \nameone\ model in \phaseone.}

\section{Additional Results}
\label{apx:results}
{In this section, we show additional results demonstrating that our \name\ method is able to learn more stable and fairer decision policies than comparable approaches. In the main paper, we reported results for the initial policy \HARSH\ and plotted different utility and fairness measures for the real-world datasets \COMPAS.
In this section, we report results with a more lenient initial policy \LENI\ for all datasets. In addition, we show plots for the real-world datasets \CREDIT\ and \MEPS\ for both initial policies, \HARSH\ and \LENI.}
\begin{figure*}[t]
	\centering
	\begin{subfigure}[b]{0.3\textwidth}
		\centering
		\includegraphics{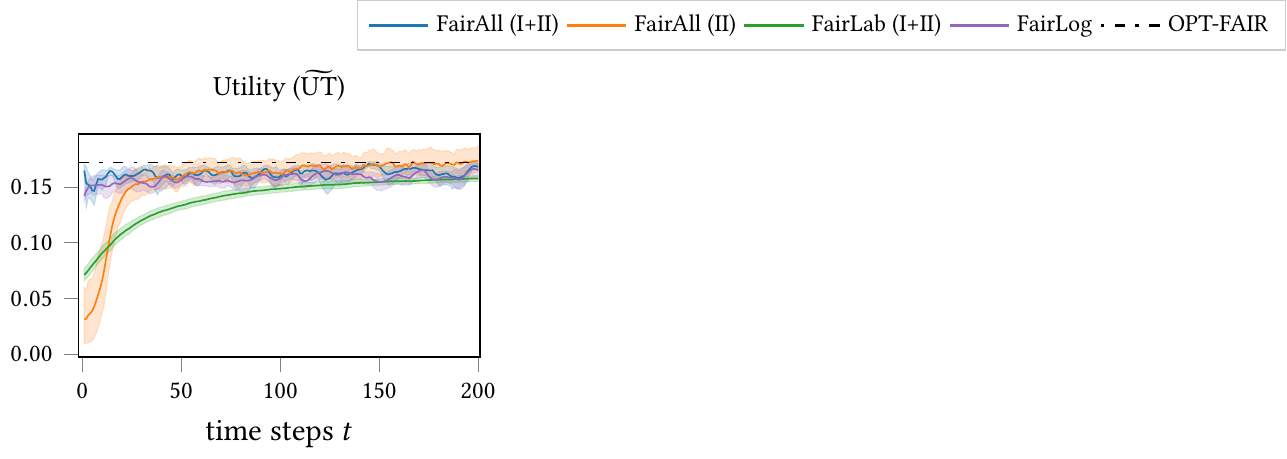}
	\end{subfigure}
	\begin{subfigure}[b]{0.3\textwidth}  
		\centering 
		\includegraphics{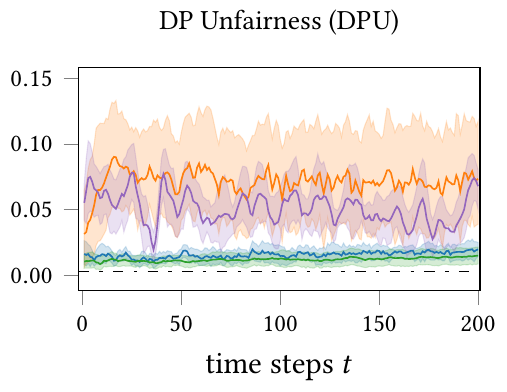}
	\end{subfigure}
	\begin{subfigure}[b]{0.3\textwidth}  
		\centering 
		\includegraphics{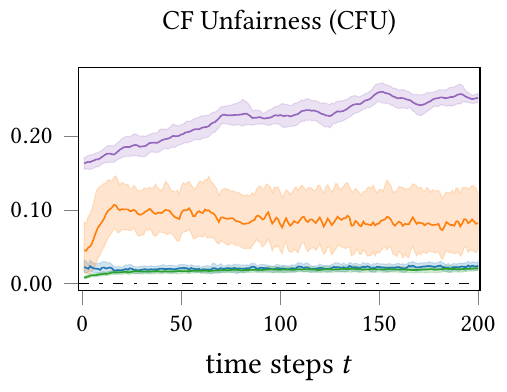}
	\end{subfigure}
\caption{Different measures for \SCB\ data. Initial policy \LENI.}
	\Description{If you require a verbal description of this figure, please contact us, the authors. We are happy to provide you with a description.}
\label{fig:scb_results_leni}
\end{figure*}

\begin{figure*}[t]
	\centering
	\begin{subfigure}[b]{0.31\textwidth}
		\centering
		\includegraphics{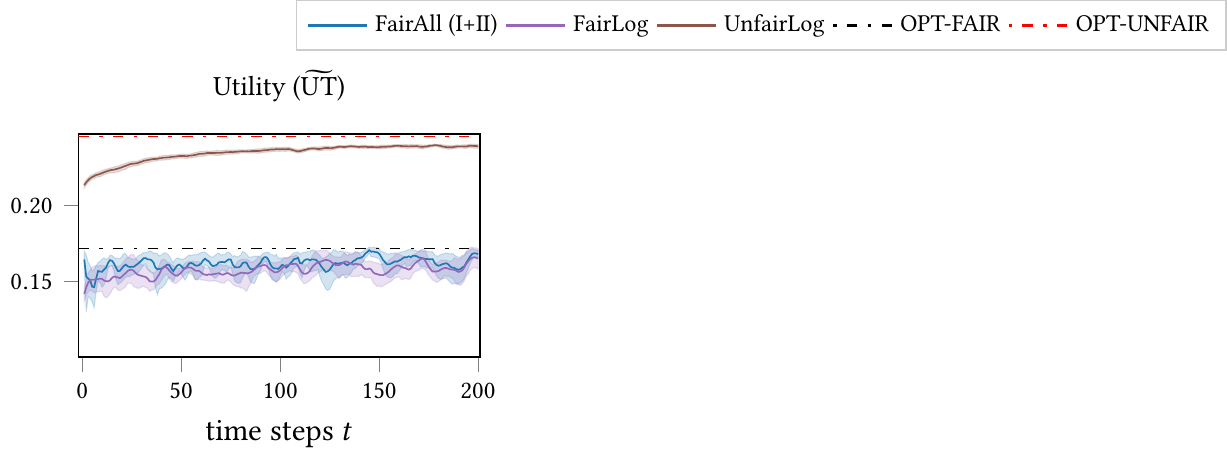}
	\end{subfigure}
	\begin{subfigure}[b]{0.31\textwidth}  
		\centering 
		\includegraphics{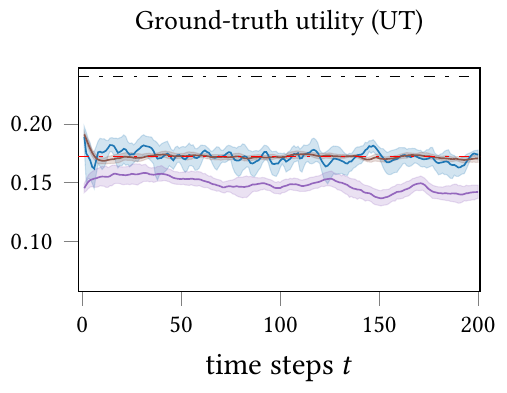}
	\end{subfigure}
	\begin{subfigure}[b]{0.31\textwidth}  
		\centering 
		\includegraphics{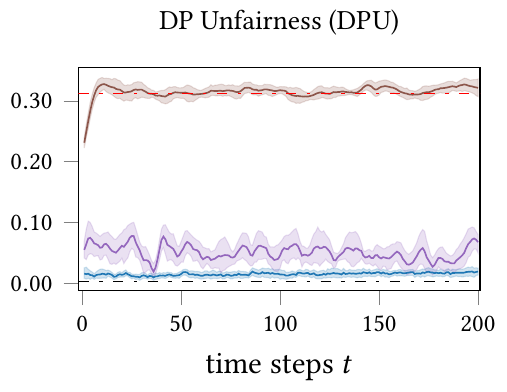}
	\end{subfigure}
\caption{Utility measured to ground truth for \SCB\ data. Initial policy \LENI.}
	\Description{If you require a verbal description of this figure, please contact us, the authors. We are happy to provide you with a description.}
\label{fig:scb_util_gnd_leni}
\end{figure*}

\subsection{Additional Synthetic Data Analysis}
We extend the results in Section~\ref{sec:eval_synth}.
Figure~\ref{fig:scb_results_leni} depicts results for the \SCB\ dataset with \LENI\ as initial policy $\pi_0$
Compared to the baseline \nikifair, we achieve similar utility convergence, while being significantly fairer regarding DP and CF.
Further,
the perceived trade-off between utility and fairness continues to be explained away if we were to measure to the unobserved ground truth $y$ instead of the proxy $\yprox$.

\subsection{Real-World: Effective Decision Learning}
Following Section~\ref{sec:eval_effective}, we measure the effectiveness of any decision learning process.
We compare the methods based on effective (accumulated) utility and DP unfairness (Def.~\ref{eq:utility_eff} and \ref{eq:dp_eff}) after 200 time steps.
Table~\ref{tab:util_fair_results_leni} summarizes the results.
We see that across all datasets, our method \name\ is able to provide significantly higher accumulated utility as well as lower unfairness.
Especially when comparing to the baseline and reference models \niki\ and \nikifair\ respectively, we see that our model performs significantly better, outperforming even the unfair \niki\ model in terms of utility.
We also see the clear benefit of using unlabeled data in both \phaseone\ and \phasetwo.
\name\ consistently manages to accumulate higher utility and lower unfairness compared to the benchmarks \ipsvae\ and \nametwo.

\begin{table*}[t]
\caption{Effective (accumulated) utility and demographic parity unfairness measured during the policy learning process for different real-world datasets after $t=200$ time steps. We consider \LENI\ initial policy here. We assume the cost of a positive decision of 0.5 for all datasets (in MEPS we assume 0.1). We report mean values over the same 10 independent seeds, with the numbers in the brackets representing the deviation. All reported values are scaled up by 100 for improved readability. Our method outperforms the baseline methods in most cases, achieving high utility while being less unfair.}
  \label{tab:util_fair_results_leni}
\small
\begin{tabular}{@{\extracolsep{4pt}}lcccccccccc}
\toprule
\multirow{2}{*}{\textbf{Model}} &
  \multicolumn{2}{c}{COMPAS} &
  \multicolumn{2}{c}{\CREDIT} &
  \multicolumn{2}{c}{\MEPS} \\
  \cline{2-3} \cline{4-5} \cline{6-7}
& Effect. \UTprox & Effect. \DPU & Effect. \UTprox & Effect. \DPU & Effect. \UTprox & Effect. \DPU \\ 
\midrule
\name &  {6.4 (0.8)} & {10.5 (0.6)} & {20.7 (0.5)} & {8.7 (1.8)} & {8.1 (0.3)} & {9.9 (1.3)} \\
\nametwo & 5.1 (0.6) & 10.6 (0.7) & 19.8 (1.0) & 10.8 (1.9) & 7.7 (0.2) & {9.4 (1.8)}\\
\ipsvae & 3.5 (0.5) & 10.9 (0.8) & 16.9 (0.9) & 10.4 (2.3) & 5.9 (0.6) & 10.3 (0.7)  \\
\nikifair & 3.5 (0.5) & 10.9 (1.0) & 19.5 (1.1) & {9.8 (1.9)} & 6.9 (0.4) & 11.2 (1.0) \\
\niki & 4.7 (0.6) & 15.1 (1.2) & {21.2 (0.5)} & 11.5 (2.0) & 7.7 (0.3) & 19.8 (2.4) \\
\bottomrule
\end{tabular}
\end{table*}

\begin{figure*}[H]
	\centering
	\begin{subfigure}[t]{0.22\textwidth}
		\centering
		\includegraphics{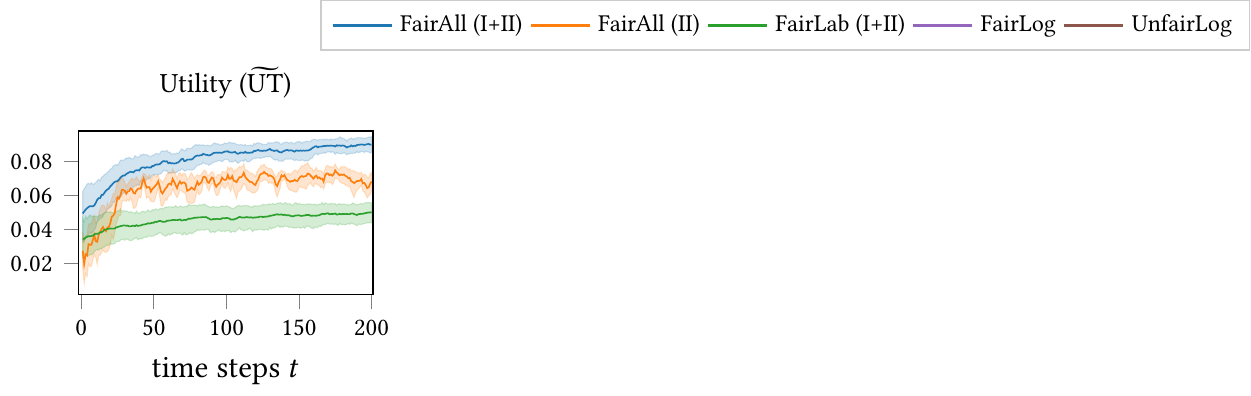}
	\end{subfigure}
	\hfill
	\begin{subfigure}[t]{0.22\textwidth}  
		\centering 
		\includegraphics{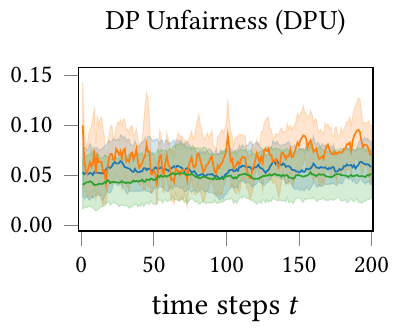}
	\end{subfigure}
	\hfill
	\begin{subfigure}[t]{0.22\textwidth}  
		\centering 
		\includegraphics{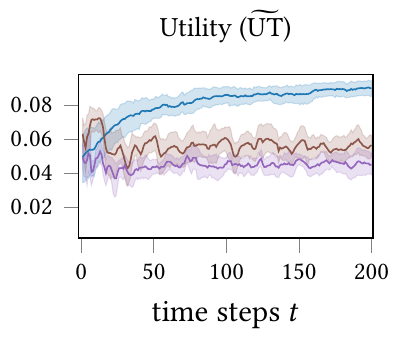}
	\end{subfigure}	
	\hfill
	\begin{subfigure}[t]{0.22\textwidth}   
		\centering 
		\includegraphics{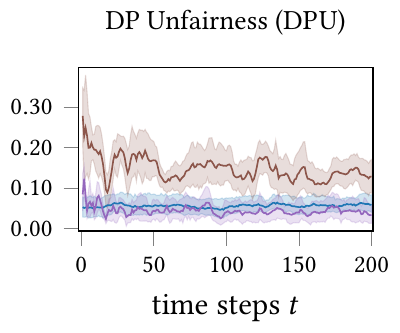}
	\end{subfigure}
	\caption{Different measures for \COMPAS\ data. Initial policy \LENI.}
    \Description{If you require a verbal description of this figure, please contact us, the authors. We are happy to provide you with a description.}
\end{figure*}

\subsection{Real-World: Deployment on Test Data}
Following Section~\ref{sec:eval_real_test}, we further show extensive evaluations for the deployment of the different methods as decision-making systems.
We see from the adjacent figures below that our method \name\ manages to converge to a significantly better utility level while being temporally more stable.
We see similar behavior with respect to DP unfairness.
Our method achieves the lower values of unfairness, which, again, is more stable temporally.
We see that when considering deployment, \name\ clearly outperforms the fair baseline and also the unfair reference model.
We also see the benefits of unlabeled data -- \ipsvae\ and \nametwo\ both fail to provide high, stable measures of utility.
These benchmark models also converge to worse levels of unfairness.
Further, Table~\ref{tab:temporal_unfair_leni} shows how our method provides the best trade-off between \emph{temporal variance} \TV\ and \emph{convergence level} $\mu$ across \emph{utility} and \emph{unfairness}.

\begin{figure*}
	\centering
	\begin{subfigure}[t]{0.22\textwidth}
		\centering
		\includegraphics{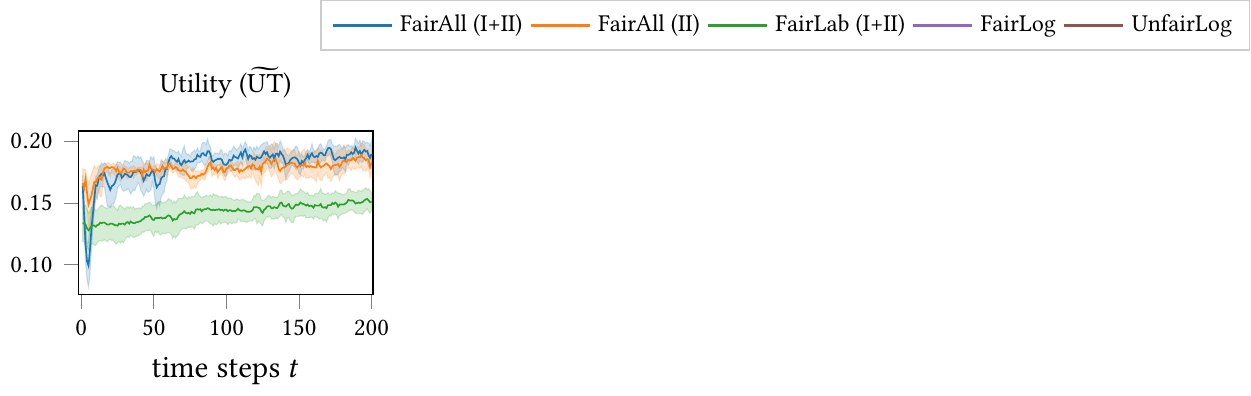}
	\end{subfigure}
	\begin{subfigure}[t]{0.22\textwidth}  
		\centering 
		\includegraphics{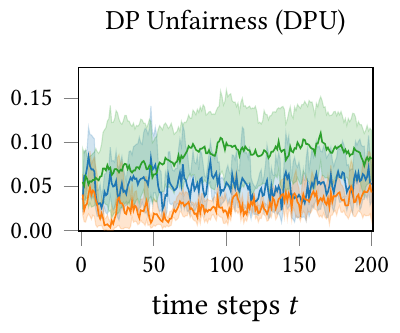}
	\end{subfigure}
	\begin{subfigure}[t]{0.22\textwidth}  
		\centering 
		\includegraphics{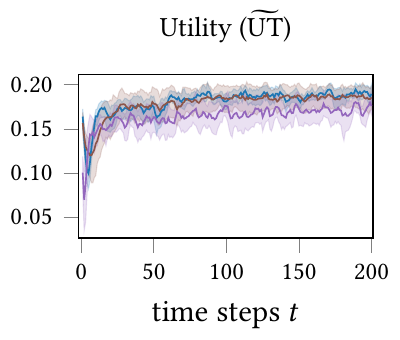}
	\end{subfigure}	
	\begin{subfigure}[t]{0.22\textwidth}   
		\centering 
		\includegraphics{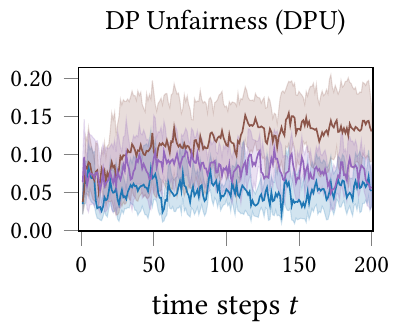}
	\end{subfigure}
	\caption{Different measures for \CREDIT\ data. Initial policy \HARSH.}
    \Description{If you require a verbal description of this figure, please contact us, the authors. We are happy to provide you with a description.}
\end{figure*}

\begin{figure*}
	\centering
	\begin{subfigure}[t]{0.22\textwidth}
		\centering
		\includegraphics{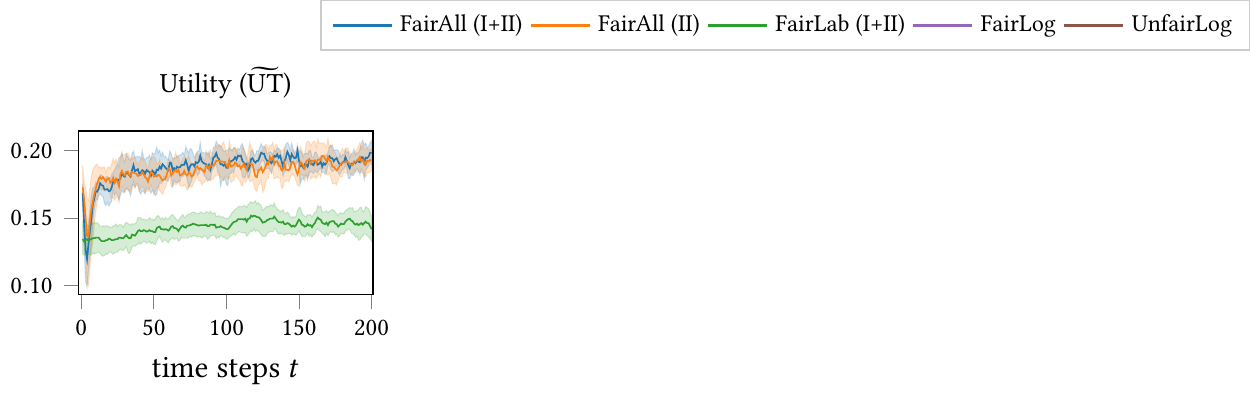}
	\end{subfigure}
	\begin{subfigure}[t]{0.22\textwidth}  
		\centering 
		\includegraphics{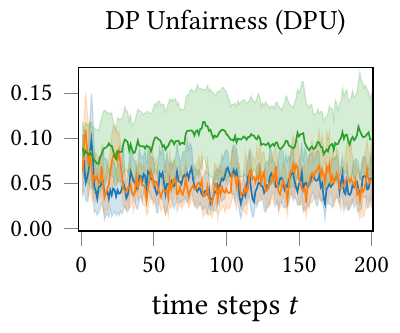}
	\end{subfigure}
	\begin{subfigure}[t]{0.22\textwidth}  
		\centering 
		\includegraphics{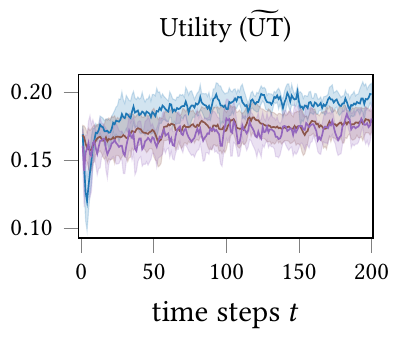}
	\end{subfigure}	
	\begin{subfigure}[t]{0.22\textwidth}   
		\centering 
		\includegraphics{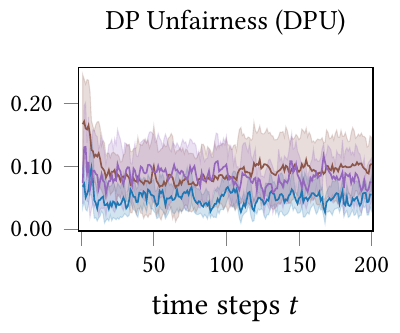}
	\end{subfigure}
	\caption{Different measures for \CREDIT\ data. Initial policy \LENI.}
	\Description{If you require a verbal description of this figure, please contact us, the authors. We are happy to provide you with a description.}
\end{figure*}

\begin{figure*}
	\centering
	\begin{subfigure}[t]{0.22\textwidth}
		\centering
		\includegraphics{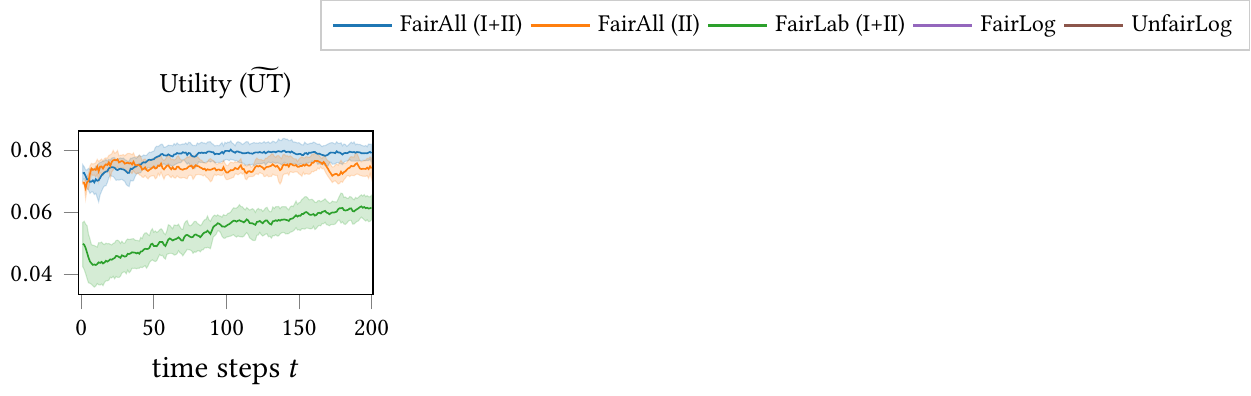}
	\end{subfigure}
	\begin{subfigure}[t]{0.22\textwidth}  
		\centering 
		\includegraphics{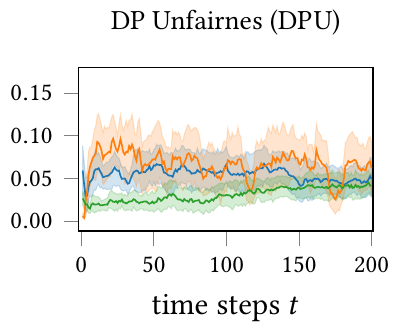}
	\end{subfigure}
	\begin{subfigure}[t]{0.22\textwidth}  
		\centering 
		\includegraphics{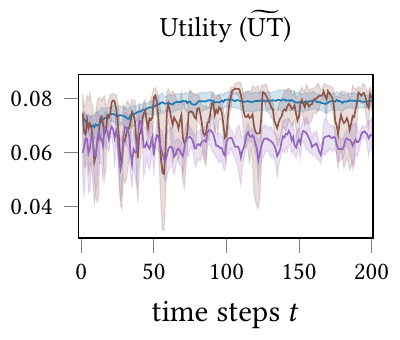}
	\end{subfigure}	
	\begin{subfigure}[t]{0.22\textwidth}   
		\centering 
		\includegraphics{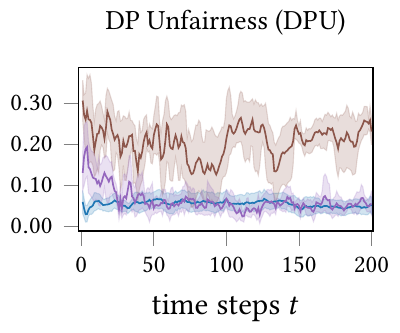}
	\end{subfigure}
	\caption{Different measures for \MEPS\ health data. Initial policy \HARSH.}
	\Description{If you require a verbal description of this figure, please contact us, the authors. We are happy to provide you with a description.}
\end{figure*}

\begin{figure*}
	\centering
	\begin{subfigure}[t]{0.22\textwidth}
		\centering
		\includegraphics{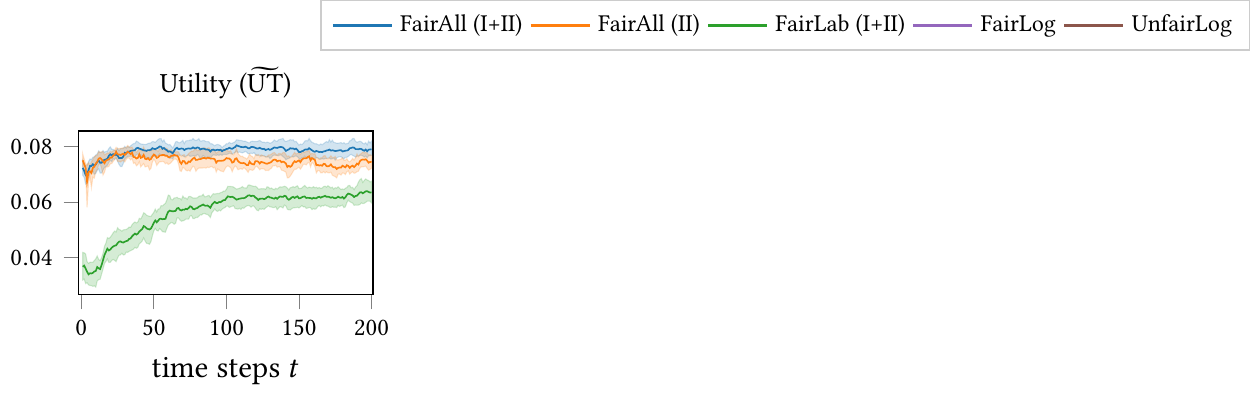}
	\end{subfigure}
	\begin{subfigure}[t]{0.22\textwidth}  
		\centering 
		\includegraphics{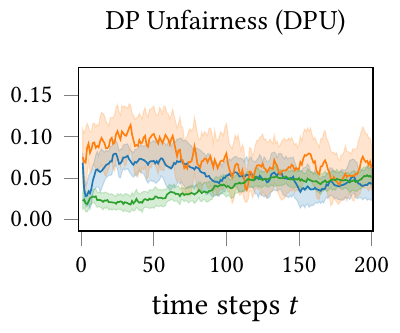}
	\end{subfigure}
	\begin{subfigure}[t]{0.22\textwidth}  
		\centering 
		\includegraphics{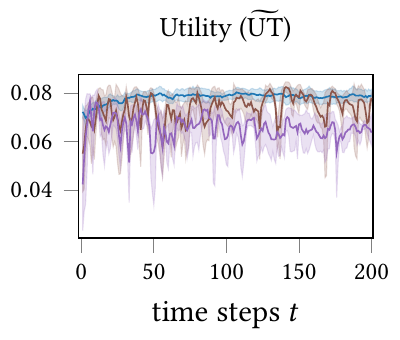}
	\end{subfigure}	
	\begin{subfigure}[t]{0.22\textwidth}   
		\centering 
		\includegraphics{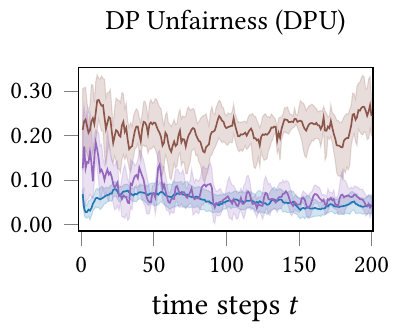}
	\end{subfigure}
	\caption{Different measures for \MEPS\ health data. Initial policy \LENI.}
	\Description{If you require a verbal description of this figure, please contact us, the authors. We are happy to provide you with a description.}
\end{figure*}

{ 
\renewcommand{\arraystretch}{1.2}
\begin{table*}[t]
\caption{Temporal variance  of utility and demographic parity unfairness, $\tvut$ and $\tvdp$, respectively, and the corresponding temporal mean $\muut$ and $\mudp$, respectively, for time interval $t=[125, 200]$. Results are shown for three real-world datasets. The assumed initial policy is \LENI. We report the mean over 10 runs with the standard deviation in brackets. For \TV, lower values are better, for $\mu$ higher (lower) is better for \UTprox\ (\DPU). Note all reported values are multiplied by 100 for improved readability.}
\label{tab:temporal_unfair_leni}
\resizebox{\textwidth}{!}{%
\begin{tabular}{@{\extracolsep{2pt}}lcccccc}
\toprule
\multirow{2}{*}{Model} &
  \multicolumn{2}{c}{\COMPAS} &
  \multicolumn{2}{c}{\CREDIT} &
  \multicolumn{2}{c}{\MEPS} \\ 
  \cline{2-3} \cline{4-5} \cline{6-7}
 &
  $\tvdp$($\downarrow$) |
   $\mudp$ ($\downarrow$)&
   $\tvut$ ($\downarrow$) |
  $\muut$ ($\uparrow$) &
  $\tvdp$ ($\downarrow$) |
  $\mudp$ ($\downarrow$) &
  $\tvut$ ($\downarrow$) |
  $\muut$ ($\uparrow$) &
   $\tvdp$ ($\downarrow$) |
  $\mudp$ ($\downarrow$) &
  $\tvut$ ($\downarrow$) |
  $\muut$ ($\uparrow$) \\ 
  \midrule
  \name &
  {1.0 (0.7)} |
  {5.8 (3.1)} &
  {0.3 (0.2)} |
  {8.8 (0.8)} &
  2.7 (1.6) |
  {4.9 (3.7)} &
  0.9 (0.4) |
  {19.3 (1.2)} &
  2.4 (2.7) |
  4.4 (3.4) &
  {0.2 (0.1)} |
  {7.9 (0.4)} \\ 
  \nametwo &
  2.9 (1.6) |
  7.6 (4.2) &
  0.7 (0.6) |
  7.0 (0.8) &
  2.8 (2.2) |
  {5.3 (4.0)} &
  0.8 (0.5) |
  19.1 (1.5) &
  3.4 (3.2) |
  6.2 (5.1) &
  0.3 (0.2) |
  7.4 (0.4) \\
  \ipsvae &
  {0.5 (0.4)} |
  {4.9 (4.1)} &
  {0.2 (0.1)} |
  4.9 (1.1) &
  {2.1 (1.8)} |
  9.5 (6.8) &
  {0.5 (0.4)} |
  14.6 (1.3) &
  {0.7 (0.4)} |
  {4.8 (1.7)} &
  {0.2 (0.2)} |
  6.2 (0.6) \\
\nikifair &
  1.6 (1.2) |
  4.2 (3.8) &
  0.5 (0.3) |
  4.5 (0.9) &
  4.0 (2.5) |
  7.9 (5.3) &
  1.2 (0.7) |
  17.3 (1.7) &
  2.2 (1.9) |
  5.6 (2.9) &
  0.7 (0.5) |
  6.4 (0.8) \\
  \bottomrule
\end{tabular}%
}
\end{table*}
}

\subsection{Cost Analysis}
\label{apx:eval_costs}
Finally, we perform a cost analysis for our decision-making algorithm.
The cost of making positive decisions in any decision-making scenario depends on the context.
It can change from one context to another, and with it, the expected utility and unfairness of any learned decision-making algorithm.
For example, in a healthcare setting, we might have to consider lower costs for positive decisions.
This would ensure we do not \emph{reject} anyone who needs access to critical healthcare and the necessary facilities.
Likewise, in loan settings, we might have to operate with higher costs.
It might happen that a decision-maker in the loan scenario needs to account for significantly higher costs whenever they provide a loan.
In this section, we show how varying the cost in a decision-making setting could affect the accumulated effective utility and unfairness, if we were to apply our decision-making algorithm in several real-world settings.

\begin{figure*}[h]
   \centering
      \begin{subfigure}[t]{0.42\textwidth}
          \centering
            \includegraphics{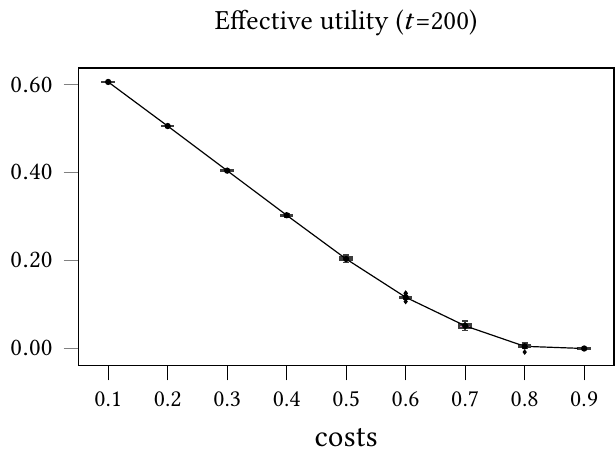}
          \caption{Effective utility ($\uparrow$) $u(d, \yprox)$ COMPAS}
      \end{subfigure}
      \begin{subfigure}[t]{0.42\textwidth}
          \centering
	       \includegraphics{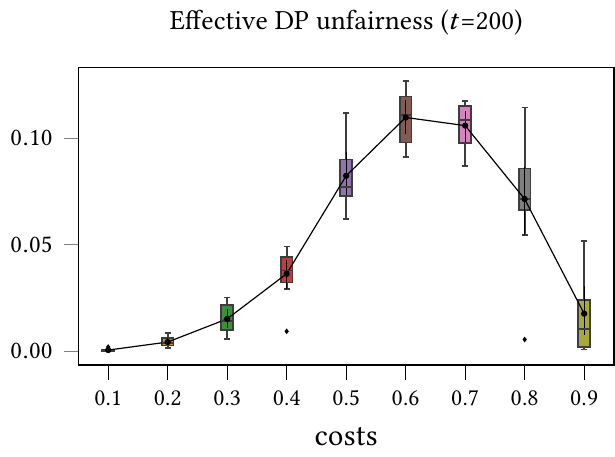}
	        \caption{Effective DP unfairness ($\downarrow$) COMPAS}
	  \end{subfigure}
	  
	  \begin{subfigure}[t]{0.42\textwidth}
		\centering
		\includegraphics{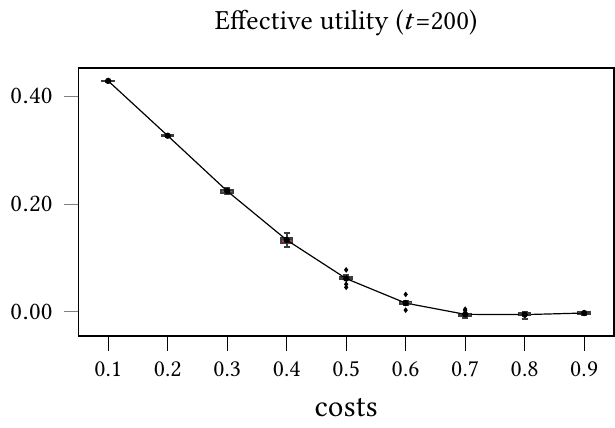}
		\caption{Effective utility ($\uparrow$) $u(d, \yprox)$ \CREDIT}
	\end{subfigure}
	\begin{subfigure}[t]{0.42\textwidth}
		\centering
		\includegraphics{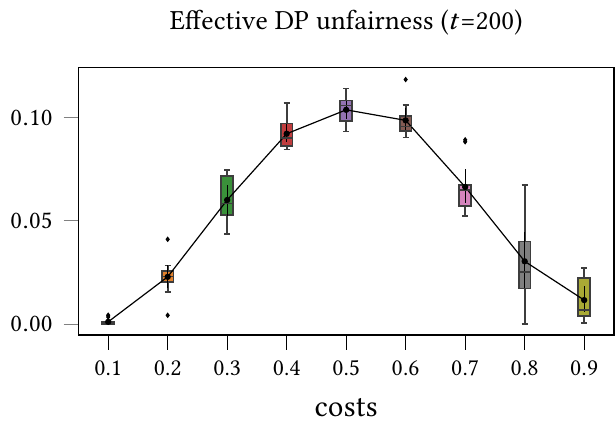}
		\caption{Effective DP unfairness ($\downarrow$) \CREDIT}
	\end{subfigure}
	
	\begin{subfigure}[t]{0.42\textwidth}
		\centering
		\includegraphics{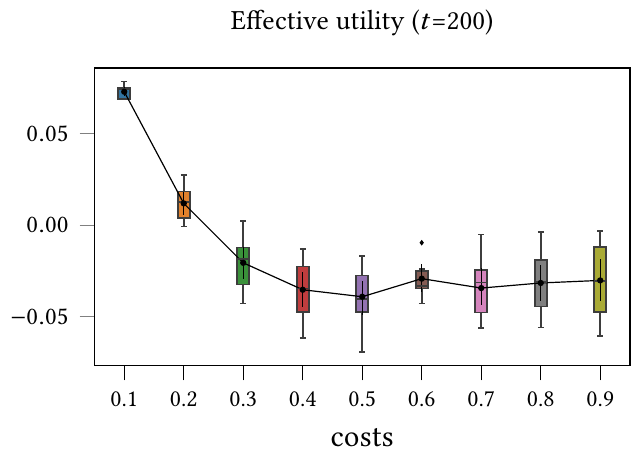}
		\caption{Effective utility ($\uparrow$) $u(d, \yprox)$ MEPS}
	\end{subfigure}
	\begin{subfigure}[t]{0.42\textwidth}
		\centering
		\includegraphics{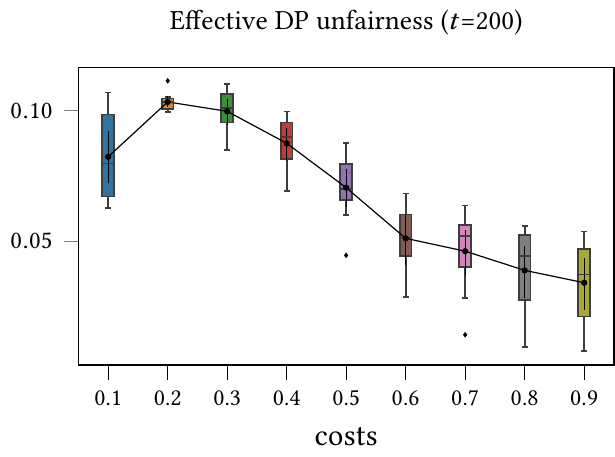}
		\caption{Effective DP unfairness ($\downarrow$) MEPS}
	\end{subfigure}
    \caption{Analysis of effective utility and effective demographic parity (DP) unfairness (at time $t=200$) on different real-world datasets for different values of the cost of the decision. We assumed initial policy $\pi_0=$ \HARSH.}
    \Description{If you require a verbal description of this figure, please contact us, the authors. We are happy to provide you with a description.}
    \label{fig:res_cost}
 \end{figure*}

Figure~\ref{fig:res_cost} shows the effect of cost on the acquired utility and fairness.
We see how at lower costs, we can acquire very high profits by accepting most people.
As cost increases, naturally, the acquired utility reduces as we start rejecting more and more people.
Correspondingly, the DP unfairness shows a more U-shaped relationship to cost.
At a very low (high) cost, unfairness is almost zero, as we start almost always accepting (rejecting) every individual.
At intermediate values of cost, we make more balanced accept/reject decisions, and thus see slightly higher unfairness.
Nonetheless, we believe performing such an analysis might give a better picture of a decision-making context, and help guide deployment and tuning the cost in related settings.

\section{Further Discussion of Assumptions}\label{apx:assumptions}

\subsection{Assumptions of the Generative Process}\label{apx:assumptions_generative}
We live in a world where the target of interest $\Y$ is not always independent of a social construct $S$. We give two prominent examples where this may not be the case.
For example, we can assume that the distribution of willingness to pay back a loan is independent of gender.
However, the ability to pay back -- which determines the utility of a bank -- may depend on gender (e.g., due to the gender pay gap). In other cases, e.g., in the medical domain, discrimination by sex may be justified. On a population level, biological sex and social gender are in general not independent, such that a decision that is dependent on sex may not be independent of gender.

\end{document}